\documentclass[twoside,11pt]{article}

\usepackage{blindtext}
\usepackage[preprint]{jmlr2e}
\usepackage{amsmath,amsfonts,amssymb}
\usepackage{bbm}
\usepackage{enumitem} 
\usepackage{floatrow} 
\usepackage{subcaption} 
\usepackage{xcolor}
\usepackage{breqn} 
\usepackage{rotating} 
\usepackage{soul} 

\usepackage{adjustbox}
\usepackage{longtable}
\usepackage{tabularx}
\usepackage{tablefootnote}
\usepackage{multirow}

\usepackage{booktabs} 

\newcommand{\cmmnt}[1]{}

\usepackage{easy-todo}

\usepackage{cancel}
\usepackage{ulem}

\usepackage{lastpage}
\jmlrheading{}{2026}{1-\pageref{LastPage}}{}{}{}{Christos Revelas, Otilia Boldea and Bas J.M. Werker}
\ShortHeadings{When do Random Forests work?}{Revelas, Boldea and Werker}
\firstpageno{1}

\begin{document}

\title{When do Random Forests work?}

\author{\name Christos Revelas \email revelas.christos@gmail.com\\
	\name Otilia Boldea \email o.boldea@tilburguniversity.edu\\
	\name Bas J.M. Werker \email b.j.m.werker@tilburguniversity.edu\\
       \addr Tilburg University \\
       Department of Econometrics and Operations Research\\
       Warandelaan 2, 5037 AB Tilburg, Netherlands
      }

\editor{My editor}

\maketitle

\begin{abstract}
We study the effectiveness of randomizing split-directions in random forests. Prior literature has shown that, on the one hand, randomization can reduce variance through decorrelation, and, on the other hand, randomization regularizes and works in low signal-to-noise ratio (SNR) environments. First, we bring together and revisit decorrelation and regularization by presenting a systematic analysis of out-of-sample mean-squared error (MSE) for different SNR scenarios based on commonly-used data-generating processes. We point out that there is no consensus in the literature on how the MSE is computed and the terms ``variance" and ``bias" are not uniquely defined. In the present paper, we obtain MSE decompositions by averaging over training sets first and then over covariates. We find that variance reduction tends to increase with the SNR and forests outperform bagging when the SNR is low because, in low SNR cases, variance dominates bias for both methods. Second, we show that the effectiveness of randomization is a question that goes beyond the SNR. We present a simulation study with fixed and moderate SNR, in which we examine the effectiveness of randomization for other data characteristics. In particular, we find that (i) randomization can increase bias in the presence of fat tails in the distribution of covariates; (ii) in the presence of irrelevant covariates randomization is ineffective because bias dominates variance; and (iii) when covariates are mutually correlated randomization tends to be effective because variance dominates bias. Beyond randomization, we find that, for both bagging and random forests, bias can be significantly reduced in the presence of correlated covariates. This last finding goes beyond the prevailing view that averaging mostly works by variance reduction. Given that in practice covariates are often correlated, our findings on correlated covariates could open the way for a better understanding of why random forests work well in many applications. 
\end{abstract}



\section{Introduction}

Random forests (\citealp{breiman2001random}) are one of the most popular methods for classification and regression estimation in machine learning and have been studied and applied extensively in the last two decades. A random forest is an average of random decision trees. The randomness comes from two sources: first, from sampling the data and, second, from randomizing the directions in which splitting is allowed. Averages of trees with data-sampling only, commonly denoted as bagging (\citealp{breiman1996bagging}) or subagging (\citealp{buhlmann2002analyzing}), have been studied in the literature as methods of their own. They have been shown to improve, in general, upon single trees, i.e., learners without any randomness given the data, through variance reduction. Averages of trees with only the split randomization, to our knowledge, have not been popular in practice, nor have they been extensively studied. Instead, split randomization was introduced on top of data sampling and has been mostly used as means to improve upon bagging. A natural question is then when split randomization improves upon bagging and why. 

The existing literature on the properties of forests and bagging is extensive. It addresses choices that a practitioner faces when implementing either method, from the number of trees in the ensemble (\citealp{scornet2017tuning}, \citealp{probst2018tune}, \citealp{10.1214/18-AOS1707}), to the size, or depth, of its underlying trees (\citealp{zhou2023trees}, \citealp{revelas2024does}), as well as theoretical properties, such as guarantees for consistency and asymptotic normality (\citealp{10.1214/15-AOS1321}, \citealp{mentch2016quantifying}, \citealp{wager2018estimation}, \citealp{cattaneo2022pointwise}, \citealp{chi2022asymptotic}, \citealp{revelas2024does}). Other work includes the study of simplified versions and generalizations of the standard CART (\citealp{breiman1984cart}) recursive partitioning scheme (\citealp{geurts2006extremely}, \citealp{biau2012analysis}, \citealp{10.1214/18-AOS1709}), ensembles where individual trees do not have the same weight (\citealp{chen2024optimal}, \citealp{beck2024hedged}), or the relation between tree-based methods and other learners (\citealp{lin2006random}, \citealp{curth2024random}). 
The existing literature addressing \textit{differences} between bagging and random forests, detailed hereafter, is narrower, and the present paper contributes to it.

The first explanation for random forest success is \textit{decorrelation}, an idea that goes back to \citet{breiman2001random}, who showed that the prediction error of an ensemble can be bounded by the product of the prediction error of any of its individual trees times the correlation between any two of its trees and then argued: the lower the correlation the better. He focused on classification forests and only provided simulations in a regression setting for forests of trees that split based on linear combinations of covariates, rather than the classical single-direction splits. The decorrelation idea was formalized in \citet{hastie2009elements}, who showed that the variance of an ensemble is, for a sufficiently large number of trees, approximately equal to the product of the pairwise correlation between trees times the variance of any individual tree in the ensemble, while the bias of the ensemble equals the bias of any individual tree. Then, they argue that forests can improve upon bagging by decorrelating trees in the ensemble, in turn reducing the ensemble's variance. They illustrate in a simulation example the existence of a bias-variance tradeoff associated with the number of directions considered at each split: the smaller this number, the smaller the correlation between trees, and hence the smaller the forest's variance, but the larger the bias, and vice versa. Still, it is not yet clear when decorrelation improves random forest performance relative to bagging and when it does not, depending on data characteristics, because split randomization tends to increase bias. 

The second explanation for random forest success prevailing in the literature is \textit{regularization}, and focuses on how much noise is in the data. \citet{ishwaran2015effect} does not compare bagging to forests but gives experimental evidence that random forests adapt well to noise and attributes this property to the CART splitting criterion. \citet{wyner2017explaining} attribute random forest performance to the combination of interpolation and averaging, arguing that this combination makes forests robust to noise: by growing large trees, any error is local, and by averaging, errors are smoothed out. However, they do not mention the split randomization and rather compare forests to trees. Moreover, as \citet{mentch2020randomization} point out, interpolation cannot happen with standard forests because of data sampling. Forests without data sampling, i.e., ensembles in which randomization only comes within the tree construction, can interpolate if fully grown. To our knowledge, \citet{mentch2020randomization} are the first to provide an in-depth comparison between forests (with both sources of randomness) and bagging (with data-sampling only). They show in simulations that split randomization reduces the degrees of freedom, and then, that forests outperform bagging in low and medium signal-to-noise ratio (SNR) data generating processes while bagging can outperform forests when the SNR is high, concluding that random forests outperform bagging in real-world datasets ``because real-world datasets tend to be noisy". \citet{arnould2023interpolation} and \citet{curth2024random} further support the arguments of interpolation and regularization respectively, but both consider forests without data sampling and compare them to trees. The second and last study, to our knowledge, to compare bagging with forests and discuss their relative performance given data characteristics, is \citet{liu2024randomization}. Contrary to \citet{mentch2020randomization}, their simulations show that in some cases, forests can outperform bagging even when the SNR is high. They introduce the concept of ``hidden pattern", essentially a narrow jump in the regression function, and argue that such patterns cannot be captured by bagging but can be captured by forests, thanks to split randomization, which, in such cases, can reduce bias. In sum, with the exception of \citet{liu2024randomization}, most of the existing literature compares the performance of forests to that of bagging with respect to the amount of noise in the data. In the present paper, we explore differences between the two methods with respect to other characteristics of the data, focusing on covariates. For the remainder of this paper, we use the term randomization to refer to split randomization. Our contribution is twofold.

First, we review previous explanations for random forest success. On the one hand, randomization has been shown to decorrelate trees. However, decorrelation alone does not guarantee a reduction in out-of-sample mean-squared error (MSE), as it can increase bias. In \citet{hastie2009elements}, the SNR is fixed at one, and we ask whether the decorrelation argument is still valid when the SNR is low or high. On the other hand, randomization has been shown to regularize predictions and be effective in noisy datasets and ineffective when the SNR is high. However, it is not clear what this (in)effectiveness is due to. \citet{mentch2020randomization} argue in favour of forests through variance reduction, yet, when reporting differences in MSE between the two methods, they do not show any bias-variance decomposition. In the present paper, we present a systematic simulation analysis of the MSE and its composition for different SNR scenarios using common data generating processes. We find that (i) in both low and high SNR environments, randomization decorrelates trees; and (ii) the decorrelation effect tends to increase with the SNR. Hence decorrelation alone does not explain the variation in performance across SNR values. We find that (iii) randomization decreases individual tree variance when the SNR is low and increases individual tree variance when the SNR is high; but (iv) thanks to decorrelation, randomization effectively reduces variance for every SNR scenario. Moreover (v) randomization increases squared bias in every SNR scenario with one exception. Still, these findings are not enough to explain the variation in performance across SNR values. We find and point out that (vi) for both bagging and forest, variance decreases and squared bias increases with the SNR. Consequently, we attribute forest success in low SNR cases to variance reduction \textit{because} variance dominates bias in these cases: when the SNR is low, variances are larger than biases, and hence, the improvement in variance due to randomization is more important than the deterioration in bias. 

Second, we explore other data characteristics than noise as defined in the literature. While prior literature has focused on the SNR as key determinant of forest performance, we present evidence of variation in MSE for datasets with a fixed, and moderate, SNR value, by extending the same common regression functions as in our review of previous explanations. We find that (vii) randomization can increase bias in the presence of tails in the distribution of covariates. when the regression function is nonlinear, making forests ineffective compared to bagging in those regions; (viii) adding irrelevant covariates increases bias for both bagging and forest, hence bias dominates variance in a MSE decomposition; (ix) in the presence of irrelevant covariates, randomization increases bias, hence bagging tends to outperform forest\footnote{ We are not the first to consider the impact of irrelevant covariates on forest performance. \citet{hastie2009elements} show, in a simulation example for classification, that the classification test error of forest increases with the number of irrelevant covariates, and attribute this deterioration to the decrease of the probability of relevant covariates being selected at each split. However, they fix the amount of randomization and do not compare forest to bagging. Yet, we find that, the presence of irrelevant covariates also affects bagging. \citet{mentch2020randomization} do consider three settings in the linear model where the number of relevant and total covariates is (5, 10), (5, 100) and (5, 1000) respectively, but they do not discuss the effect of adding irrelevant covariates and treat each case as a model on its own. \citet{liu2024randomization} discuss the effect of adding irrelevant covariates and show in a simulation example that, the more such covariates are added, the larger the optimal number of splitting directions becomes, hence, indirectly, supporting that for many irrelevant covariates, bagging can outperform. Yet, they consider up to 5 irrelevant covariates in a data generating process containing 7 covariates, and as we see in our simulations, few irrelevant covariates are often not enough to observe bagging outperformance. }; (x) when covariates are mutually correlated, for both bagging and forest, the MSE can be reduced compared to when covariates are independent and the stronger the correlation, the larger the improvement; (xi) this improvement is mainly due to bias and, while randomization increases bias, forests tend to outperform bagging in the presence of correlated covariates because variance dominates bias; (xii) in the presence of perfect mutual correlation randomization has zero effect in MSE: bagging and forest are essentially the same; and, finally, (xiii) while in the presence of irrelevant covariates, randomization increases MSE, if, additionally, covariates are correlated, then randomization reduces MSE enough to outperform bagging. Table \ref{table_intro} summarizes our findings.
\begin{table}[htbp]
\centering
\begin{tabular}{ccccc}
  & noise $\nearrow$ & irrelevant $\nearrow$ & $\rho \nearrow$ & tails \\
  \hline
  squared bias & $\searrow$ & $\nearrow$ & $\searrow$ & $\nearrow$ \\ 
  variance & $\nearrow$ & $\searrow$ & = & $\nearrow$ \\ 
  dominant & variance & bias & variance & bias \\
  preferred & forest & bagging & forest & bagging  \\
  \hline
  \hline
\end{tabular}
\caption{Summary of findings. The first two rows show the typical effect of changes in data characteristics on the bias and variance for \textbf{both} bagging and forests. Randomization almost always reduces variance and increases squared bias. Whether randomization is effective or not ultimately depends on whether bias or variance dominate mean-squared error.}
\label{table_intro}
\end{table}

For both of our contributions just presented, our task consists of comparing the performance of bagging with forests across data-generating processes (DGPs) by calculating the MSE, and, when possible, explaining the obtained differences by looking for example at bias and variance separately. In doing so, we encounter two challenges. 

First, we observe that there is no consensus in the literature on how the MSEs of the two methods are compared. Some studies provide the difference in MSE, others the relative difference or the difference divided by, e.g., the noise. In some DGPs, the MSE for both methods is large, i.e., both methods perform poorly, the MSE difference can be large in absolute value, while the relative difference is small. In other DGPs, the MSE for both methods is small, i.e., both methods perform well, and yet the relative difference in MSE can be large. In either case, can we really say that one method is empirically significantly better than the other? In the present paper, we show that the relative difference in MSE between bagging and forest is invariant to a normalization of the regression function's variance, as long as we keep the SNR fixed. We argue that, presenting differences in a normalized setup can increase the empirical significance of observed differences. For example, an improvement of 5\% due to randomization, may be more significant when the two methods have a MSE of roughly 20, compared to when their MSE is of the order of 1000. However, whether a 5\% improvement is empirically significant, independently of whether the MSE of the two methods is of the order of 20 or 1000, still depends on the application at hand. 

Second, we observe that there is no consensus on how the MSE is calculated, and in particular, decomposed into bias, variance, and irreducible error. In simulated datasets, typically several train and test sets are generated; a model is fit, or estimated, in each training set, before being evaluated on test sets. The unconditional MSE is an average of squared differences between the observed, test-set values of the dependent variable, and the predicted values obtained on observations of the training sets. The average is ultimately taken over both the training sets, i.e., prediction models, and the covariate values in the test sets. In some studies, the MSE is obtained by averaging over training sets first, for a given covariate value, while in others, squared differences are first averaged over covariate values for a given training set. While the order of averaging, or conditioning, does not impact the MSE, we find that, it does change bias and variance: decomposing the MSE conditionally on the prediction model first and then averaging each term in the decomposition over prediction models gives an unconditional bias term and an unconditional variance term. But, decomposing the MSE conditionally on covariate values first, then averaging, also gives an unconditional bias term and an unconditional variance term. We point out that the above unconditional bias and variance terms are not the same in the two cases and, hence, are not uniquely defined. Therefore, we need to be careful, when making statements of the form ``randomization increases bias" or ``randomization reduces variance", to be explicit in how ``bias" and ``variance" are defined. Finally, we note that in real datasets the MSE cannot be calculated or decomposed into bias and variance because the regression function is not known. We discuss the empirical applicability of our results in Section \ref{section_guidelines}.

The remainder of the paper is organized as follows. Section \ref{section_framework} sets the mathematical and simulation frameworks of our study. In particular, we fix a bias-variance decomposition of MSE and show that the relative difference in MSE between bagging and forests is normalization invariant. In Section \ref{section_replications} we review previous explanations for random forest success by replicating and complementing experiments presented in \citet{hastie2009elements}, \citet{mentch2020randomization} and \citet{liu2024randomization}. Section \ref{section_findings} contains our findings beyond SNR: we compare bagging to forests in the presence of tails, irrelevant, and correlated covariates. In Section \ref{section_guidelines} we discuss applications and future work. Section \ref{section_conclusion} concludes. Proofs and additional simulations are gathered in the appendices.


\section{Random Forests and Bagging for Regression}
\label{section_framework}
In this section we set the mathematical framework for the remainder of the paper. We are interested in a regression model of the form
\begin{equation}
Y=f(X)+\varepsilon
\label{regression_model}
\end{equation}
with $(X,Y)\in\mathbb{R}^p\times \mathbb{R}$ and $\varepsilon\in\mathbb{R}$ such that $\mathrm{E}[\varepsilon|X]=0$ and $\mathrm{E}[\varepsilon^2|X]=\sigma_{\varepsilon}^2<\infty$. We assume throughout $0<\mathrm{Var}(f(X))<\infty$. We want to estimate $f(x)=\mathrm{E}[Y|X=x]$ based on an i.i.d sample $D_n=\{(X_1,Y_1),\dots,(X_n,Y_n)\}$ satisfying (\ref{regression_model}). 
Tree estimators can be written as
\begin{equation}
T_n(x) = \sum_{i=1}^n W_{n,i}(x) Y_i \text{\ \ with \ } W_{n,i}(x) = \frac{\mathbbm{1}_{X_i\in C_n(x)}}{\sum_{j=1}^n\mathbbm{1}_{X_j\in C_n(x)}}
\label{tree_estimator}
\end{equation}
where $C_n(x)$ is the cell that contains $x$ after partitioning $\mathbb{R}^p$ based on $D_n$. The bagged estimator can be written as
\begin{equation}
\bar{T}^{*}_{n}(x) = \frac{1}{B}\sum_{b=1}^B\sum_{i=1}^n W^*_{n,b,i}(x) Y_i
\label{bagging_estimator}
\end{equation}
where $B$ is the number of trees and $W^*_{n,b,i}(x)$ is the random weight for $Y_i$ in the $b^{\text{th}}$ tree. $W^*_{n,b,i}(x)$ is not only random due to $D_n$ but also conditionally on $D_n$. This conditional randomness comes from bootstrapping and is represented in notation by the star $(^*)$. 
Similarly, the random forest estimator can be written as
\begin{equation}
\bar{T}^{*\dagger}_{n,m}(x) = \frac{1}{B}\sum_{b=1}^B\sum_{i=1}^n W^{*\dagger}_{n,m,b,i}(x) Y_i
\label{forest_estimator}
\end{equation}
where again $B$ is the number of trees and $W^{*\dagger}_{n,m,b,i}(x)$ is the random weight for $Y_i$ in the $b^{\text{th}}$ tree. Here the conditional randomness comes from bootstrapping \textit{and} random selection of covariates at each split. The subscript $m$ denotes the number of covariates drawn at each split and is fixed and the same for every split and every tree in the forest. It corresponds to the $m_{try}$ parameter in \textbf{R}'s randomForest package. Randomness due to the selection of covariates, i.e., split randomization, is represented by a dagger $(^\dagger)$.

\subsection{Performance comparison in practice: unconditional MSE}
We wish to compare the performance of forest to that of bagging. To do so, for each method we consider the unconditional, or ``out-of-sample", mean-squared error
\begin{equation}
\text{MSE}(\hat{f}):=\mathrm{E}\Big[\Big(Y-\hat{f}(X)\Big)^2\Big]
\label{unconditional_MSE_def}
\end{equation}
where $\hat{f}$ is either $\bar{T}^{*}_{n}$ or $\bar{T}^{*\dagger}_{n,m}$. The expectation is taken with respect to both the estimate $\hat{f}$ and out-of-sample, or ``test", observations $(X,Y)$. We can show the following.
\begin{proposition}[bias - variance decomposition]\label{proposition_decomposition_MSE_starting_given_X}
For $X$, $\varepsilon$, and $Y$ satisfying (\ref{regression_model}) and any estimator $\hat{f}$ which is independent of $X$ and $\varepsilon$, we have
\begin{equation}
\mathrm{E}\Big[\Big(Y-\hat{f}(X)\Big)^2\Big] = \mathrm{E}\Big[\Big(f(X)-\mathrm{E}[\hat{f}(X)|X]\Big)^2\Big] + \mathrm{E}\Big[\mathrm{Var}\big[\hat{f}(X)|X\big]\Big] + \sigma_{\varepsilon}^2
\label{precise_decomposition}
\end{equation}
\end{proposition} 
A proof is given in Appendix \ref{appendix_proofs}. The independence assumption in Proposition \ref{proposition_decomposition_MSE_starting_given_X} is satisfied in practice when datasets to train our models are independent from datasets used to evaluate them, i.e., the common train and test sets distinction. The right hand side of (\ref{precise_decomposition}) is a decomposition of the MSE into three terms: a bias term, a variance term, and an irreducible error term. The irreducible error is the best possible performance and can only be achieved by an estimator with zero bias and zero variance. A natural question when we observe in practice that a method outperforms another in terms of MSE is whether improvements come from ``bias reduction" or from ``variance reduction". We use the above decomposition to answer this question in the present paper. We point out, however, that decomposition (\ref{precise_decomposition}) is obtained by first calculating the MSE conditionally on $X$ and then using the law of total expectation to average $X$ out. Alternatively, we could condition on $\hat{f}$ first. In such case, we obtain the following decomposition 
\begin{equation}
\mathrm{E}\Big[\Big(Y-\hat{f}(X)\Big)^2\Big] = \mathrm{E}\Big[\Big(\mathrm{E}[f(X)-\hat{f}(X)|\hat{f}]\Big)^2\Big] + \mathrm{E}\Big[\mathrm{Var}\big[f(X)-\hat{f}(X)|\hat{f}\big]\Big] + \sigma_{\varepsilon}^2
\label{precise_decomposition_alternative}
\end{equation}
where, again, we have three terms: a bias term, a variance term, and an irreducible error. A proof of (\ref{precise_decomposition_alternative}) is also given in the appendix. The irreducible error is the same in both decompositions. However, the bias and variance terms are different between both decompositions, and hence, ``unconditional bias" and ``unconditional variance" are \textit{not} uniquely defined. We also observe this in practice: in simulations, one method can outperform another in terms of MSE, the improvement coming from the bias term in one decomposition and from the variance term in the other decomposition, and vice versa. We point this out to stress the importance of being explicit in statements of the form ``forests outperform bagging in this or that example due to, e.g., bias reduction". Based on decomposition (\ref{precise_decomposition}), which is the one we use in the present paper, an improvement in bias means an improvement, on average over $X$, of the expected, on average over trained models $\hat{f}$, at a given $X=x$, squared bias, and similarly for the variance. We've encountered both decompositions in the literature. For example, in \citet{hastie2009elements}, Equation (7.9) in Section 7.3, gives the bias-variance decomposition of the MSE conditionally on $X$, and the expectations are taken over the prediction models $\hat{f}$. This is also the approach adopted in \citet{curth2024random}, Section 4.1.1, where they define ``statistical bias" as the difference between the true regression function and its expectation over prediction models for a given value of $X$. In \citet{liu2024randomization}, the MSE decomposition presented in Equation (8) of Section 3.3, is, although not precisely written, conditional on $\hat{f}$, and expectations are taken over $X$. In our case, we use decomposition (\ref{precise_decomposition}) because some of our findings are dependent on $X$, as well as because we formulate \citet{breiman2001random}'s decorrelation argument in favour of forests conditionally on $X$, which is the focus of the next subsection. Finally, we note that in the present paper we compare bagging to forest in simulated datasets. When the two methods are studied in real datasets, the true regression function $f$ is not known, and hence, neither decomposition (\ref{precise_decomposition}) or (\ref{precise_decomposition_alternative}) can be applied. What we've observed in prior literature is that the average prediction at a given value is used as a ``proxy" for the true regression function. However, in that case, the bias term as presented in our decomposition (\ref{precise_decomposition}) is effectively eliminated.

\subsection{Decorrelation of trees and limitations}
We now derive the bias and variance of bagging and forests in terms of the bias and pairwise correlation of each method's individual trees, conditionally on $X$. Rewrite (\ref{bagging_estimator}) and (\ref{forest_estimator}) as averages of trees, i.e., 
\begin{equation}
\bar{T}^{*}_{n}(x) = \frac{1}{B}\sum_{b=1}^B T^{*}_{n,b}(x)
\label{bagging_estimator_2}
\end{equation}
and
\begin{equation}
\bar{T}^{*\dagger}_{n,m}(x) = \frac{1}{B}\sum_{b=1}^BT^{*\dagger}_{n,m,b}(x)
\label{forest_estimator_2}
\end{equation}
respectively. Again, we will use the notation $\hat{f}$ for either $\bar{T}^{*}_{n}$ or $\bar{T}^{*\dagger}_{n,m}$. Moreover, we note $\hat{f}_b$ for the $b^{th}$ individual tree in the ensemble, i.e., either $\bar{T}^{*}_{n,b}$ or $\bar{T}^{*\dagger}_{n,m,b}$. 
\begin{proposition}\label{proposition_decorrelation}
For both bagging and forests, for any $x$, we have 
\begin{equation}
\mathrm{E}[\hat{f}(x)] = \mathrm{E}[\hat{f}_1(x)]
\label{bias_ensemble}
\end{equation}
and, as $B\rightarrow\infty$, 
\begin{equation}
\mathrm{Var}[\hat{f}(x)]  \rightarrow \mathrm{Corr}[\hat{f}_1(x), \hat{f}_2(x)] \ \mathrm{Var}[\hat{f}_1(x)]
\label{variance_ensemble}
\end{equation}
where $\mathrm{Corr}$ denotes the correlation between predictions $\hat{f}_1(x)$ and $\hat{f}_2(x)$. 
\end{proposition} 
A proof is given in Appendix \ref{appendix_proofs}. The above expectations, variances and correlation, are all taken with respect to $\hat{f}$. Equation (\ref{bias_ensemble}) says that, the bias of bagging and the bias of a random forest each equal the bias of any of their respective individual trees, conditionally on $X$. However, $\mathrm{E}[\bar{T}^{*}_{n,1}(x)]$ and $\mathrm{E}[\bar{T}^{*\dagger}_{n,m,1}(x)]$ are in general \textbf{not} equal, i.e., for a given $x$, the bias of an individual tree in bagging is different from the bias of an individual tree in a forest. Similarly, (\ref{variance_ensemble}) says that, the variance of bagging and the variance of a random forest each approximately equal the product of the variance of any of their individual trees times their pairwise correlation, but the variance of a tree in bagging is not in general the same as the variance of a tree in a forest, and the same holds for the correlations. The rationale behind the decorrelation argument in favour of random forests, as hinted by \citet{breiman2001random} and illustrated in \citet{hastie2009elements}, and which is expressed here locally\footnote{ Note that (\ref{bias_ensemble}) and  (\ref{variance_ensemble}) hold for \textit{every} $x$, and hence, they also hold globally, i.e., on average over $x$.}, is that, by introducing randomness in the directions in which splitting is allowed, individual trees within the ensemble might lose in terms of bias, as some splits are not considered, hence, the ensemble's bias will increase, while at the same time, trees within the ensemble will be less correlated, due to the additional randomness, effectively reducing the correlation factor in the ensemble's variance. In simulations (Section \ref{section_replications}), we observe that the effect of randomization on individual tree variance can go both ways: in low SNR cases, randomization reduces individual tree variance, while when the SNR is high, randomization tends to increase individual tree variance. Ultimately, in our simulations, randomization always reduces ensemble variance. However, randomization can increase bias, and the effect on MSE depends on the data-generating process (DGP).

\subsection{Normalization, SNR and relative difference in MSE}
We aim to compare bagging and forests across different DGPs. Since the MSE is sensitive to the variance of $f(X)$, performance across regression functions and distributions of $X$ can be misleading. In order to address this, except when mentioned otherwise\footnote{ This is the case in Section \ref{section_replications} where we replicate empirical findings of previous literature.}, we will normalize the variance of $f(X)$ to be equal to one, i.e., we replace $f(X)$ by
\begin{equation}
\frac{f(X)}{\sqrt{\mathrm{Var}(f(X))}}
\label{regression_model_normalized}
\end{equation}
where $\mathrm{Var}(f(X))$ is either known for a given $f$ or calculated numerically. We note $\sigma_{f}^2:=\mathrm{Var}(f(X))$. 
We then consider datasets with varying amount of noise. We follow the definition of signal-to-noise ratio (SNR) of \citet{mentch2020randomization}, also used in \citet{liu2024randomization}, given by
\begin{equation}
\text{SNR} := \frac{\sigma_{f}^2}{\sigma_{\varepsilon}^2}
\label{SNR_definition}
\end{equation}
which, when $f(X)$ has variance one, simplifies to $\text{SNR} = 1/\sigma_{\varepsilon}^2$. 
In order to compare the performance of bagging and forest we consider the relative difference of their MSEs
\begin{equation}
\Delta_r := \frac{\text{MSE(bagging)}-\text{MSE(forest)}}{\text{MSE(forest)}}
\label{relative_performance}
\end{equation}
where each term is defined as in (\ref{unconditional_MSE_def}). Then, we can easily show the following. 
\begin{proposition}\label{proposition_relative_performance}
Given (\ref{regression_model}), define $Z=\frac{Y}{\sigma_{f}}$, $g(x)=\frac{f(x)}{\sigma_{f}}$ and $\eta=\frac{\varepsilon}{\sigma_{f}}$ such that
\begin{equation}
Z=g(X)+\eta.
\label{regression_model_by_sigma_f}
\end{equation} 
Note $\Delta_r(Z)$ and $\Delta_r(Y)$ the relative difference in MSE between bagging and forest for models (\ref{regression_model_by_sigma_f}) and (\ref{regression_model}) respectively. Then $\Delta_r(Z) = \Delta_r(Y)$. 
\end{proposition} 
A proof is given in Appendix \ref{appendix_proofs}. Proposition \ref{proposition_relative_performance} says that, for a \textit{fixed} SNR, whether we normalize the regression function to have variance one or not, the relative difference in MSE between bagging and the random forest does not change. 

Finally, in simulations we add an indication of the statistical significance of MSE differences between bagging and forests, something we did not see in the literature. This is useful because differences in MSE between bagging and forest are often of small order, and, in simulations, we find that, even with many replications, MSE differences may or may not be statistically significant. 
Letting $Z:=MSE(bagging)-MSE(forest)$ for any \textit{given} training set, if $\bar{Z}$ is the average of $Z$ over training sets, and $S$ is the standard deviation of $Z$ over training sets, then, noting $W$ the number of training sets, for large $W$,
\begin{equation}
T:=\frac{\bar{Z}}{S/\sqrt{W}} \approx \mathcal{N}(0,1)
\label{test_statistic_distribution}
\end{equation}
the proof of which follows from the independence of training sets and finite variances.


\section{Literature Replications: Review of Previous Findings}
\label{section_replications}
In this section, we review previous explanations for random forest success. We find that, decorrelation alone does not explain differences in MSE across SNR scenarios: randomization tends to increase bias; randomization always reduces variance, and does so by more in case of high SNR than for low SNR. With high SNR both bagging and forest tend to have smaller variance and larger bias relative to low SNR. Hence, in balance, randomization tends to be effective in reducing MSE when the SNR is low and ineffective when the SNR is high.  

We consider the following three DGPs taken from \citet{hastie2009elements}, \citet{mentch2020randomization}, and \citet{liu2024randomization}. 
\begin{enumerate}
\item $\mathcal{N}$-LINEAR: 
\begin{equation}
Y=\sum_{j=1}^{5} X_j+\varepsilon
\label{model_N_linear}
\end{equation}
with i.i.d. $\mathcal{N}(0,1)$ covariates. 
\item $\mathcal{U}$-MARS: 
\begin{equation}
Y=10\sin(\pi X_1 X_2) + 20(X_3-0.05)^2 + 10X_4 + 5X_5+\varepsilon
\label{model_U_mars}
\end{equation}
with i.i.d. $\mathcal{U}(0,1)$ covariates\footnote{ Note that in older literature the MARS model is defined with $20(X_3-\textbf{0.5})^2$ instead of $20(X_3-\textbf{0.05})^2$; we follow more recent references, but we did not observe any differences when we tried the original version.}. 
\item $\mathcal{U}$-HIDDEN: 
\begin{equation}
Y = X_1 - \mathbbm{1}(0.6\leq X_2\leq0.65)+\varepsilon
\label{model_U_hidden}
\end{equation}
with i.i.d. $\mathcal{U}(0,1)$ covariates. 
\end{enumerate}
Note that these are not normalized using (\ref{regression_model_normalized}). On the one hand, we chose these DGPs to replicate the above studies, and on the other hand, due to their characteristics. The linear and mars models have been extensively used in papers on trees and forests. The latter, due to its interaction terms and nonlinearity, resembles real datasets. The hidden pattern has the particularity of being discontinuous. In every model, $\varepsilon|X\sim\mathcal{N}(0,\sigma_{\varepsilon}^2)$. In this section, the experimental setup varies: for the LINEAR model, we use training sets of size $n=100$ as in \citet{hastie2009elements}\footnote{ Note that in \citet{hastie2009elements} the linear model is already normalized; moreover, it has more covariates, but this does not affect what we wish to replicate here, i.e., the decorrelation effect of forests.}; for the forest, the number of covariates to draw, denoted $m_{try}$, is set to $\lfloor p/3 \rfloor$, where $p$ is the number of covariates and $\lfloor \ \rfloor$ denotes the integer part; for the MARS model, we take $n=200$ and $m_{try}=\lfloor p/3 \rfloor$ as in \citet{mentch2020randomization}; for the HIDDEN pattern model, $n=500$ and $m_{try}= \lfloor p/2 \rfloor$ as in \citet{liu2024randomization}. Note that effectively, $m_{try}=1$ in every case, i.e., the maximum possible randomization.

For each model, we generate 500 training sets of the above respective sizes $n$. On each training set, we evaluate bagging ($m_{try}=p$) and forests ($m_{try}=1$). All other parameters of the randomForest package in \textbf{R}\footnote{ Implementations throughout this paper were executed with \textbf{R} version 4.4.2 (2024-10-31).} are left to their default values, i.e., ensembles consist of 500 trees, and trees are grown large, with terminal nodes having 5 observations, in line with the above references. We generate a \textit{single} test set of size $10,000$ and evaluate each trained model on this test set. We note that, in the literature, we often observe that for each training set a new test set is generated. From a statistical point of view, this means adding variation to the evaluation, which is something we chose to avoid, but we verified that this choice does not impact our findings. From the obtained predictions, for each method, we report the unconditional MSE (\ref{unconditional_MSE_def}) estimated on the test set, as well as the squared bias, variance, and irreducible error as in decomposition (\ref{precise_decomposition}). Moreover, we report the average, over $X$ (test observations), individual tree variance and pairwise correlation as in the right-hand side of (\ref{variance_ensemble}). Then, we report the test statistic value as in (\ref{test_statistic_distribution}) for the difference between the two methods' MSE. Finally, we report the relative difference in MSE as defined in (\ref{relative_performance}). We repeat the above for three values of SNR: 0.05 (low), 1 (moderate) and 6 (high) and report the obtained results in the first three columns of Tables \ref{replication_table_linear}, \ref{replication_table_mars}, and \ref{replication_table_hidden} for each model respectively. Then, for each model, we repeat the same procedure but we normalize the regression function: we start by drawing 100,000 separate observations of $X$ which we use to calculate $\sigma_{f}=\sqrt{\mathrm{Var}(f(X))}$ and replace $f(X)$ by $f(X)/\sigma_{f}$ and $\varepsilon$ by $\varepsilon/\sigma_{f}$. These results are given in columns four to six. 

We start with the LINEAR model shown in Table \ref{replication_table_linear} and look at the original regression function (first three columns). 
\begin{table}[htbp]
\small
\centering
\begin{tabular}{ccccccc}
  & \multicolumn{3}{c}{Original} & \multicolumn{3}{c}{Normalized} \\
  \hline
  SNR & 0.05 & 1 & 6 & 0.05 & 1 & 6 \\ 
  $\sigma_{f}$ & 2.23 & 2.23 & 2.23 & 1 & 1 & 1 \\ 
  $\sigma_{\varepsilon}$ & 9.99 & 2.23 & 0.91 & 4.47 & 1 & 0.41 \\ 
  \hline
  $Bias^2$ bagging & 0.38 & 0.66 & 0.93 & 0.08 & 0.13 & 0.18 \\ 
  $Bias^2$ forest & 0.89 & 1.13 & 1.32 & 0.18 & 0.23 & 0.26 \\ 
  Variance bagging & 15.15 & 1.06 & 0.45 & 2.98 & 0.22 & 0.09 \\ 
  Variance forest & 8.44 & 0.53 & 0.21 & 1.66 & 0.11 & 0.04 \\ 
  Tree variance bagging & 101.48 & 7.03 & 3.11 & 19.85 & 1.42 & 0.63 \\ 
  Tree variance forest & 88.31 & 7.03 & 3.57 & 17.85 & 1.43 & 0.73 \\ 
  Correlation bagging & 0.15 & 0.15 & 0.15 & 0.14 & 0.15 & 0.14 \\
  Correlation forest & 0.09 & 0.07 & 0.06 & 0.09 & 0.07 & 0.07 \\ 
  Irreducible & 100.32 & 5.02 & 0.84 & 20.11 & 1.01 & 0.17 \\ 
  MSE bagging & 115.73 & 6.70 & 2.20 & 23.13 & 1.35 & 0.44 \\ 
  MSE forest & 109.47 & 6.63 & 2.34 & 21.91 & 1.33 & 0.47 \\ 
  test statistic & 48.37 & 5.90 & -21.41 & 45.80 & 7.12 & -20.97 \\ 
  \hline
  relative difference (\%) & \textbf{5.72} & \textbf{1.09} & \textbf{-6.28} & \textbf{5.59} & \textbf{1.40} & \textbf{-6.40} \\ 
   \hline
\end{tabular}
\caption{Decorrelation and SNR in the case of $\mathcal{N}$-LINEAR,  $n = 100$. Randomization decorrelates more when the SNR is high. The terms $Bias^2$, Variance and MSE are defined as in (\ref{precise_decomposition}). Tree variance (respectively Correlation) are measures for individual trees (respectively pairs of trees) in the ensemble. From Proposition \ref{proposition_decorrelation}, Variance $\approx$ Tree variance $\times$ Correlation. A positive relative difference, defined in (\ref{relative_performance}), means that the forest outperforms bagging.}
\label{replication_table_linear}
\end{table}
The relative difference between the two methods applied to (\ref{model_N_linear}) for a SNR of 1 is 1.09\%. This is an example that illustrates the decorrelation effect of randomization, in line with \citet{hastie2009elements}. Indeed, forests reduce variance by $1.06-0.53=0.53$ while increasing squared bias by $1.13-0.66=0.47$, hence, in balance, because the reduction in variance is larger than the increase in bias, the MSE is lower for forests compared to bagging. Moreover, the reduction in variance comes from decorrelation. Indeed, the individual tree variances are approximately the same for bagging and forest ($\approx7.03$) while the pairwise correlation is 0.15 for bagging and 0.07 for forest. Then, we ask what happens when the SNR is low or high. For every SNR, forest has larger squared bias and lower variance compared to bagging. For both methods, the squared bias increases and the variance decreases with the SNR. Hence, in balance for the MSE, changes in variance due to randomization are more important when the SNR is low, while changes in bias due to randomization are more important when the SNR is high. This is indeed what we observe: for a low SNR, forest outperforms bagging by more than 1\% (the relative difference is 5.72\%) and for a high SNR bagging outperforms forest (-6.28\%), and this change in relative performance is due to whether bias, or variance, dominate the out-of-sample MSE. Finally, we look at pairwise correlation and individual tree variance. For bagging, the pairwise correlation is similar across SNRs (0.15 in all three SNR cases when rounded to two decimals). For forest, the pairwise correlation decreases from 0.09 to 0.06. Individual tree variance is reduced by randomization when SNR is low (101.48 for bagging versus 88.31 for forest) and is increased when SNR is high (3.11 for bagging versus 3.57 for the forest). This shows that, in this example, the fact that forest outperforms bagging in low SNR, is \textit{not} due to ``more decorrelation" between trees. In fact, randomization decorrelates more when the SNR is high. Instead, what we observe, is that, while decorrelation helps in reducing variance (from 0.45 for bagging to 0.21 for forest), bagging outperforms in the high SNR case because of an increase in squared bias due to randomization (from 0.93 for bagging to 1.32 for forest). Note that the above observations are all also true in the normalized framework. 

Next, we look at the MARS model. Table \ref{replication_table_mars} replicates and extends simulations presented in \citet{mentch2020randomization}. 
\begin{table}[htbp]
\small
\centering
\begin{tabular}{ccccccc}
  & \multicolumn{3}{c}{Original} & \multicolumn{3}{c}{Normalized} \\
  \hline
  SNR & 0.05 & 1 & 6 & 0.05 & 1 & 6 \\ 
  $\sigma_{f}$ & 7.11 & 7.11 & 7.11 & 1 & 1 & 1 \\ 
  $\sigma_{\varepsilon}$ & 31.78 & 7.11 & 2.90 & 4.47 & 1 & 0.41 \\ 
  \hline
  $Bias^2$ bagging & 2.80 & 2.90 & 3.67 & 0.05 & 0.06 & 0.07 \\ 
  $Bias^2$ forest & 6.40 & 7.29 & 8.10 & 0.12 & 0.14 & 0.16 \\ 
  Variance bagging & 120.09 & 8.51 & 2.92 & 2.41 & 0.17 & 0.06 \\ 
  Variance forest & 70.13 & 4.23 & 1.51 & 1.39 & 0.09 & 0.03 \\ 
  Tree variance bagging & 1022.15 & 61.44 & 20.34 & 20.17 & 1.23 & 0.40 \\ 
  Tree variance forest & 901.82 & 68.58 & 32.74 & 17.97 & 1.39 & 0.66 \\ 
  Correlation bagging & 0.11 & 0.14 & 0.14 & 0.12 & 0.14 & 0.14 \\ 
  Correlation forest & 0.07 & 0.06 & 0.05 & 0.07 & 0.06 & 0.05 \\ 
  Irreducible & 1026.01 & 51.30 & 8.55 & 20.32 & 1.02 & 0.17 \\ 
  MSE bagging & 1148.27 & 62.74 & 15.18 & 22.77 & 1.25 & 0.30 \\ 
  MSE forest & 1101.73 & 62.72 & 18.14 & 21.81 & 1.24 & 0.36 \\ 
  test statistic & 61.82 & 0.17 & -58.38 & 56.79 & 1.84 & -61.25 \\ 
  \hline
  relative difference (\%) & \textbf{4.22} & \textbf{0.02} & \textbf{-16.31} & \textbf{4.39} & \textbf{0.27} & \textbf{-16.61} \\ 
   \hline
\end{tabular}
\caption{SNR and relative performance for $\mathcal{U}$-MARS,  $n = 200$. In low SNR, randomization is effective because variance dominates bias. The terms $Bias^2$, Variance and MSE are defined as in (\ref{precise_decomposition}). Tree variance (respectively Correlation) are measures for individual trees (respectively pairs of trees) in the ensemble. From Proposition \ref{proposition_decorrelation}, Variance $\approx$ Tree variance $\times$ Correlation. A positive relative difference, defined in (\ref{relative_performance}), means that the forest outperforms bagging.}
\label{replication_table_mars}
\end{table}
In their Figure 4, the top right plot shows the difference between the MSE of bagging minus the MSE of forest in the case of $\mathcal{U}$-MARS for $n=200$. We find differences of 46.53, 0.02 and -2.96 for low, moderate and high SNR respectively, which indeed replicates their results. What is not shown in \citet{mentch2020randomization} is the underlying MSE for each method, which we provide here. Notice the drastic change in the scale of MSE, for example for forest, ranging from 1101.73 in low SNR to 18.14 in high SNR. While in absolute terms the difference between the two methods is much larger in low SNR ($1148.27-1101.73=46.53$) than it is in high SNR ($15.18-18.14=-2.96$), in relative terms, the difference is much larger in high SNR ($-16.31\%$) than it is in low SNR ($4.22\%$). Our observations on decorrelation made for the linear model are also true here. In particular, our results reinforce \citet{mentch2020randomization}'s regularization argument in favour of forests. Indeed, in low SNR, forest performance is due to variance reduction (variance decreases from 120.09 to 70.13, while the increase in bias is of the order of 3), and in high SNR, forests increase squared bias enough (from 3.67 to 8.10, compared to a reduction in variance from 2.92 for bagging to 1.51 for forest) to be outperformed by bagging. Moreover, these findings are also true in a normalized framework, hence independent of the scale of MSE. Noting that in the original model in low SNR both methods seemingly perform badly (very large MSE), one could question whether a 5\% improvement is empirically relevant. This depends on the application at hand. Showing the same improvement in the normalized framework (in which MSEs are much smaller), further supports the significance of the differences in MSE. Finally, we observe that in the case of $\mathcal{U}$-MARS, for moderate SNR, MSEs for bagging and forest are \textit{not} statistically different.

\citet{liu2024randomization} argue in favour of forests through bias reduction, yet, in their simulations, trees are fully grown and we would a priori expect both methods to have little to no bias. For their hidden-pattern example, they do not show any bias-variance decomposition. We replicate this example and find that indeed, in high signal-to-noise, forests reduce bias, and this happens on top of variance reduction. Table \ref{replication_table_hidden} shows performance for the HIDDEN model. 
\begin{table}[htbp]
\small
\centering
\begin{tabular}{ccccccc}
  & \multicolumn{3}{c}{Original} & \multicolumn{3}{c}{Normalized} \\
  \hline
  SNR & 0.05 & 1 & 6 & 0.05 & 1 & 6 \\ 
  $\sigma_{f}$ & 0.36 & 0.36 & 0.36 & 1 & 1 & 1 \\ 
  $\sigma_{\varepsilon}$ & 1.63 & 0.36 & 0.15 & 4.47 & 1 & 0.41 \\ 
  \hline
  $Bias^2$ bagging & 0.012 & 0.010 & 0.013 & 0.096 & 0.07 & 0.10 \\ 
  $Bias^2$ forest & 0.012 & 0.007 & 0.006 & 0.095 & 0.05 & 0.04 \\ 
  Variance bagging & 0.47 & 0.03 & 0.010 & 3.51 & 0.22 & 0.08 \\ 
  Variance forest & 0.39 & 0.02 & 0.005 & 2.92 & 0.17 & 0.04 \\  
  Tree variance bagging & 2.43 & 0.15 & 0.05 & 18.35 & 1.10 & 0.38 \\ 
  Tree variance forest & 2.26 & 0.14 & 0.04 & 17.20 & 1.04 & 0.32 \\ 
  Correlation bagging & 0.19 & 0.20 & 0.20 & 0.19 & 0.20 & 0.20 \\ 
  Correlation forest & 0.17 & 0.17 & 0.13 & 0.16 & 0.16 & 0.13 \\
  Irreducible & 2.59 & 0.13 & 0.02 & 19.62 & 0.98 & 0.16 \\ 
  MSE bagging & 3.07 & 0.17 & 0.04 & 23.21 & 1.27 & 0.34 \\ 
  MSE forest & 2.99 & 0.16 & 0.03 & 22.61 & 1.20 & 0.24 \\ 
  test statistic & 101.98 & 65.13 & 62.24 & 95.75 & 62.91 & 58.54 \\ 
  \hline
  relative difference (\%) & \textbf{2.66} & \textbf{6.22} & \textbf{37.21} & \textbf{2.62} & \textbf{6.08} & \textbf{37.64} \\ 
   \hline
\end{tabular}
\caption{$\mathcal{U}$-HIDDEN,  $n = 500$. The terms $Bias^2$, Variance and MSE are defined as in (\ref{precise_decomposition}). Tree variance (respectively Correlation) are measures for individual trees (respectively pairs of trees) in the ensemble. From Proposition \ref{proposition_decorrelation}, Variance $\approx$ Tree variance $\times$ Correlation. A positive relative difference, defined in (\ref{relative_performance}), means that the forest outperforms bagging.}
\label{replication_table_hidden}
\end{table}
In their Section 3.1.2, \citet{liu2024randomization} give the test MSE for bagging (0.046) and forest (0.029) and their relative difference (37\%) in a simulation study of the $\mathcal{U}$-HIDDEN model with $n=500$ in high SNR ($=6$). In the same framework, we indeed\footnote{ Note that we used default tree depth (5 observations per terminal node) while in \citet{liu2024randomization} trees are reportedly grown at full depth (1 observation per node).} find test MSEs respectively of 0.044 and 0.032 for a relative difference of 37.21\%. Even though MSEs are small, all the observed differences are statistically significant. Here, MSEs in high SNR are extremely small, and showing that the relative difference is invariant by normalization, in which case MSEs are larger, only further supports the 37\% improvement's significance. A bias-variance decomposition as in (\ref{precise_decomposition}) shows a reduction in squared bias by a factor of 2 in favour of forests (from 0.013 to 0.006) in high SNR. This is the main difference between this example and the previous two examples. Our findings on variance do not change: randomization reduces variance and the decorrelation effect tends to increase with the SNR. 



\section{Tails, Irrelevant, and Correlated Covariates}
\label{section_findings}
We extend previous findings starting from the LINEAR, MARS, and HIDDEN models, introduced in (\ref{model_N_linear}), (\ref{model_U_mars}), and (\ref{model_U_hidden}) respectively, and show that differences in MSE between bagging and forest can occur in data generating processes with a fixed and moderate SNR. Therefore, the choice between the two methods goes beyond noise as defined in the literature. First, we find that randomization can increase bias in the presence of tails in the distribution of covariates. Second, in the presence of irrelevant covariates, MSE increases for both bagging and forest due to bias and, because randomization increases bias in balance with variance, randomization can hurt predictions. Third, we find that, for both bagging and forest, when covariates are correlated, the MSE can be reduced compared to when covariates are independent of each other. Moreover, the stronger the correlation the larger the improvement. This improvement in MSE for both methods is mainly due to bias and, hence, because randomization increases bias, forests tend to outperform bagging when covariates are correlated. Moreover, in the presence of perfect correlation, bagging and forests are essentially the same. 
Throughout this section, with the exception of Section \ref{subsection_varying_SNR}, we consider moderate signal to noise in a normalized setup: SNR $= 1$ and $\sigma_{f} = \sigma_{\varepsilon} = 1$. We fix training set size at $n=250$ for every model and $m_{try}=\lfloor p/3 \rfloor$ for forests (where $p$ is the number of covariates). We chose $n=250$ in order to be close to the average dataset size observed in previous studies. Performance evaluation is again done on a single test set of size 10,000 based on 500 training sets. 

\subsection{Tails in the distribution of covariates}
\label{subsection_distribution_of_X}
We show that a simple change in the distribution of covariates can drastically change the relative difference in MSE between bagging and forests. 

Table \ref{extension_table_distribution_of_X} shows the changes in MSE when changing the distribution of every covariate from uniform to normal and vice versa. 
\begin{table}[htbp]
\small
\centering
\begin{tabular}{ccccccc}
  & \multicolumn{2}{c}{LINEAR} & \multicolumn{2}{c}{MARS} & \multicolumn{2}{c}{HIDDEN} \\
    \hline
  distribution of $X_j$ & $\mathcal{N}(0,1)$ & $\mathcal{U}(0,1)$ & $\mathcal{N}(0,1)$ & $\mathcal{U}(0,1)$ & $\mathcal{N}(0,1)$ & $\mathcal{U}(0,1)$ \\
  \hline
  $Bias^2$ bagging & 0.08 & 0.05 & 0.10 & 0.05 & 0.01 & 0.12 \\ 
  $Bias^2$ forest & 0.15 & 0.12 & 0.36 & 0.13 & 0.02 & 0.10 \\ 
  Variance bagging & 0.15 & 0.15 & 0.18 & 0.16 & 0.22 & 0.25 \\ 
  Variance forest & 0.08 & 0.08 & 0.10 & 0.08 & 0.17 & 0.19 \\ 
  Tree variance bagging & 1.29 & 1.28 & 1.19 & 1.20 & 0.92 & 1.15 \\ 
  Tree variance forest & 1.34 & 1.34 & 1.49 & 1.35 & 0.97 & 1.11 \\ 
  Correlation bagging & 0.12 & 0.12 & 0.13 & 0.13 & 0.23 & 0.21 \\ 
  Correlation forest & 0.05 & 0.05 & 0.07 & 0.06 & 0.18 & 0.17 \\ 
  Irreducible & 1.01 & 1.02 & 1.01 & 1.02 & 1.00 & 0.98 \\ 
  MSE bagging & 1.234 & 1.219 & 1.30 & 1.23 & 1.23 & 1.35 \\ 
  MSE forest & 1.230 & 1.217 & 1.48 & 1.22 & 1.19 & 1.26 \\ 
  test statistic & 3.80 & 1.40 & -98.72 & 2.73 & 46.18 & 50.16 \\ 
  \hline
  relative difference (\%)  & \textbf{0.37} & \textbf{0.15} & \textbf{-12.37} & \textbf{0.36} & \textbf{3.21} & \textbf{7.13} \\ 
   \hline
\end{tabular}
\caption{Changes in the distribution of $X_j$. In the presence of tails and nonlinearities, bagging outperforms forests. The terms $Bias^2$, Variance and MSE are defined as in (\ref{precise_decomposition}). Tree variance (respectively Correlation) are measures for individual trees (respectively pairs of trees) in the ensemble. From Proposition \ref{proposition_decorrelation}, Variance $\approx$ Tree variance $\times$ Correlation. A positive relative difference, defined in (\ref{relative_performance}), means that the forest outperforms bagging.} 
\label{extension_table_distribution_of_X}
\end{table}
In the case of the LINEAR model, changing the distribution of covariates has no significant impact on relative difference in MSE. In the case of MARS, the relative difference drops from 0.36\% to -12.37\%. Moreover, this drop is due to bias: when changing covariate distribution from $\mathcal{U}(0,1)$ to $\mathcal{N}(0,1)$, the squared bias of bagging increases by a factor 2 (from 0.05 to 0.1) while that of forest increases by a factor of almost 3 (from 0.13 to 0.36). Similarly for HIDDEN: forest deteriorates, relative to bagging, when changing from uniform to normal covariates and this deterioration seems to also come from an increase in bias. Note in particular that while forest decreases squared bias relative to bagging in the case $\mathcal{U}(0,1)$, which is what \citet{liu2024randomization} claimed, this is no longer true in the case $\mathcal{N}(0,1)$. 

\begin{figure}[htbp]
\begin{subfigure}{0.19\textwidth}
	\includegraphics[width=\textwidth]{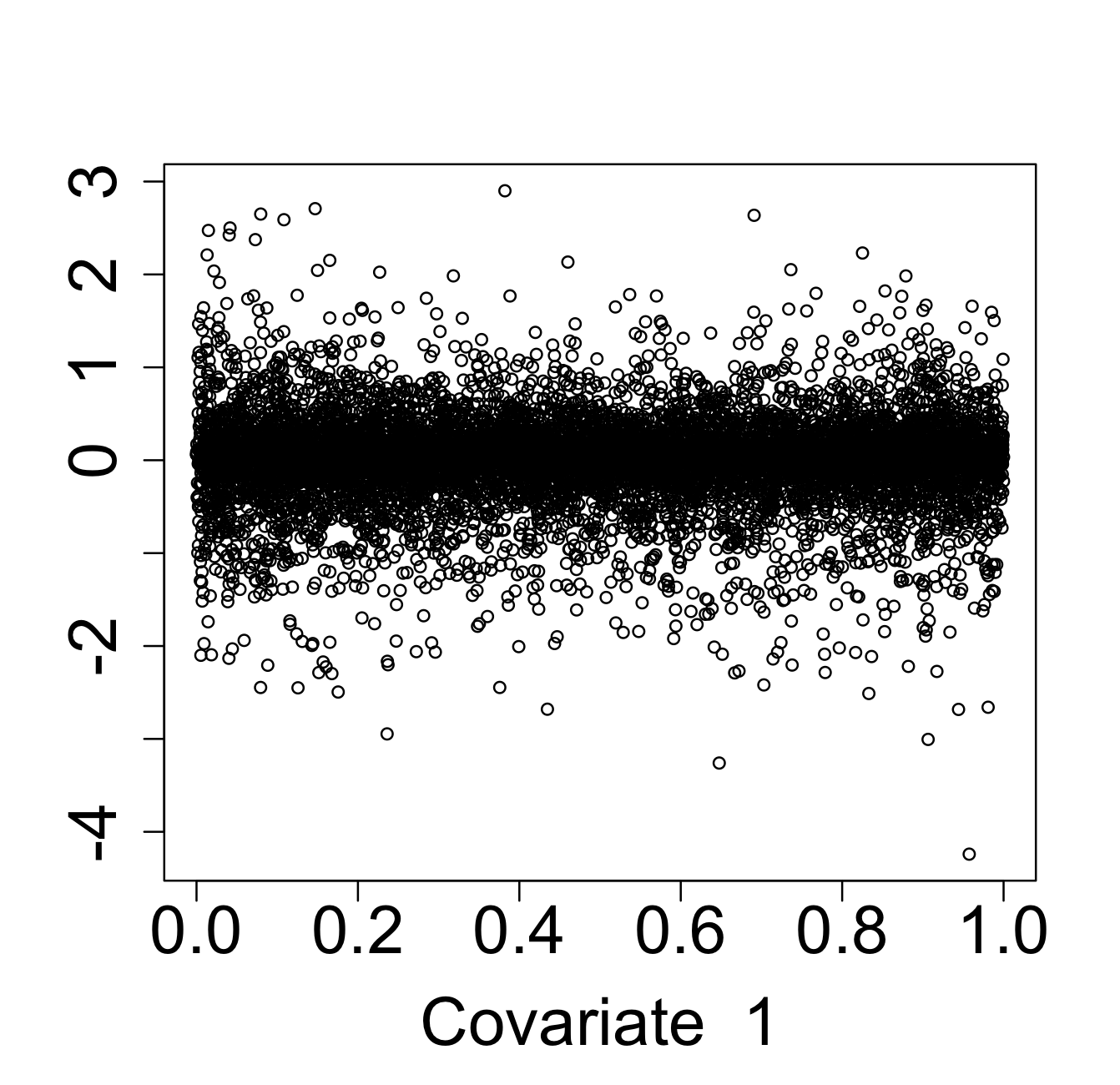}
\end{subfigure}
\begin{subfigure}{0.19\textwidth}
	\includegraphics[width=\textwidth]{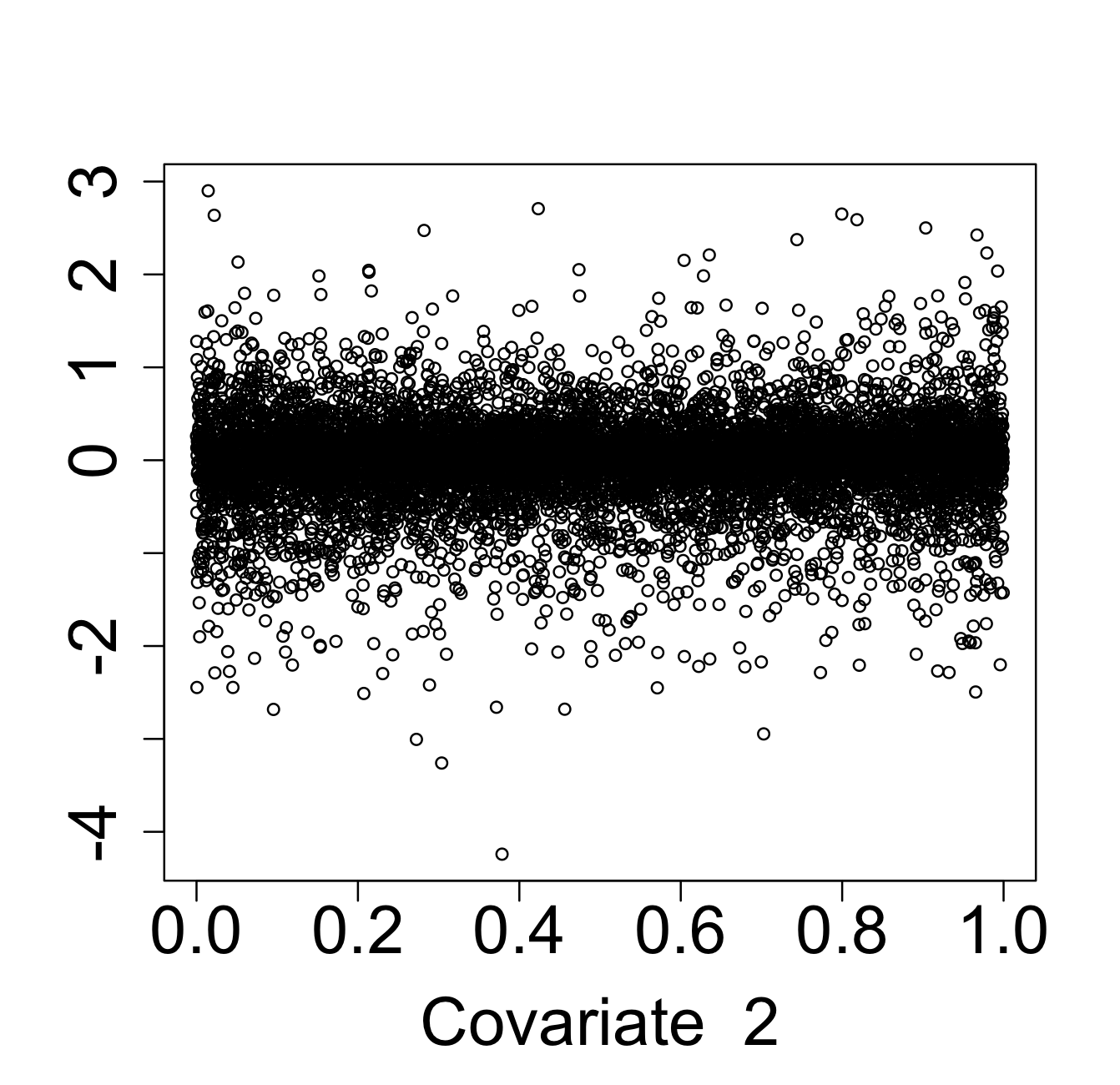}
\end{subfigure}
\begin{subfigure}{0.19\textwidth}
	\includegraphics[width=\textwidth]{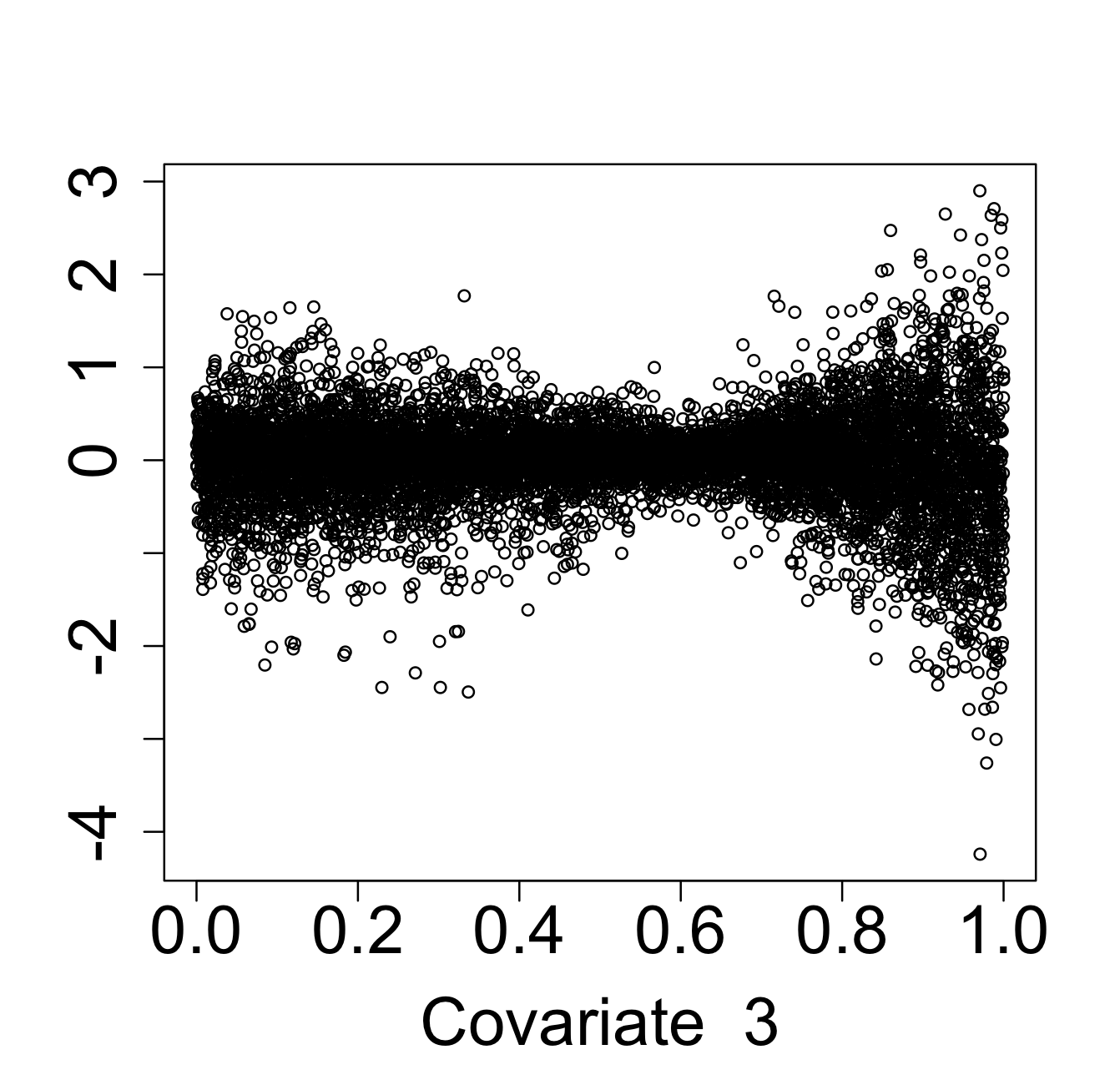}
\end{subfigure}
\begin{subfigure}{0.19\textwidth}
	\includegraphics[width=\textwidth]{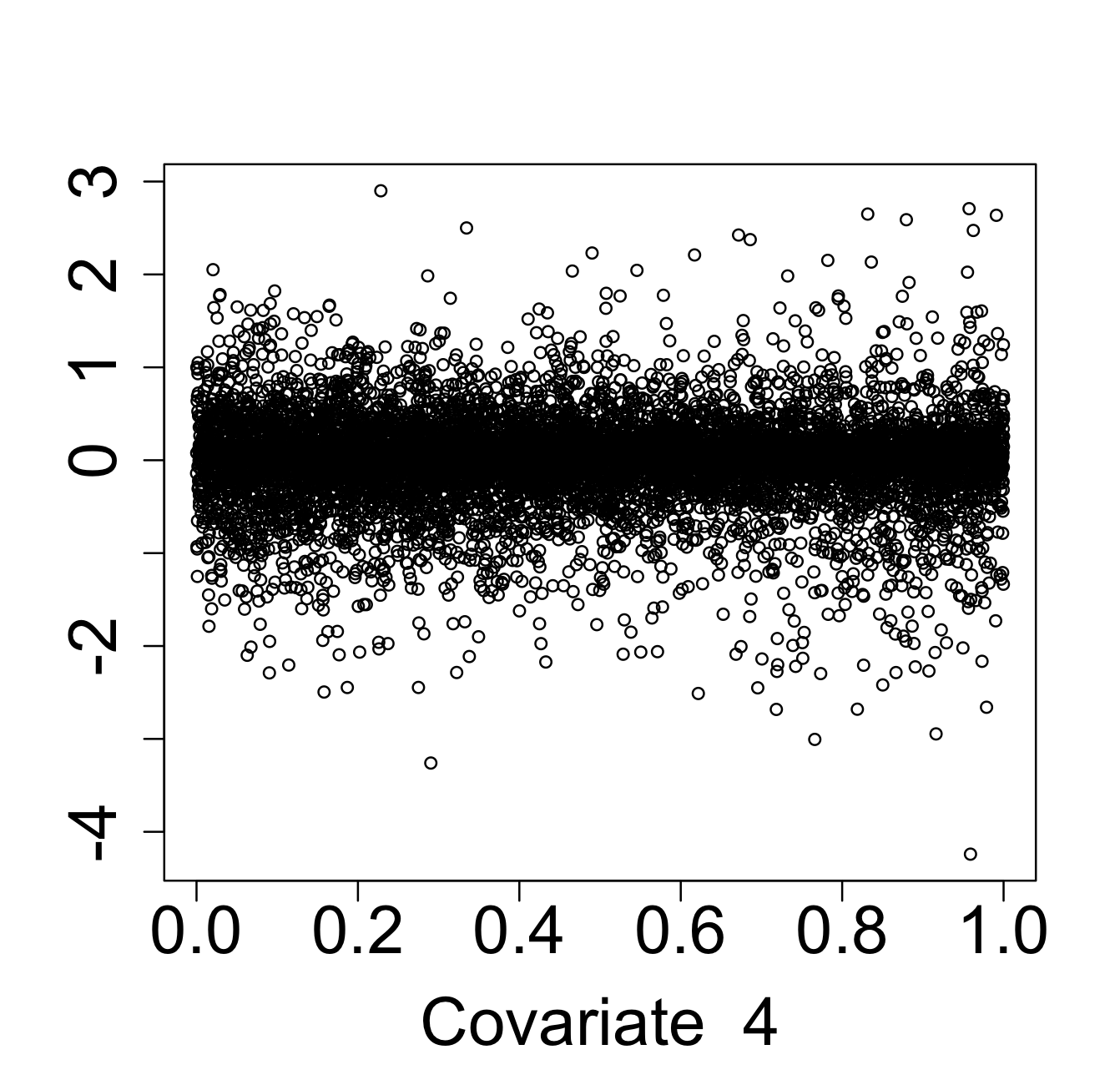}
\end{subfigure}
\begin{subfigure}{0.19\textwidth}
	\includegraphics[width=\textwidth]{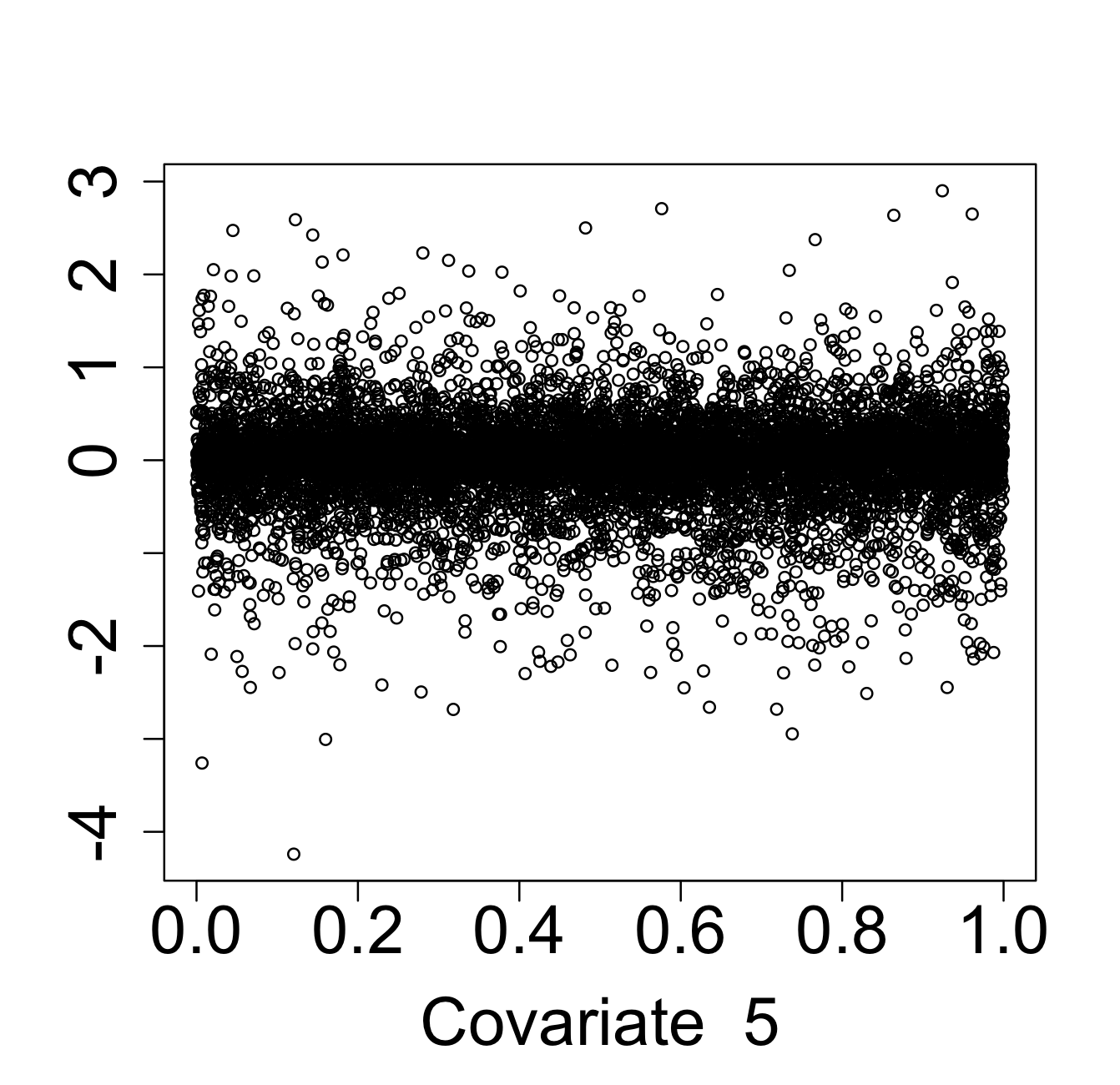}
\end{subfigure}
\begin{subfigure}{0.19\textwidth}
	\includegraphics[width=\textwidth]{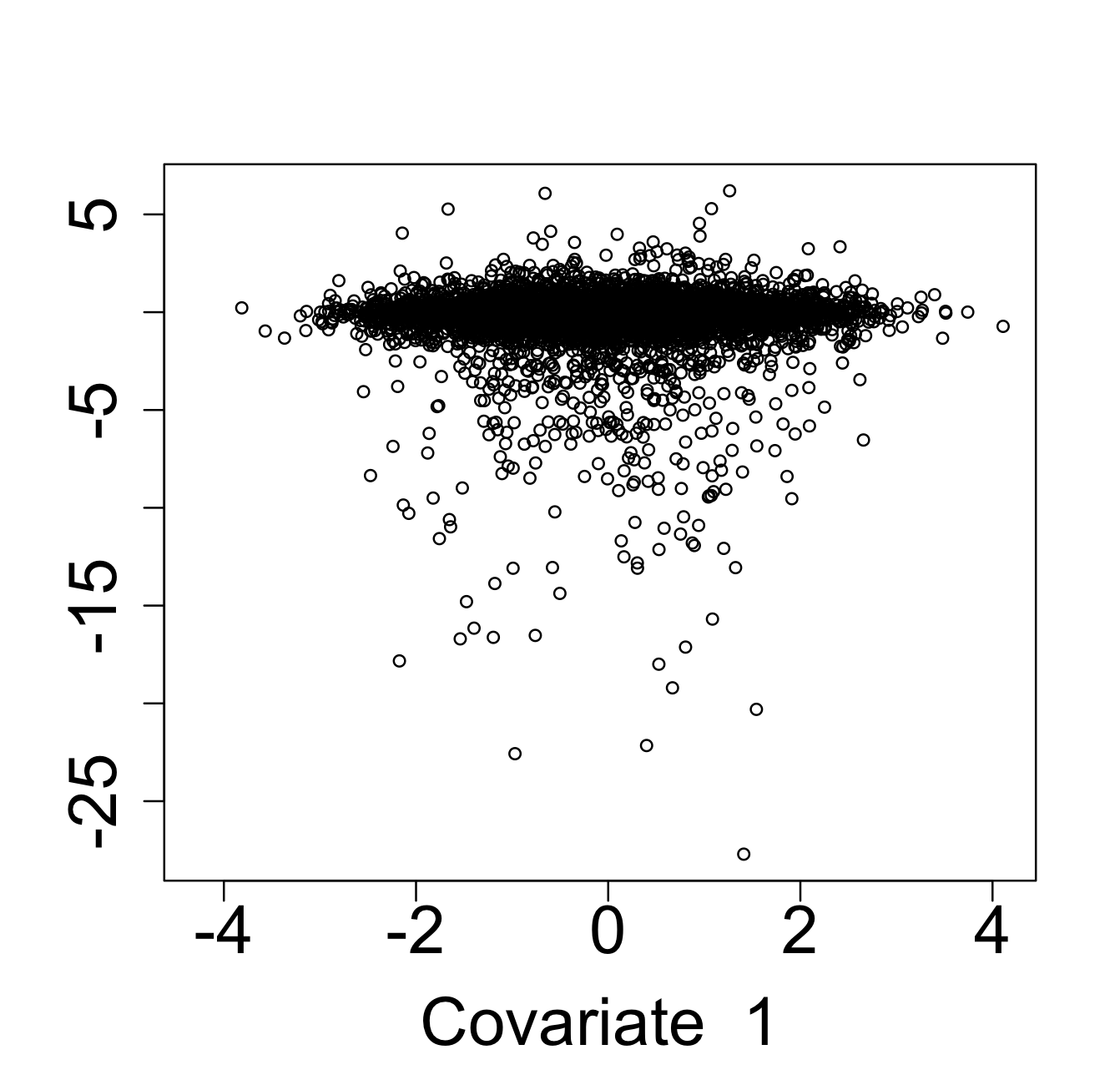}
\end{subfigure}
\begin{subfigure}{0.19\textwidth}
	\includegraphics[width=\textwidth]{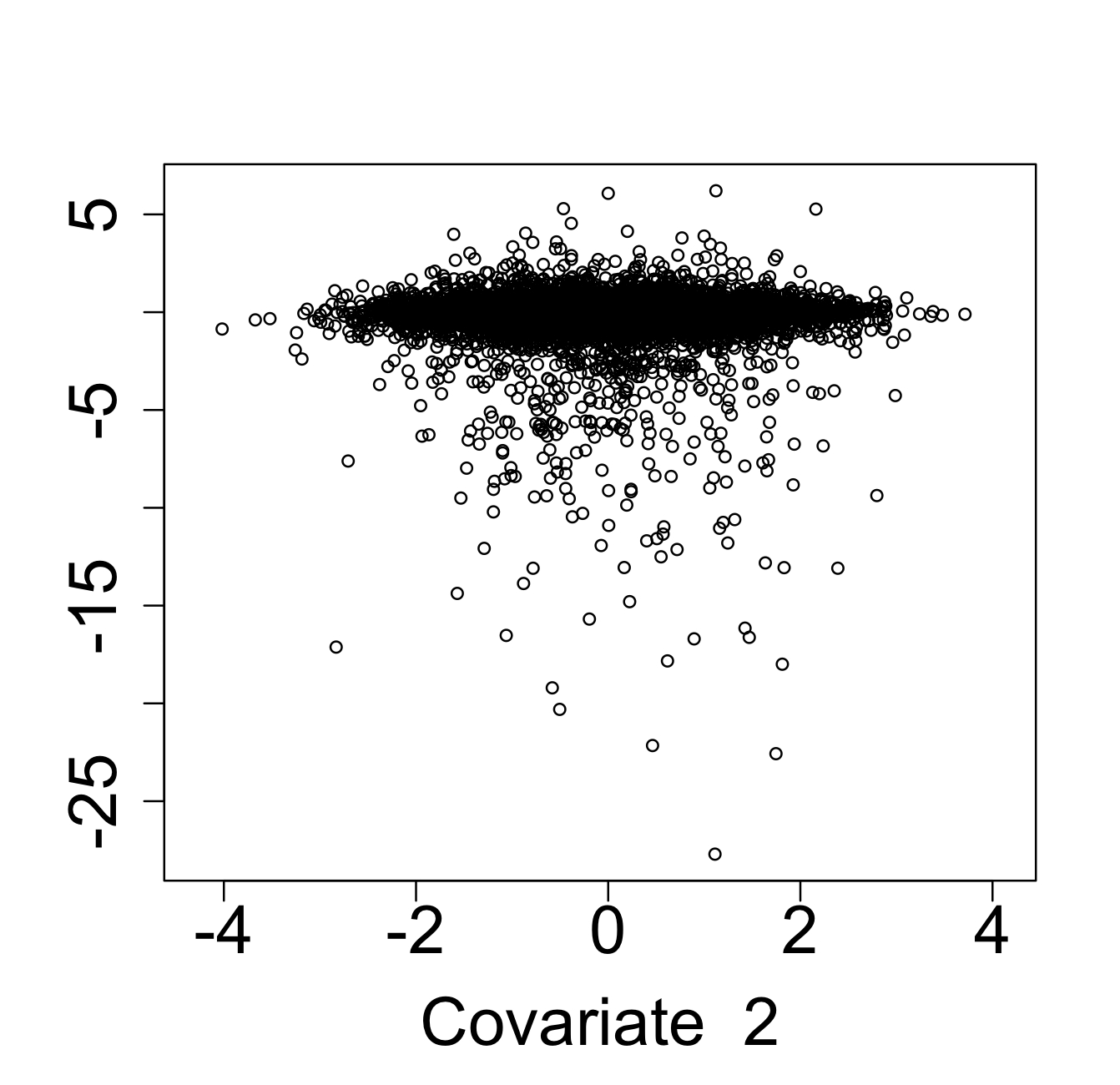}
\end{subfigure}
\begin{subfigure}{0.18\textwidth}
	\includegraphics[width=\textwidth]{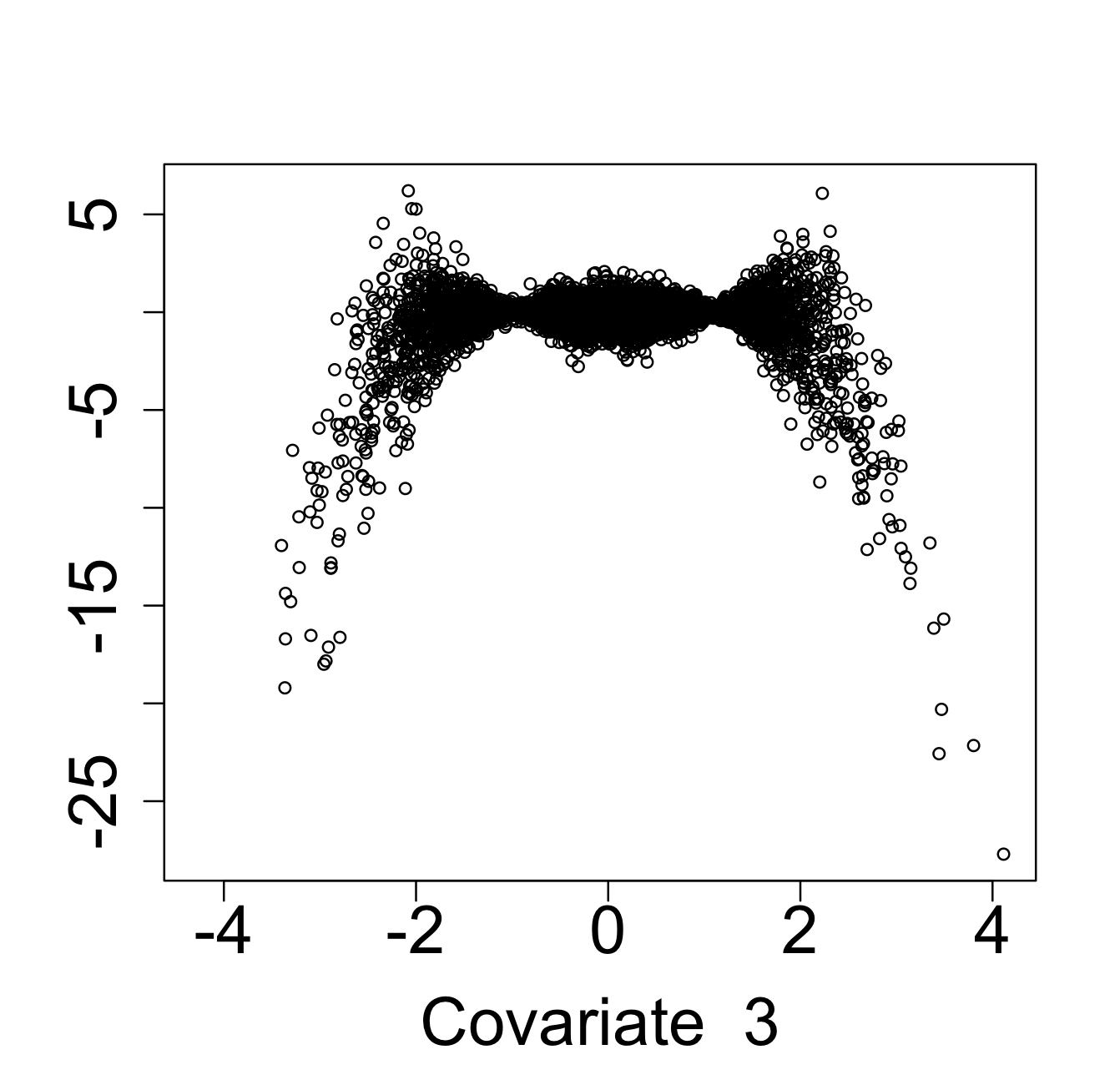}
\end{subfigure}
\begin{subfigure}{0.19\textwidth}
	\includegraphics[width=\textwidth]{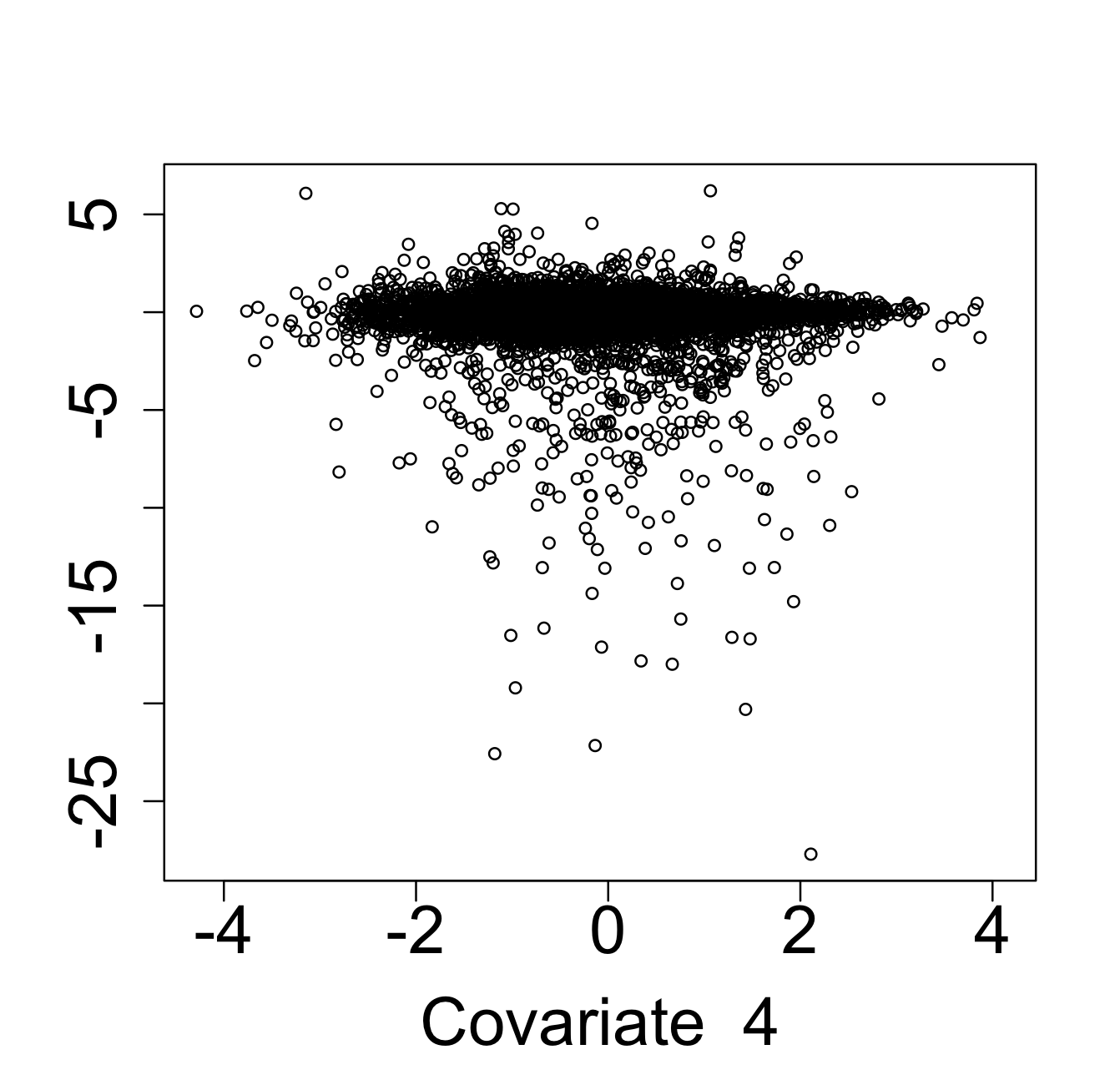}
\end{subfigure}
\begin{subfigure}{0.19\textwidth}
	\includegraphics[width=\textwidth]{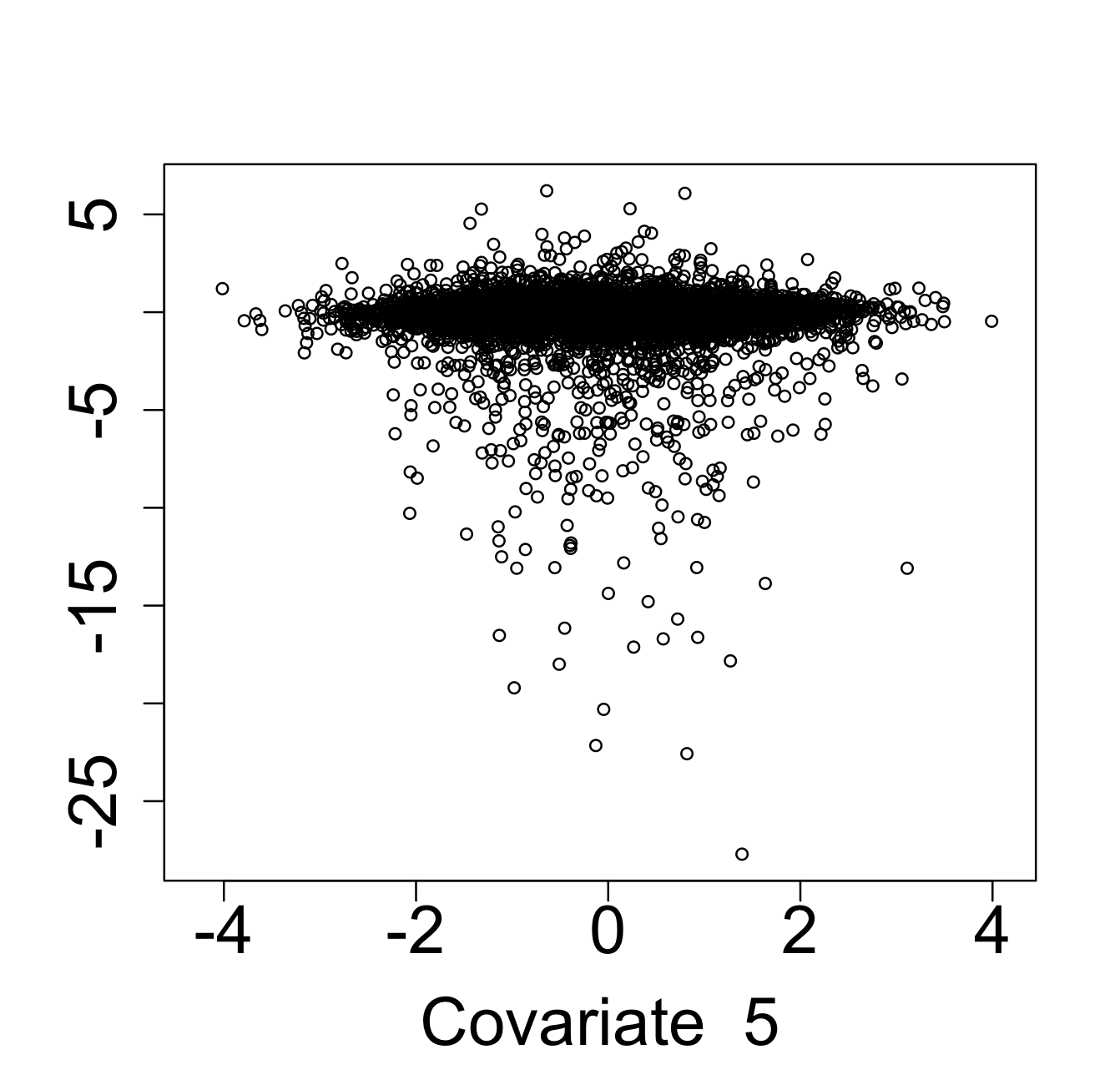}
\end{subfigure}
\caption{ Difference MSE(bagging) $-$ MSE(forest), plotted by covariate $x_j$, for $\mathcal{U}$-MARS (top row) and $\mathcal{N}$-MARS (bottom). In the case of $\mathcal{N}$-MARS, forest performance deteriorates in the tails of $X_3$.}
\label{plots_given_X_MARS}
\end{figure}

In order to understand where the observed deterioration of forest, relative to bagging, comes from, we look at the conditional MSE for different $x$-values in the test-set used for evaluation. We do this for the MARS model, in which the effect is largest. We find that, randomization can increase MSE and this is due to an increase in squared bias for tail values of the third covariate. Each plot in Figure \ref{plots_given_X_MARS} shows, on the vertical axis, the difference $MSE$(bagging) - $MSE$(forest), conditionally on $x$, plotted by covariate $x_j$ on the horizontal axis, for uniform (top row) and normally distributed (bottom row) covariates. In the case $\mathcal{U}(0,1)$, the difference in MSE does not change with $x$, except slightly along the third covariate. In contrast, in the case $\mathcal{N}(0,1)$, we observe that forest does drastically worse, compared to bagging, in the tails of the third covariate. 

\begin{figure}[htbp]
\begin{subfigure}{0.28\textwidth}
	\includegraphics[width=\textwidth]{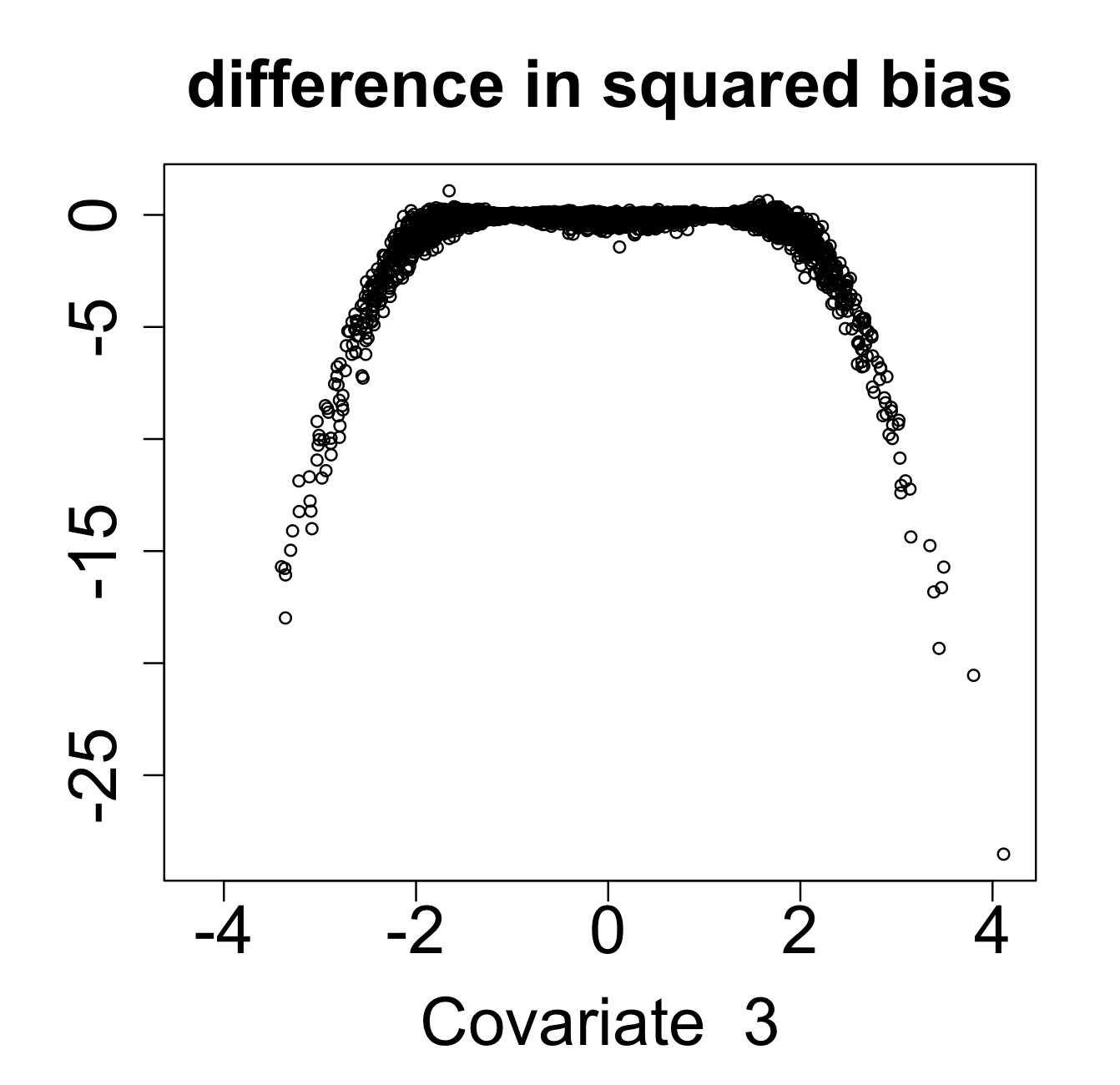}
\end{subfigure}
\begin{subfigure}{0.28\textwidth}
	\includegraphics[width=\textwidth]{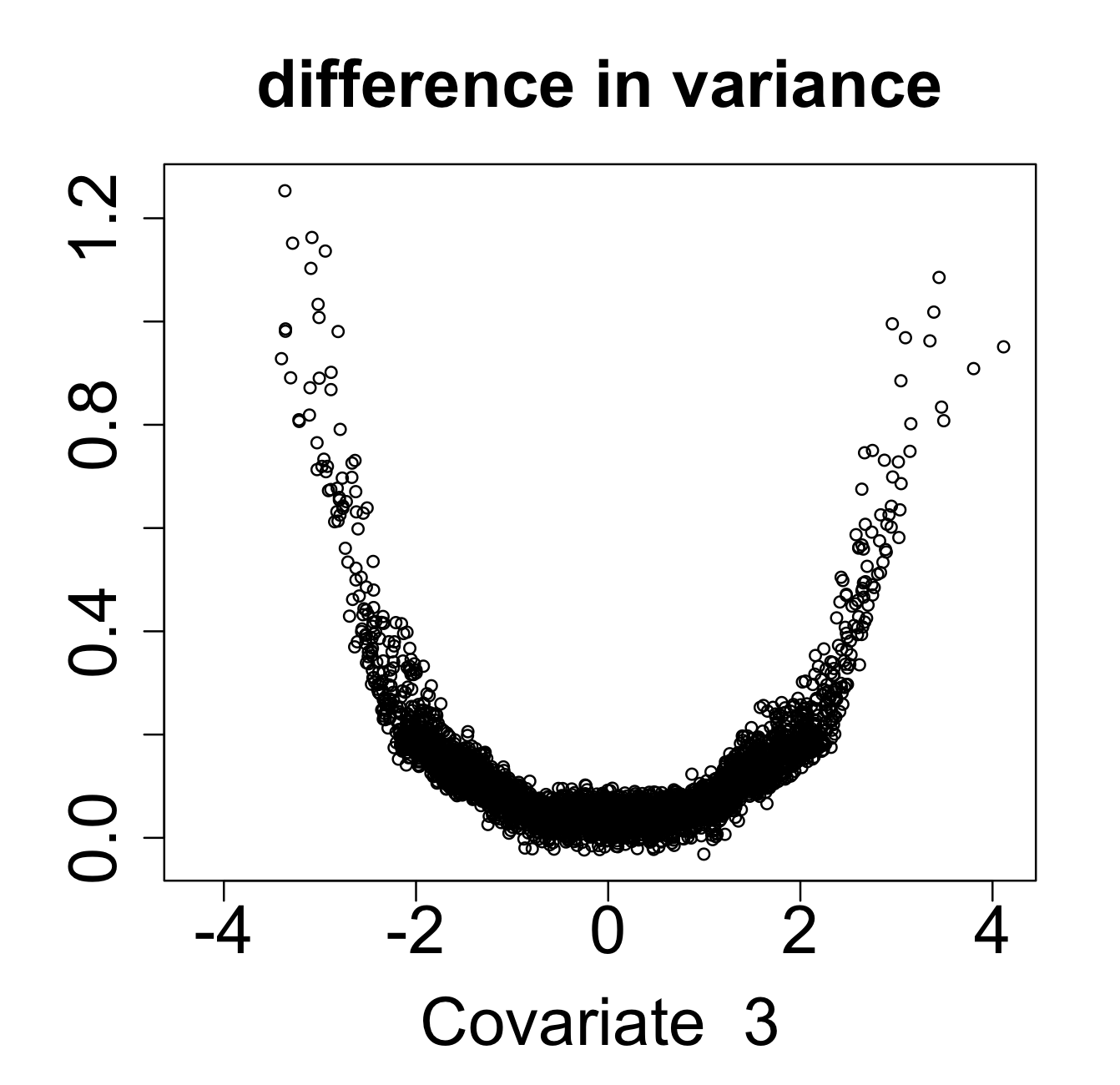}
\end{subfigure}
\caption{ Difference in squared bias and variance between bagging and forests as a function of the third covariate for $\mathcal{N}$-MARS. Negative differences correspond to bagging outperforming forests: randomization increases variance in the presence of tails.}
\label{plots_given_X_N_MARS_bias_and_variance}
\end{figure}

Figure \ref{plots_given_X_N_MARS_bias_and_variance} shows the differences $Bias^2$(bagging) - $Bias^2$(forest) and $Variance$(bagging) - $Variance$(forest), conditionally on $x$, for the third covariate, in the case $\mathcal{N}(0,1)$. We find that, in the tails, randomization increases squared bias, and, while it also decreases variance, it does so at smaller scale. We argue that, by construction, the additional randomness of forests implies that, at each split, some directions are excluded. If we don't split enough in some directions, then regions might be large, in Euclidian volume, while still having only few observations. This implies that bias cannot be further reduced in those regions\footnote{ For a formal relation between prediction bias and cell size see, e.g., Section 3 in \citet{revelas2024does}.}. 

\subsection{Irrelevant covariates}
\label{subsection_irrelevant}
Next we look at the relative difference in MSE between bagging and forests in the presence of irrelevant covariates, namely, letting $X=(X_{rel}, X_{irrel})$, covariates satisfying 
\begin{equation}
\mathrm{E}[Y|X_{rel}, X_{irrel}] = \mathrm{E}[Y|X_{rel}].
\label{definition_irrelevant}
\end{equation}
Here, we assume that all covariates, i.e., both relevant and irrelevant, are iid. 

Table \ref{extension_table_irrelevant_covariates} shows changes in MSE when adding irrelevant covariates to the three basis models $\mathcal{N}$-LINEAR, $\mathcal{U}$-MARS and $\mathcal{U}$-HIDDEN\footnote{ Note that the distribution of covariates is not relevant here, and we kept the distributions of the original models (\ref{model_N_linear}), (\ref{model_U_mars}) and (\ref{model_U_hidden}) as presented in the literature. Note also that, for each model, we started with adding one irrelevant covariate rather than no irrelevant covariate, due to $m_{try}=\frac{p}{3}$: for the LINEAR and MARS models, no irrelevant covariate means forest draws one out of five at each split, while when adding one irrelevant covariate forest draws two (out of six) covariates at each split. This can change performance of forest (and hence performance relative to bagging). In other words we want to fix the total number of covariates to a multiple of three, which allows us to obtain the observed monotonicity in changes. Similarly, for HIDDEN, we start from 3 covariates.}. 
\begin{table}[htbp]
\small
\centering
\begin{tabular}{ccccccc}
  & \multicolumn{2}{c}{$\mathcal{N}$-LINEAR} & \multicolumn{2}{c}{$\mathcal{U}$-MARS} & \multicolumn{2}{c}{$\mathcal{U}$-HIDDEN} \\
  \hline
  Relevant covariates & 5 & 5 & 5 & 5 & 2 & 2 \\
  Irrelevant covariates & 1 & 25 & 1 & 25 & 1 & 13 \\
  Total covariates ($p$) & 6 & 30 & 6 & 30 & 3 & 15 \\
  \hline
  $Bias^2$ bagging & 0.09 & 0.24 & 0.06 & 0.16 & 0.19 & 0.32 \\ 
  $Bias^2$ forest & 0.12 & 0.32 & 0.09 & 0.22 & 0.18 & 0.34 \\ 
  Variance bagging & 0.15 & 0.13 & 0.16 & 0.12 & 0.21 & 0.14 \\ 
  Variance forest & 0.10 & 0.08 & 0.10 & 0.07 & 0.13 & 0.09 \\ 
  Tree variance bagging & 1.36 & 1.66 & 1.24 & 1.47 & 1.27 & 1.48 \\ 
  Tree variance forest & 1.39 & 1.74 & 1.33 & 1.62 & 1.27 & 1.62 \\ 
  Correlation bagging & 0.11 & 0.07 & 0.13 & 0.08 & 0.16 & 0.09 \\ 
  Correlation forest & 0.07 & 0.04 & 0.07 & 0.04 & 0.11 & 0.05 \\ 
  Irreducible & 1.01 & 0.98 & 0.98 & 1.01 & 0.99 & 1.00 \\
  MSE bagging & 1.26 & 1.35 & 1.18 & 1.27 & 1.39 & 1.45 \\ 
  MSE forest & 1.25 & 1.37 & 1.15 & 1.29 & 1.31 & 1.41 \\ 
  test statistic & 12.75 & -15.75 & 24.33 & -10.90 & 38.02 & 19.92 \\ 
  \hline
  relative difference (\%)  & \textbf{0.93} & \textbf{-1.45} & \textbf{2.55} & \textbf{-1.15} & \textbf{6.17} & \textbf{2.20} \\ 
   \hline
\end{tabular}
\caption{Effect of irrelevant covariates. Bagging outperforms forests in the presence of irrelevant covariates. The terms $Bias^2$, Variance and MSE are defined as in (\ref{precise_decomposition}). Tree variance (respectively Correlation) are measures for individual trees (respectively pairs of trees) in the ensemble. From Proposition \ref{proposition_decorrelation}, Variance $\approx$ Tree variance $\times$ Correlation. A positive relative difference, defined in (\ref{relative_performance}), means that the forest outperforms bagging.} 
\label{extension_table_irrelevant_covariates}
\end{table}
First, we find that, when we add irrelevant covariates, the MSE increases for \textit{both} bagging and forests. For both methods, squared bias increases and variance decreases, but bias increases by more. We argue that not only forest, but also bagging, sometimes splits at irrelevant covariates and this is due to the presence of noise: the CART criterion does not always distinguish between good (i.e., at relevant covariates) and bad (i.e., at irrelevant covariates) splits. Second, we find that randomization increases bias, while reducing variance. Bias is increased because excluding possible split directions increases the probability of splitting at irrelevant covariates. Variance is decreased due to decorrelation. In balance, \textit{because} the bias becomes more important, relative to the variance, in the presence of irrelevant covariates, either bagging outperforms forest (this is the case for the $\mathcal{N}$-LINEAR and $\mathcal{U}$-MARS models here) or the improvement due to randomization becomes smaller (this is the case for $\mathcal{U}$-HIDDEN). 

\begin{figure}[htbp]
\begin{subfigure}{0.32\textwidth}
	\includegraphics[width=\textwidth]{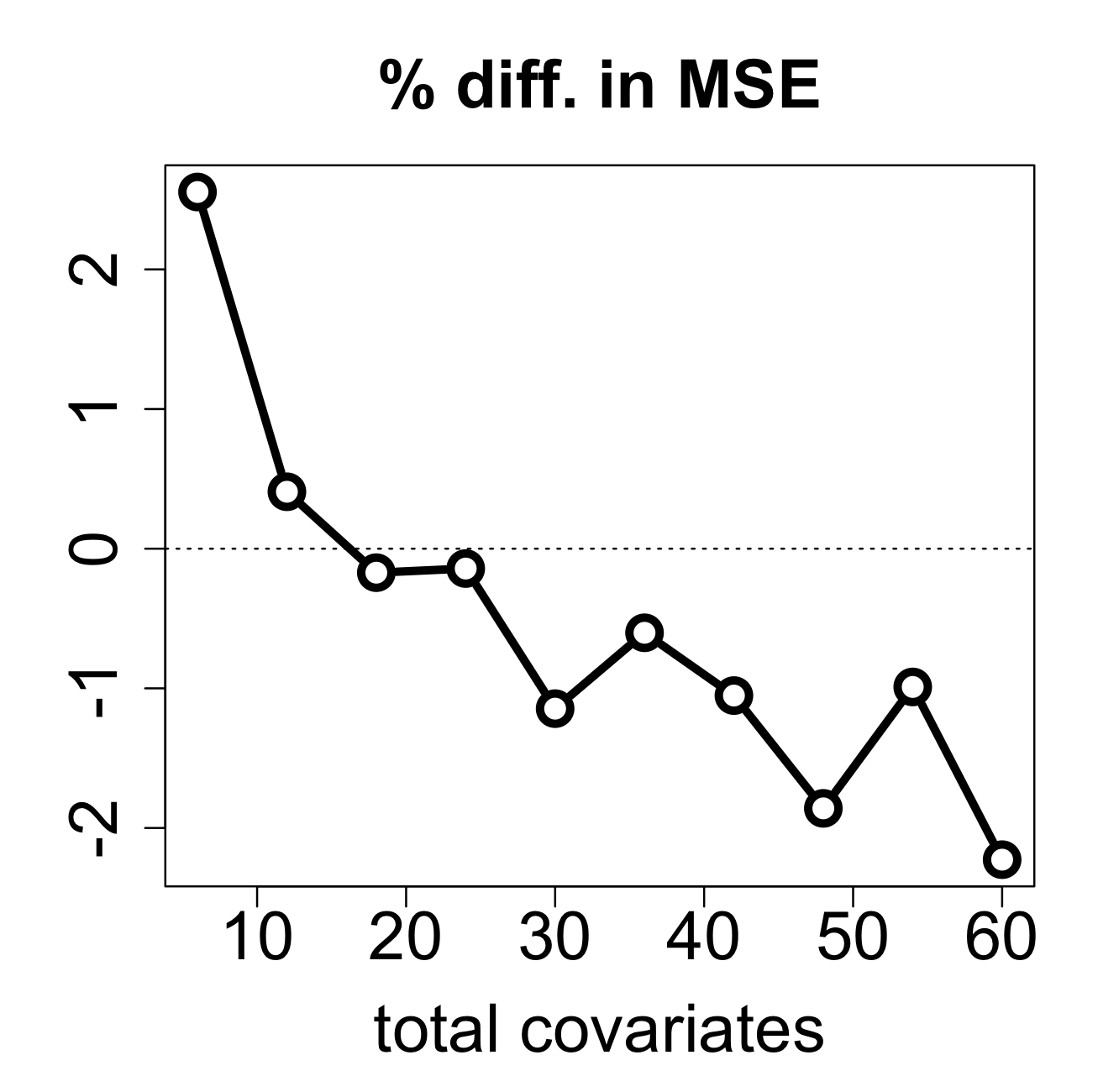}
\end{subfigure}
\begin{subfigure}{0.32\textwidth}
	\includegraphics[width=\textwidth]{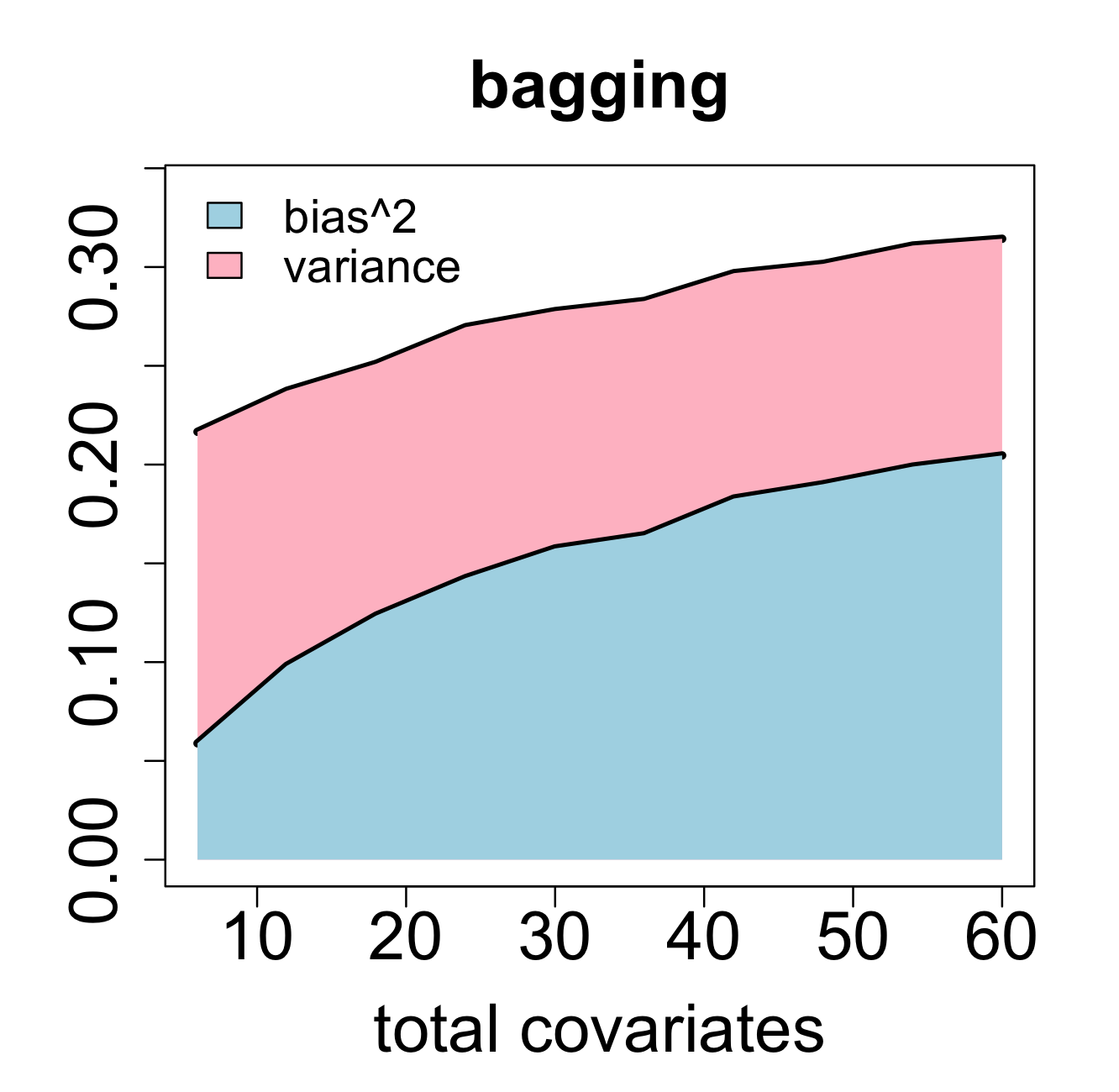}
\end{subfigure}
\begin{subfigure}{0.32\textwidth}
	\includegraphics[width=\textwidth]{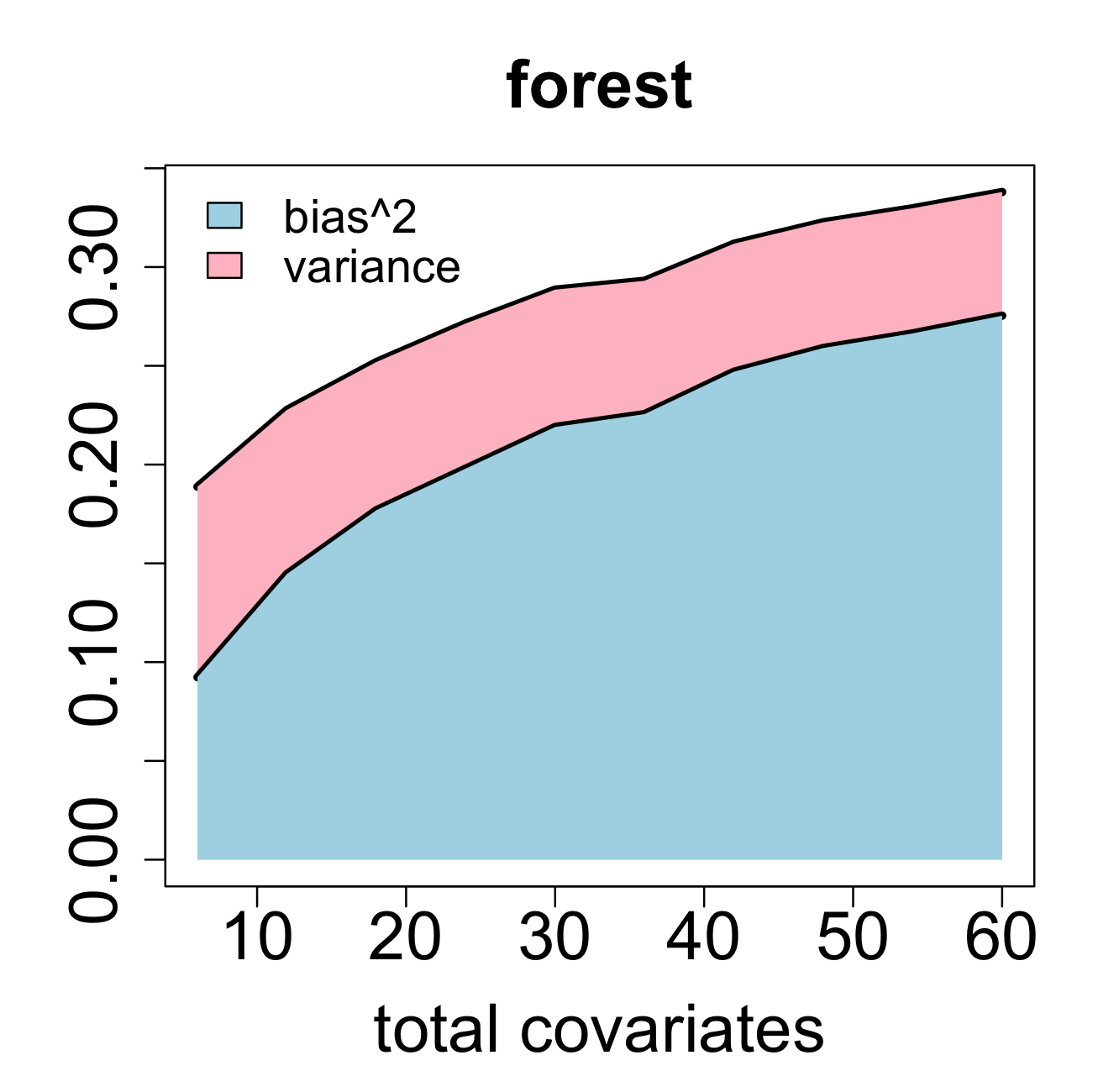}
\end{subfigure}
\caption{ Effect of irrelevant covariates for $\mathcal{U}$-MARS. Left: relative difference in MSE, as in (\ref{relative_performance}). Middle: MSE decomposition (\ref{precise_decomposition}) for bagging as a function of the total number of covariates. Right: MSE decomposition as a function of the total number of covariates for forests. There are five relevant and (1, 7, 13, \dots, 49, 55) irrelevant covariates. Plots are interpolated between these 10 values. The more irrelevant covariates, the more the performance of bagging improves relative to forests.}
\label{plots_irrelevant_covariates_U_MARS}
\end{figure}

Figure \ref{plots_irrelevant_covariates_U_MARS} shows performance as function of the number of irrelevant covariates in the case $\mathcal{U}$-MARS: the left plot shows the relative difference in MSE as defined in (\ref{relative_performance}); the middle and right plots show for bagging and forest respectively the unconditional squared bias and variance terms as in decomposition (\ref{precise_decomposition}). On the horizontal axis is the total number of covariates: there are five relevant covariates, and the number of irrelevant covariates varies from 1 to 25. For both methods, when irrelevant covariates are added, bias increasingly dominates variance. Moreover, bias increases by more for forest than for bagging: randomization is ineffective in the presence of many irrelevant covariates. We observe the same patterns for $\mathcal{N}$-LINEAR and $\mathcal{U}$-HIDDEN, which we added to Appendix \ref{appendix_irrelevant_covariates_plots}. 

\subsection{Correlated Covariates}
\label{subsection_dependencies}
We now look at the performance of bagging and forests when covariates are mutually correlated. We consider again the three basis DGPs and use normally distributed covariates in order to add a dependence structure. Precisely, we take a multivariate normal $X\sim\mathcal{N}(\textbf{0},\Sigma_{\rho})$ where $\Sigma_{\rho}$ is a matrix containing 1's in the diagonal and each off-diagonal element equals some common value $\rho\in[0,1]$. In other words, we assume for simplicity constant pairwise correlation among covariates. 

Table \ref{extension_table_dependencies} shows changes in the relative difference in out-of-sample MSE when passing from no correlation between covariates ($\rho=0$) to mutually correlated covariates with pairwise correlation $\rho=0.5$. For the $\mathcal{N}$-LINEAR model, forest and bagging have similar MSE when $\rho=0$ and forest outperforms bagging when $\rho=0.5$. For the $\mathcal{N}$-MARS model, bagging outperforms forest for $\rho=0$ and does so by less for $\rho=0.5$. For $\mathcal{N}$-HIDDEN, the change in relative difference from $\rho=0$ to $\rho=0.5$ is small; we expect this to be due to the small number of covariates in this example, and we exclude it from the analysis hereafter. First, we find that, for both bagging and forests, the MSE decreases in the presence of correlations and this decrease is mostly due to a reduction in squared bias (e.g., in $\mathcal{N}$-LINEAR, for bagging, bias decreases by $0.08-0.01=0.07$ and variance decreases by $0.15-0.14=0.01$). Second, we find that the reduction in MSE is larger for forests than for bagging (e.g., in $\mathcal{N}$-LINEAR, reduction of $1.23-1.09=0.14$ for forests versus $1.23-1.16=0.07$ for bagging). Randomization increases bias independently of whether covariates are correlated or not. Given that randomization is ineffective when its increase in bias dominates its decrease in variance, and because bias decreases with correlation, this explains why randomization can be effective in reducing MSE in the presence of correlated covariates.

\begin{table}[htbp]
\small
\centering
\begin{tabular}{ccccccc}
  & \multicolumn{2}{c}{$\mathcal{N}$-LINEAR} & \multicolumn{2}{c}{$\mathcal{N}$-MARS} & \multicolumn{2}{c}{$\mathcal{N}$-HIDDEN} \\
  \hline
  correlation $\rho$ & 0 & 0.5 & 0 & 0.5 & 0 & 0.5 \\
  \hline
  $Bias^2$ bagging & 0.08 & 0.01 & 0.10 & 0.09 & 0.01 & 0.01 \\ 
  $Bias^2$ forest & 0.15 & 0.02 & 0.36 & 0.28 & 0.02 & 0.02 \\ 
  Variance bagging & 0.15 & 0.14 & 0.18 & 0.18 & 0.22 & 0.22 \\ 
  Variance forest & 0.08 & 0.07 & 0.10 & 0.10 & 0.17 & 0.17 \\ 
  Tree variance bagging & 1.29 & 1.09 & 1.19 & 1.17 & 0.92 & 0.93 \\ 
  Tree variance forest & 1.34 & 1.03 & 1.49 & 1.38 & 0.97 & 0.93 \\ 
  Correlation bagging & 0.12 & 0.13 & 0.13 & 0.14 & 0.23 & 0.24 \\ 
  Correlation forest & 0.05 & 0.06 & 0.07 & 0.07 & 0.18 & 0.18 \\ 
  Irreducible & 1.01 & 1.01 & 1.01 & 1.01 & 1.00 & 1.00 \\ 
  MSE bagging & 1.23 & 1.16 & 1.29 & 1.28 & 1.23 & 1.23 \\ 
  MSE forest & 1.23 & 1.09 & 1.48 & 1.39 & 1.19 & 1.19 \\ 
  test statistic & 3.80 & 64.80 & -98.72 & -60.64 & 46.18 & 48.92 \\ 
  \hline
  relative difference (\%)  & \textbf{0.37} & \textbf{5.58} & \textbf{-12.37} & \textbf{-7.73} & \textbf{3.21} & \textbf{3.47} \\ 
   \hline
\end{tabular}
\caption{Effect of correlated covariates. For both bagging and forests, bias is reduced in the presence of correlated covariates. The terms $Bias^2$, Variance and MSE are defined as in (\ref{precise_decomposition}). Tree variance (respectively Correlation) are measures for individual trees (respectively pairs of trees) in the ensemble. From Proposition \ref{proposition_decorrelation}, Variance $\approx$ Tree variance $\times$ Correlation. A positive relative difference, defined in (\ref{relative_performance}), means that the forest outperforms bagging.} 
\label{extension_table_dependencies}
\end{table}

\begin{figure}[htbp]
\begin{subfigure}{0.32\textwidth}
	\includegraphics[width=\textwidth]{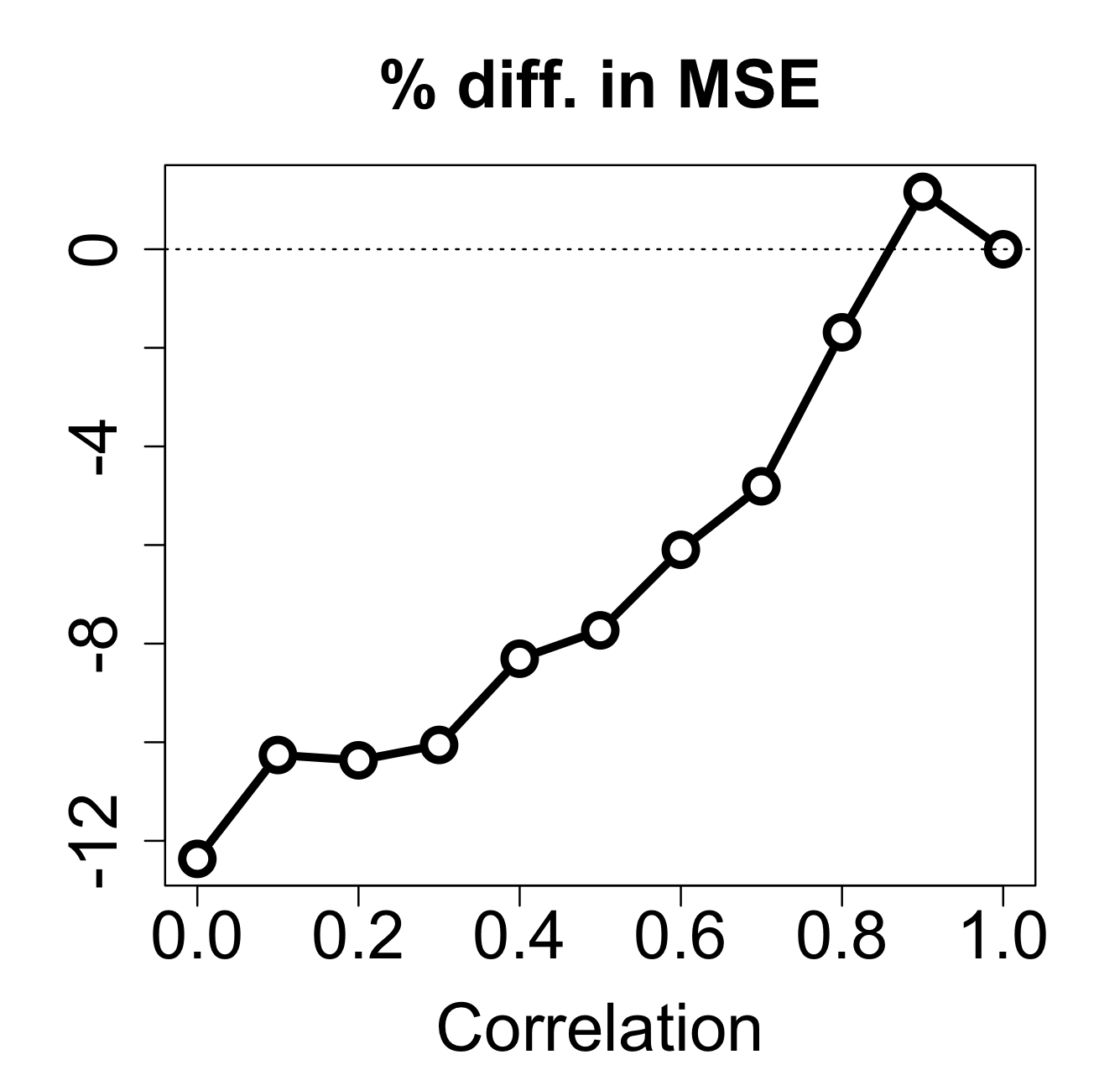}
\end{subfigure}
\begin{subfigure}{0.32\textwidth}
	\includegraphics[width=\textwidth]{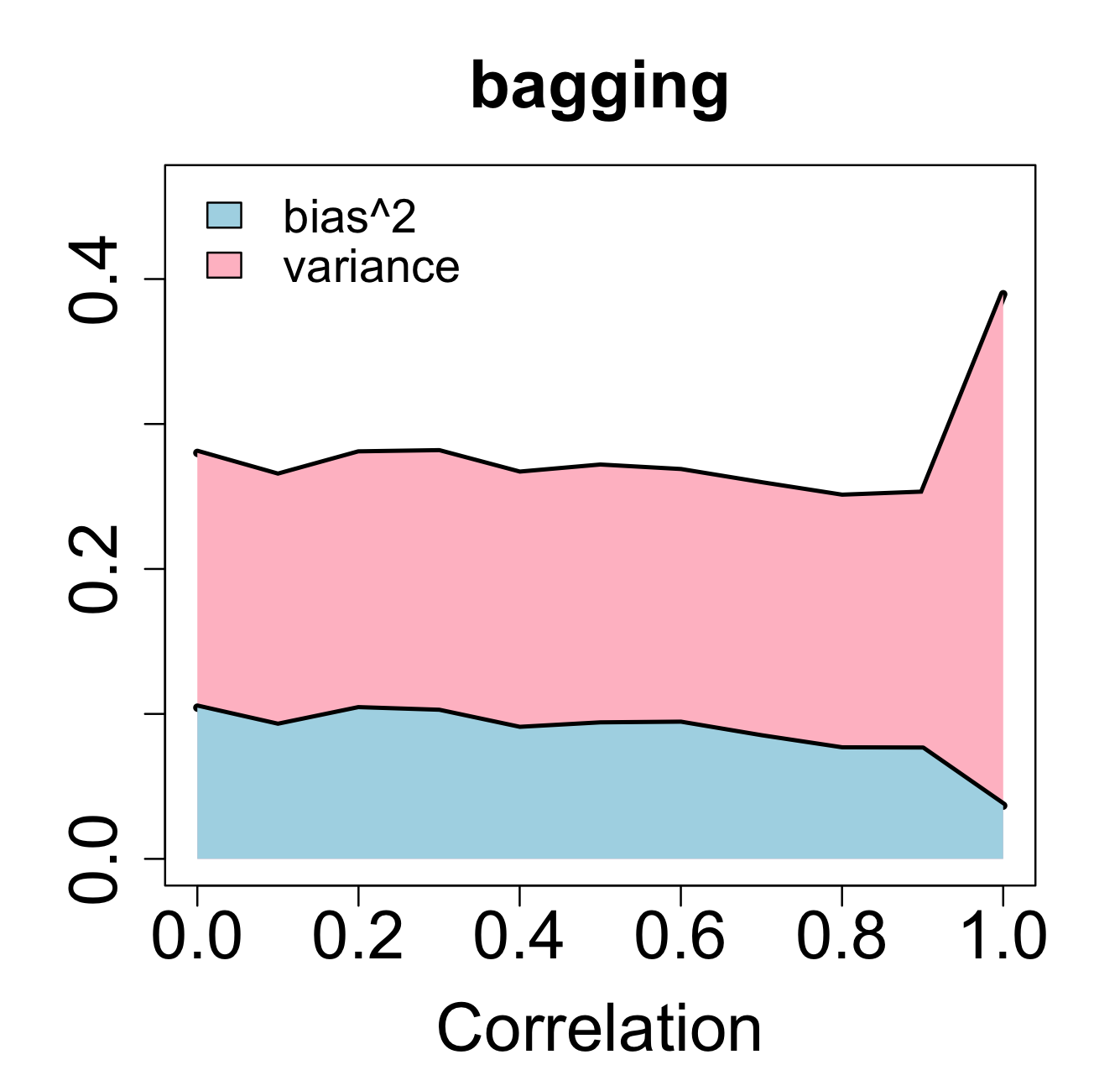}
\end{subfigure}
\begin{subfigure}{0.32\textwidth}
	\includegraphics[width=\textwidth]{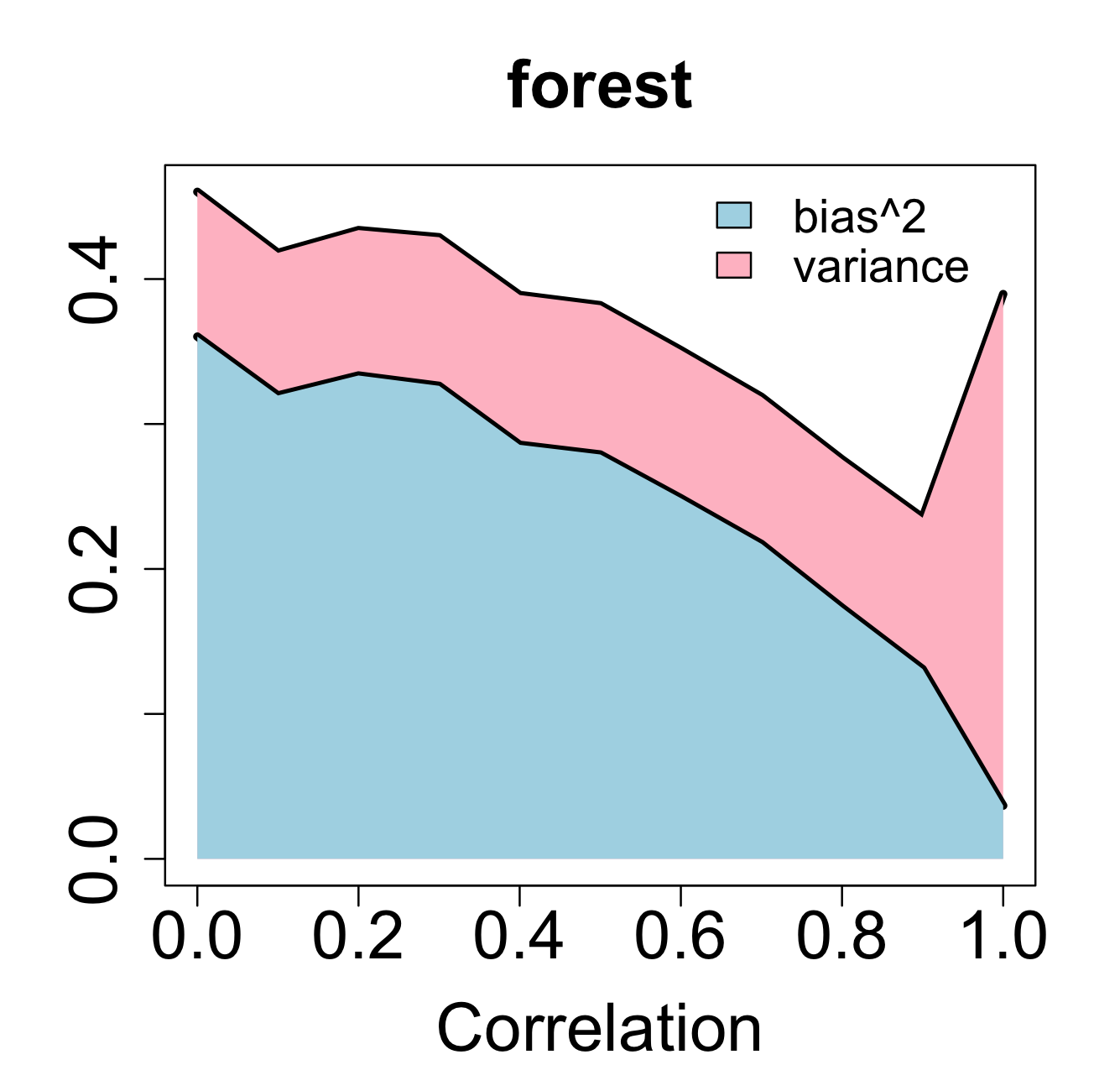}
\end{subfigure}
\caption{ Effect of correlated covariates for $\mathcal{N}$-MARS. Left: relative difference in MSE, as in (\ref{relative_performance}). Middle: MSE decomposition (\ref{precise_decomposition}) as a function of $\rho$ for bagging. Right: MSE decomposition as a function of $\rho$ for forests. Plots are interpolated between the 10 values of $\rho$: in the presence of perfect correlation, the two methods always have the same MSE.}
\label{plots_dependencies_N_MARS}
\end{figure}

Figure \ref{plots_dependencies_N_MARS} shows the performance of bagging and forest for different values of $\rho\in[0,1]$ in the $\mathcal{N}$-MARS example. The left plot shows the relative difference in MSE as in (\ref{relative_performance}). The middle and right plots show the squared bias and variance, as in (\ref{precise_decomposition}), for each method respectively. On the horizontal axis is the correlation $\rho$. We find that randomization improves monotonically with $\rho$. For large enough correlation, forests even outperform bagging in this example. Again, improvement comes from bias: for both methods variance remains roughly constant across correlations; squared bias slightly decreases with $\rho$ for bagging and significantly decreases with $\rho$ for the forest. We argue that, while in the case $\rho=0$ randomization increases bias because some ``good splits" are avoided, in the presence of correlated covariates, the loss is less harmful: not splitting on a ``good" covariate is less harmful when this covariate is correlated with the other covariates, and, the stronger the correlation, the smaller the harm due to randomization. Why increasing $\rho$ can result in forests even outperforming bagging, as happens for $\rho=0.9$ in this example, is something we cannot explain. Yet, we observe this even more in the case $\mathcal{N}$-LINEAR (shown in Table \ref{extension_table_dependencies} above, as well as in Figure \ref{plots_dependencies_N_LINEAR} in Appendix \ref{appendix_dependencies_plots}). Finally, we find that, in the presence of perfect correlation ($\rho=1$), bagging and forest perform exactly the same (i.e., their relative difference is zero). This is something we observe for every model that we tried. Following the same reasoning as for the previous observation, we argue that, with perfect correlation, it does not matter whether or not we randomize the directions in which splits are allowed: both methods essentially split on one and only direction all the time.

\subsection{Correlated and irrelevant covariates}
We extend our findings of the previous sections by looking at the joint presence of correlated and irrelevant covariates. We assume that all covariates, i.e., both relevant and irrelevant ones, are mutually correlated with constant pairwise correlation $\rho$. Table \ref{table_joint_irrelevant_correlated} shows the relative difference in MSE between bagging and forests for the three basis models. 
\begin{table}[htbp]
\small
\centering
\begin{tabular}{ccccc}
  & irrelevant (total) &  $\rho=0$ & $\rho=0.5$ & $\rho=0.9$ \\
  \hline
  \multirow{3}{*}{$\mathcal{N}$-LINEAR} & 1 ($p=6$) & 0.93 & 3.35 & 3.66 \\ 
  & 25 ($p=30$) & -1.45 & 1.34 & 1.96 \\ 
  & 55 ($p=60$) & -1.94 & 0.79 & 1.42 \\ 
  \hline
  \multirow{3}{*}{$\mathcal{N}$-MARS} & 1 ($p=6$) & -5.68 & -3.73 & 1.06 \\ 
  & 25 ($p=30$) & -9.59 & -6.92 & 0.52 \\ 
  & 55 ($p=60$) & -10.58 & -6.72 & 0.26 \\ 
  \hline
  \multirow{3}{*}{$\mathcal{N}$-HIDDEN} & 1 ($p=3$) & 1.93 & 2.90 & 4.47 \\
  & 13 ($p=15$) & -2.66 & -1.61 & 1.61 \\ 
  & 28 ($p=30$) & -4.59 & -3.92 & 1.18 \\ 
  \hline
\end{tabular}
\caption{Relative difference in MSE (\ref{relative_performance}) in the presence of irrelevant and correlated covariates. The more irrelevant covariates, the larger the improvement of forest as $\rho$ increases.}
\label{table_joint_irrelevant_correlated}
\end{table}

We find that, (i) for any $\rho$, the performance of forest deteriorates, relative to bagging, with the number of irrelevant covariates; (ii) for any number of irrelevant covariates, the performance of forest improves, relative to bagging, with $\rho$, and (iii) the amount of improvement due to correlation varies with the number of irrelevant covariates: the more irrelevant covariates, the larger the improvement of forest tends to be as $\rho$ increases.

\subsection{Varying SNR from low to high}
\label{subsection_varying_SNR}
Finally, we extend the findings presented in Sections \ref{subsection_distribution_of_X}, \ref{subsection_irrelevant} and \ref{subsection_dependencies} to low and high SNR scenarios. We ran again each of Tables \ref{extension_table_distribution_of_X}, \ref{extension_table_irrelevant_covariates}, and \ref{extension_table_dependencies}, replacing $\text{SNR}=1$ with $\text{SNR}=0.05$ and $\text{SNR}=6$. Detailed results are given in Appendix \ref{appendix_simulations_SNR}. Here, we summarize what we find. 

In the presence of tails in the distribution of covariates, randomization increases bias in both low and high SNR. However, in low SNR, the increase in bias is negligible compared to the decrease in variance, and forests always outperform bagging. In high SNR, with the exception of the linear model, changes in relative difference when switching from uniform to normal covariates not only are in line with our findings but they are also amplified. 

Adding irrelevant covariates consistently works in favour of bagging across models and SNR, i.e., the relative difference in MSE between bagging and forests decreases. Again, when the SNR is low, randomization reduces MSE in every example, because variance dominates bias for both methods. When the SNR is high, randomization is ineffective, because the increase in bias dominates the decrease in variance. 

In the presence of correlated covariates, with the exception of the hidden pattern model, bias decreases for both methods. When the SNR is low, variance dominates bias and hence randomization is effective in every model. When the SNR is high, for the linear model, for uncorrelated covariates bagging outperforms forest because the increase in bias due to randomization dominates variance, and when covariates are correlated, forests outperform bagging because the variance reduction dominates. In the case of the Mars model, randomization is still ineffective compared to bagging, but less ineffective in the presence of correlated covariates compared to non-correlated covariates.


\section{Practical considerations and future work}\label{section_guidelines}
We end this paper with a discussion on practical considerations and possible future work. MSE and its decompositions cannot be computed in real applications because the true model, or regression function $f$, is unknown. Similarly, the SNR also cannot be computed exactly. What can be computed, is the variation of predictions over training sets. In other words, the variance term in our first MSE decomposition (\ref{precise_decomposition}), where we first average over training sets and then over covariate values. Hence, we recommend preferring this decomposition over the alternative (\ref{precise_decomposition_alternative}). Moreover, decomposition (\ref{precise_decomposition}) allows for an analysis of MSE locally in the covariates, which can be useful in practice if the researcher is interested in evaluating methods at particular regions of the feature space. In practice, one can first estimate $f$ with a preferred of-the-shelf prediction method and use this estimate as true regression model to compute the MSE and its decomposition. However, any interpretation of the obtained results should be done with caution. When it comes to the SNR, the practitioner could either again use an of-the-shelf estimate for $f$ to compute the SNR as defined in (\ref{SNR_definition}), or use the variance of different methods as a proxy for noise: if non-flexible methods exhibit high variance then this could indicate that the environment is noisy. For particular applications the practitioner might also have an a priori idea of the expected amount of noise (if, e.g., the goal is predicting income and only age and weight are used as explanatory variables, then we can be confident that the SNR will be low).

We have shown that forests can perform worse than bagging in the presence of tails or irrelevant covariates, while they seem to be at an advantage in the presence of correlated covariates (even when, at the same time, there are many irrelevant covariates). In practice, tails and correlations can be observed directly in the data. One way to examine the relevance of covariates is by looking at the performance of forests and bagging separately with and without a given covariate, e.g., via cross-validation. Alternatively, one could assume a linear model and apply statistical tests. Our experiments are limited to a small number of covariates. Extending our experiments by, e.g., increasing the sample size together with the number of irrelevant covariates, or varying pairwise correlations between relevant and irrelevant covariates, we believe, could be of practical relevance and are left for future research. We did not conduct an empirical application, which is something we also leave for future research. 
One way of illustrating the advantage of forests in real datasets would be to group covariates by levels of pairwise correlation, start by applying both bagging and forests with only the least correlated covariates, then progressively add covariates that are more and more correlated, and observe the change in relative difference in MSEs.



\section{Conclusion}
\label{section_conclusion}

In this paper, we have brought together and extended prior literature towards a better understanding of when split randomization in random forests is effective and when it is not. We have shown that the amount of noise in the data impacts bias and variance for both bagging and forests, and, while randomization always decorrelates trees, its effectiveness ultimately depends on the balancing dominance of bias and variance in a mean-squared error decomposition. Then, we have shown that, in order to answer the question of when randomization works and when it does not, we need to also look beyond noise. We have shown changes in the relative difference in mean-squared error between the two methods, of at least the same order of magnitude as in prior literature, for a fixed and moderate signal-to-noise ratio. In particular, we found that forests can increase bias in the presence of tails in the distribution of covariates and, while in the presence of irrelevant covariates randomization increases mean-squared error, when covariates are correlated, for both methods, bias is reduced. Hence, in balance with variance, randomization tends to be effective in the presence of correlations. 

We have put forward a normalized framework for comparison of the two methods and have underlined the importance of being explicit in how bias and variance are defined when decomposing out-of-sample mean-squared error to explain differences. In most simulated datasets that we considered, we found that differences often do not exceed the order of 5\%, and therefore, depending on the application at hand, choosing between bagging or forests could be irrelevant. Moreover, when bagging outperforms forests, it does so by more, compared to when forests outperforms bagging, and hence, in practice, when bagging should be preferred and is not, the loss can be larger. 

Our findings on correlated covariates are, to our knowledge, new of their kind and can possibly open the way for a better understanding of why random forests work well in many applications, given that in practice covariates are often correlated. In particular, we find that, independently of randomization, bias is significantly reduced in the presence of correlated covariates. This finding goes beyond the prevailing view that averaging mostly works by variance reduction, and better understanding why this happens is something that we leave for future research.


\newpage
\appendix
\section{Proofs}
\label{appendix_proofs}

\subsection{Proof of Proposition \ref{proposition_decomposition_MSE_starting_given_X}}
Here we prove (\ref{precise_decomposition}). We have
\begin{align}
\mathrm{E}\Big[\Big(Y-\hat{f}(X)\Big)^2\Big|X\Big] = & \ \mathrm{E}\Big[\Big(f(X)+\varepsilon-\hat{f}(X)\Big)^2\Big|X\Big] \\
= & \ \mathrm{E}\Big[\Big(f(X)-\hat{f}(X)\Big)^2\Big|X\Big] + 2\ \mathrm{E}\Big[\varepsilon\Big(f(X)-\hat{f}(X)\Big)\Big|X\Big] + \mathrm{E}[\varepsilon^2|X] \nonumber \\
= & \ \mathrm{E}\Big[\Big(f(X)-\hat{f}(X)\Big)^2\Big|X\Big] +  \sigma_{\varepsilon}^2 \nonumber 
\end{align}
using that $\mathrm{E}[\varepsilon|X]=0$ and the independence of $\hat{f}$ from $X$ and $\varepsilon$. Also 
\begin{align}
\mathrm{E}\Big[\Big(f(X)-\hat{f}(X)\Big)^2\Big|X\Big] = & \ \Big(\mathrm{E}[f(X)-\hat{f}(X)|X]\Big)^2 + \mathrm{Var}\big[f(X)-\hat{f}(X)|X\big]\\
= & \ \Big(f(X)-\mathrm{E}[\hat{f}(X)|X]\Big)^2 + \mathrm{Var}[\hat{f}(X)|X]. \nonumber 
\end{align}
Hence, by the law of iterated expectations, 
\begin{align}
\mathrm{E}\Big[\Big(Y-\hat{f}(X)\Big)^2\Big] 
= & \ \mathrm{E}\Big[\Big(f(X)-\mathrm{E}[\hat{f}(X)|X]\Big)^2\Big] + \mathrm{E}\Big[\mathrm{Var}\big[\hat{f}(X)|X\big]\Big] + \sigma_{\varepsilon}^2. 
\end{align}

\subsection{Alternative decomposition of the unconditional MSE}
Here we prove (\ref{precise_decomposition_alternative}). Instead of conditioning on $X$ first, we condition on $\hat{f}$ first. Again, 
\begin{align}
\mathrm{E}\Big[\Big(Y-\hat{f}(X)\Big)^2\Big|\hat{f}\Big] = & \ \mathrm{E}\Big[\Big(f(X)+\varepsilon-\hat{f}(X)\Big)^2\Big|\hat{f}\Big] \\
= & \ \mathrm{E}\Big[\Big(f(X)-\hat{f}(X)\Big)^2\Big|\hat{f}\Big] + 2\ \mathrm{E}\Big[\varepsilon\Big(f(X)-\hat{f}(X)\Big)\Big|\hat{f}\Big] + \mathrm{E}[\varepsilon^2|\hat{f}] \nonumber \\
= & \ \mathrm{E}\Big[\Big(f(X)-\hat{f}(X)\Big)^2\Big|\hat{f}\Big] +  \sigma_{\varepsilon}^2 .\nonumber 
\end{align}
Moreover, 
\begin{align}
\mathrm{E}\Big[\Big(f(X)-\hat{f}(X)\Big)^2\Big|\hat{f}\Big] = & \ \Big(\mathrm{E}[f(X)-\hat{f}(X)|\hat{f}]\Big)^2 + \mathrm{Var}\big[f(X)-\hat{f}(X)|\hat{f}\big].
\end{align}
Hence, 
\begin{align}
\mathrm{E}\Big[\Big(Y-\hat{f}(X)\Big)^2\Big] 
= & \ \mathrm{E}\Big[\Big(\mathrm{E}[f(X)-\hat{f}(X)|\hat{f}]\Big)^2\Big] + \mathrm{E}\Big[\mathrm{Var}\big[f(X)-\hat{f}(X)|\hat{f}\big]\Big] + \sigma_{\varepsilon}^2. 
\end{align}

\subsection{Proof of Proposition \ref{proposition_decorrelation}}
Showing (\ref{bias_ensemble}) is straightforward from the law of total expectation and the linearity of the expectation, noting that $\hat{f}_1(x),\dots,\hat{f}_B(x)$ are i.i.d conditionally on $D_n$. To prove (\ref{variance_ensemble}), note that, since $\hat{f}_1(x),\dots,\hat{f}_B(x)$ are i.i.d conditionally on $D_n$, the pairs $\{(\hat{f}_i(x),\hat{f}_j(x))\}_{i<j}$ are also i.i.d conditionally on $D_n$, and so are the squares $\hat{f}_1(x)^2,\dots,\hat{f}_B(x)^2$. Unconditionally with respect to $D_n$, tree estimates are no longer independent but remain identically distributed, and so are pairs and squares. Hence, by linearity of the expectation, we have 
\begin{equation}
\mathrm{E}\Bigg[\sum_{b=1}^B\hat{f}_b(x)\Bigg] = B \mathrm{E}[\hat{f}_1(x)]
\end{equation}
and 
\begin{equation}
\mathrm{E}\Bigg[\Bigg(\sum_{b=1}^B\hat{f}_b(x)\Bigg)^2\Bigg] = B(B-1)\mathrm{E}[\hat{f}_1(x)\hat{f}_2(x)]+B\mathrm{E}[\hat{f}_1(x)^2].
\end{equation}
Therefore
\begin{align}
\mathrm{Var}[\hat{f}(x)] 
&=\frac{1}{B^2}\Big\{B(B-1)\mathrm{E}[\hat{f}_1(x)\hat{f}_2(x)]+B\mathrm{E}[\hat{f}_1(x)^2]-B^2\mathrm{E}[\hat{f}_1(x)]^2\Big\}  \\
&=\frac{1}{B^2}\Big\{B(B-1)\Big(\mathrm{Cov}[\hat{f}_1(x), \hat{f}_2(x)]+\mathrm{E}[\hat{f}_1(x)]^2)\Big) \nonumber \\
& \ \ \ \ \ \ \ \ \ \ +B\Big(\mathrm{Var}[\hat{f}_1(x)]+\mathrm{E}[\hat{f}_1(x)]^2\Big)-B^2\mathrm{E}[\hat{f}_1(x)]^2\Big\}\notag \nonumber \\
&=\mathrm{Cov}[\hat{f}_1(x),\hat{f}_2(x)] + \frac{1}{B}\Bigg(\mathrm{Var}[\hat{f}_1(x)]-\mathrm{Cov}[\hat{f}_1(x),\hat{f}_2(x)]\Bigg). \nonumber 
\end{align}
Finally, take $B$ large to get rid of the second term in the last equality, and notice that $\hat{f}_1(x)$ and $\hat{f}_2(x)$ have same standard deviation.

\subsection{Proof of Proposition \ref{proposition_relative_performance}}
For both bagging and forests we have
\begin{align}
\mathrm{E}\Big[\Big(Z-\hat{g}(X)\Big)^2\Big] &= \mathrm{E}\Big[\Big(g(X)-\hat{g}(X)\Big)^2\Big] + \sigma_{\eta}^2 \\
& = \frac{1}{\sigma_{f}^2}\Bigg\{\mathrm{E}\Big[\Big(f(X)-\hat{f}(X)\Big)^2\Big] + \sigma_{\varepsilon}^2\Bigg\}  \nonumber \\
& = \frac{1}{\sigma_{f}^2} \mathrm{E}\Big[\Big(Y-\hat{f}(X)\Big)^2\Big]  \nonumber 
\end{align}
hence, when taking the relative difference of the two methods, the factor $1/\sigma_{f}^2$ simplifies from all terms, which concludes the proof.

\newpage
\section{Additional simulations for Section \ref{section_findings}}
\label{appendix_plots_extensions}

\subsection{Irrelevant Covariates}
\label{appendix_irrelevant_covariates_plots}

\begin{figure}[htbp]
\begin{subfigure}{0.32\textwidth}
	\includegraphics[width=\textwidth]{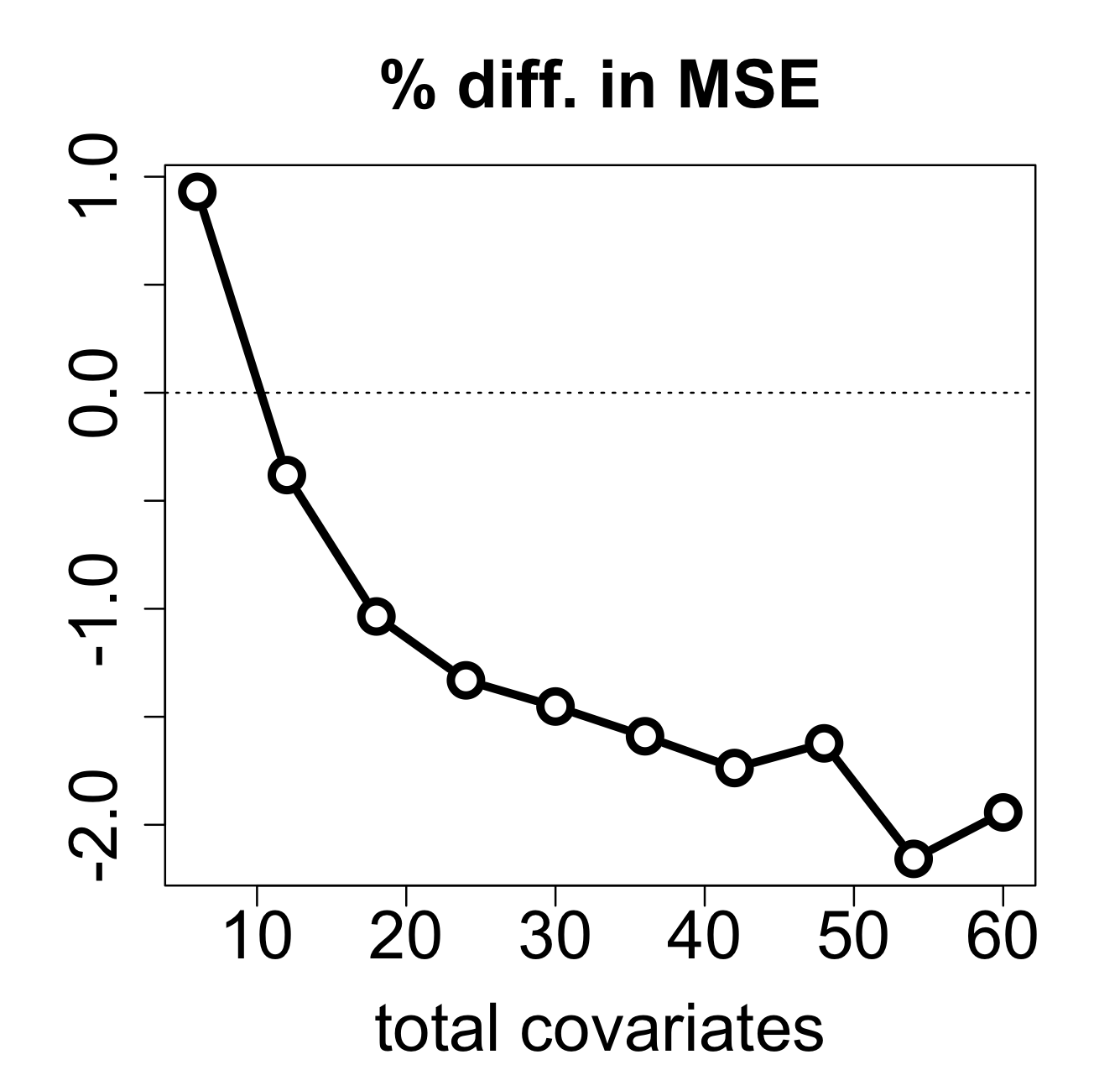}
\end{subfigure}
\begin{subfigure}{0.32\textwidth}
	\includegraphics[width=\textwidth]{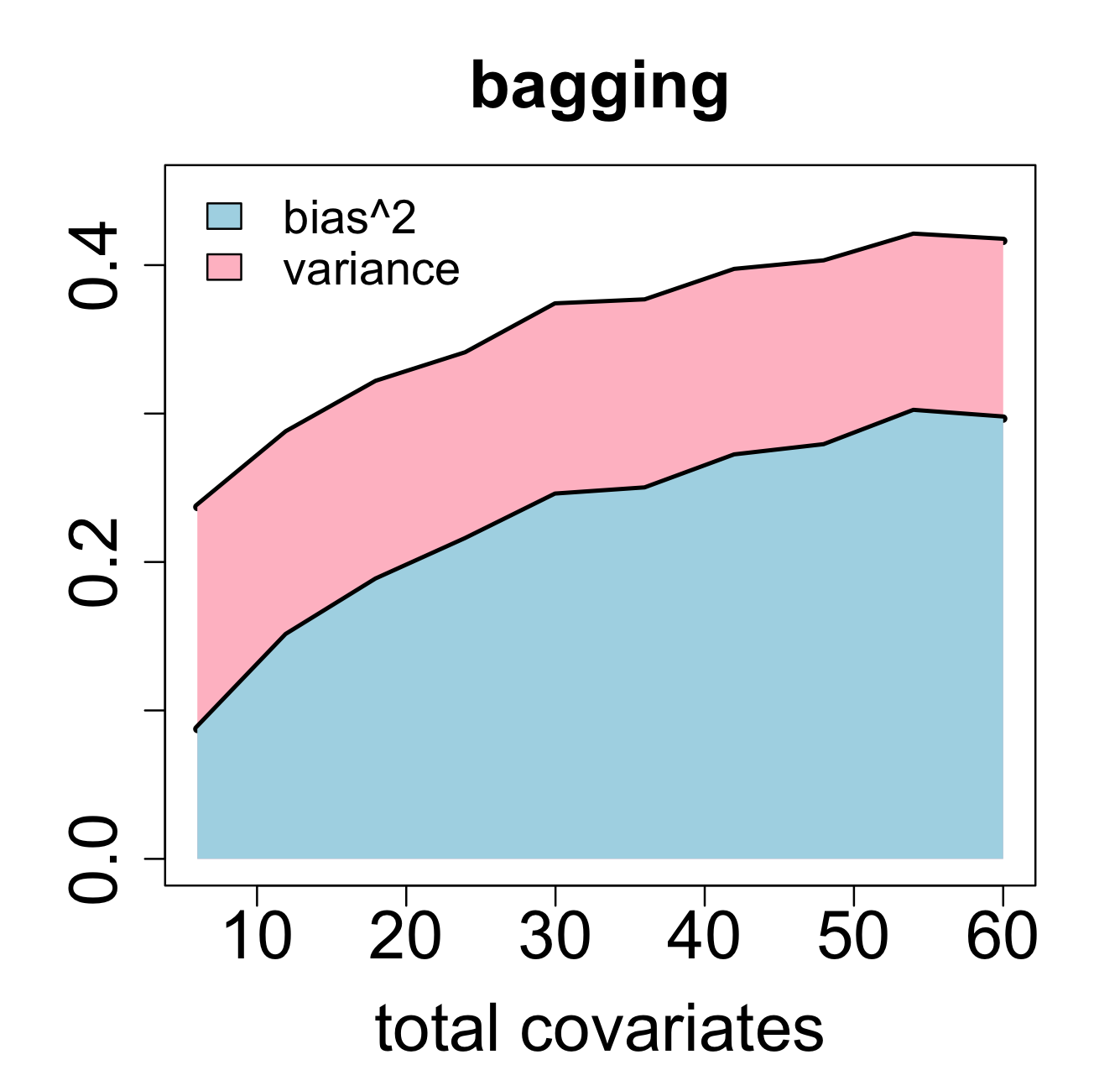}
\end{subfigure}
\begin{subfigure}{0.32\textwidth}
	\includegraphics[width=\textwidth]{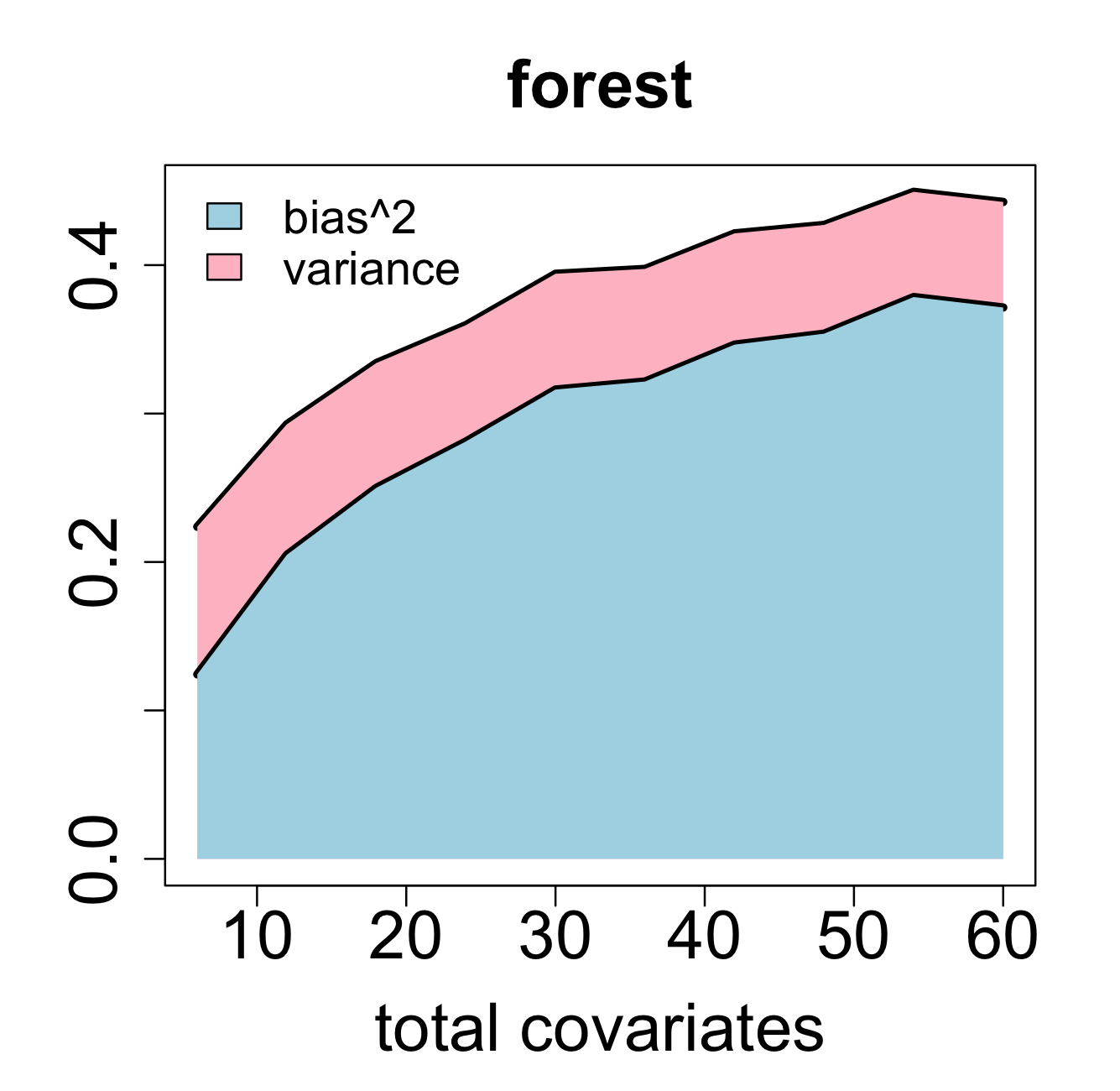}
\end{subfigure}
\caption{ Effect of irrelevant covariates for $\mathcal{N}$-LINEAR.}
\label{plots_irrelevant_covariates_N_LINEAR}
\end{figure}

\begin{figure}[htbp]
\begin{subfigure}{0.32\textwidth}
	\includegraphics[width=\textwidth]{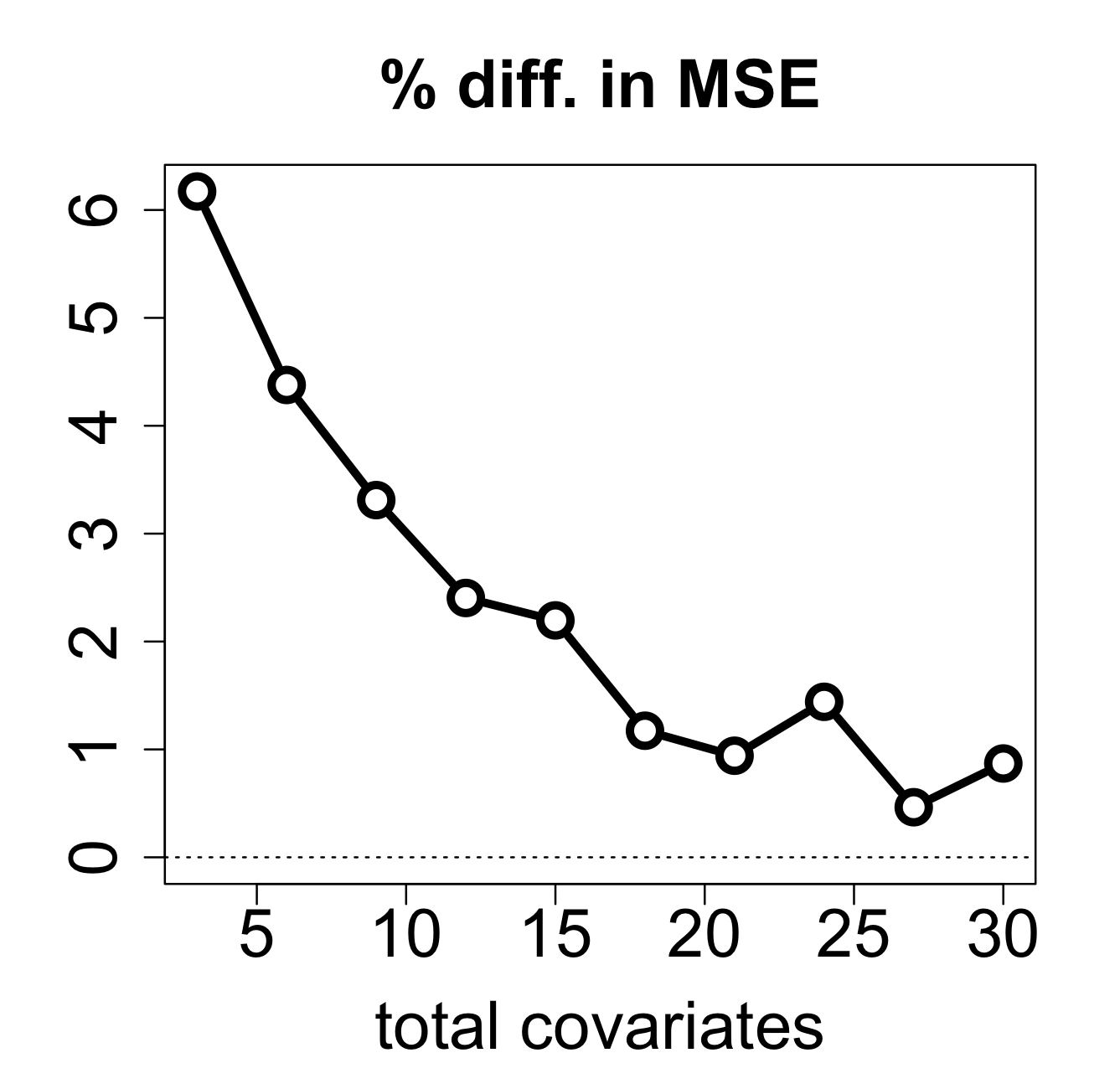}
\end{subfigure}
\begin{subfigure}{0.32\textwidth}
	\includegraphics[width=\textwidth]{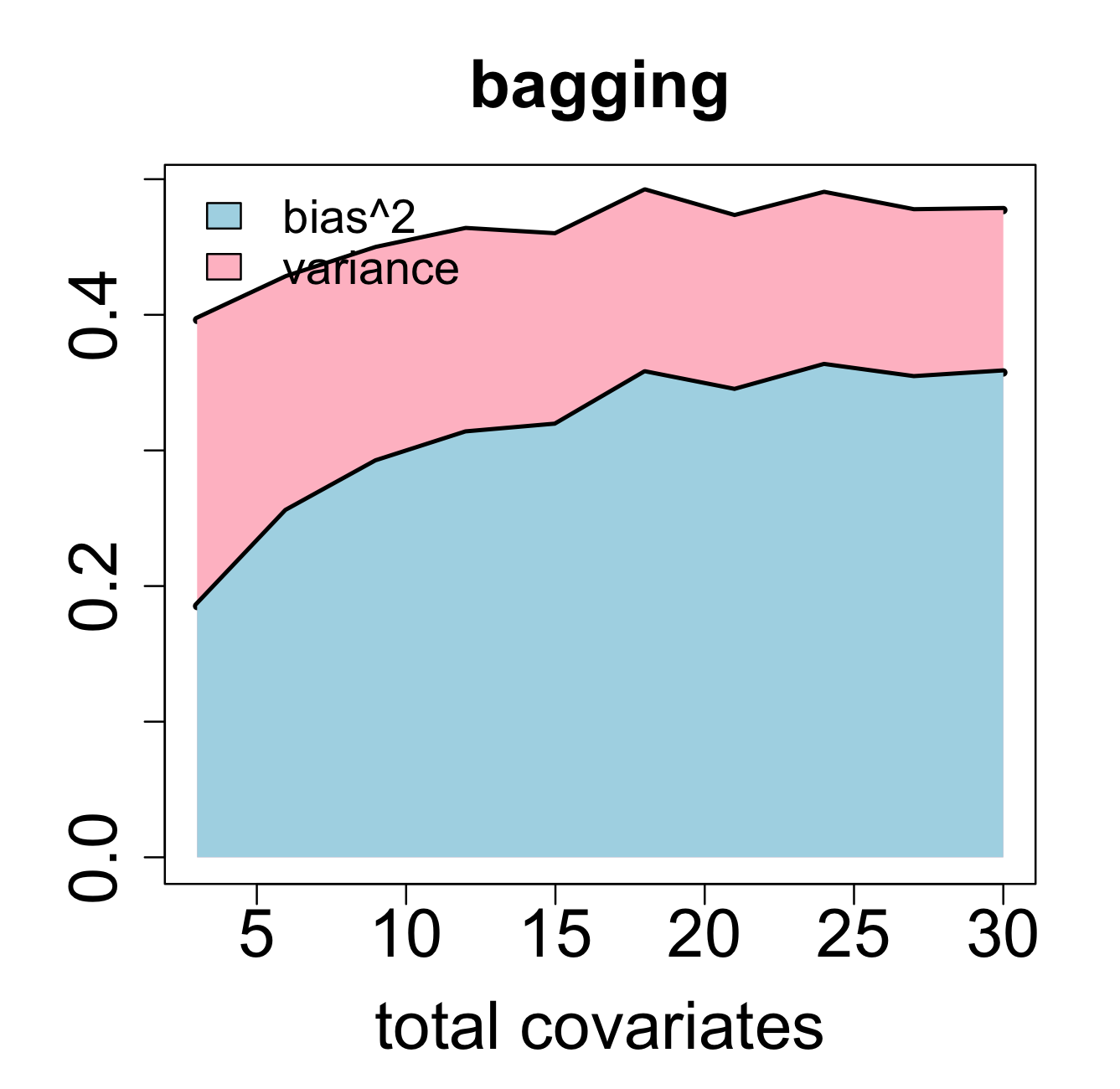}
\end{subfigure}
\begin{subfigure}{0.32\textwidth}
	\includegraphics[width=\textwidth]{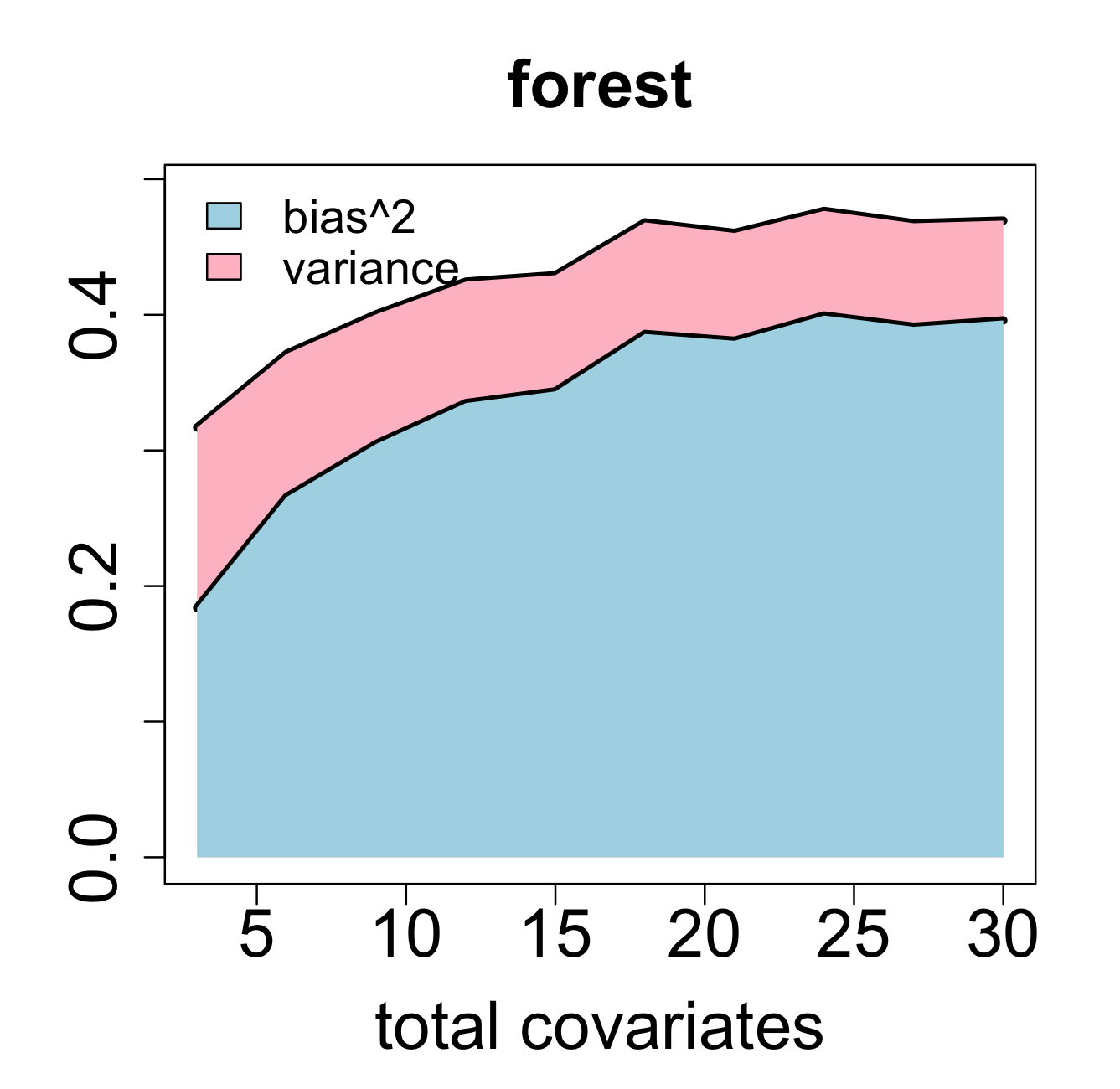}
\end{subfigure}
\caption{ Effect of irrelevant covariates for $\mathcal{U}$-HIDDEN.}
\label{plots_irrelevant_covariates_U_HIDDEN}
\end{figure}

\newpage
\subsection{Correlated Covariates}
\label{appendix_dependencies_plots}

\begin{figure}[htbp]
\begin{subfigure}{0.32\textwidth}
	\includegraphics[width=\textwidth]{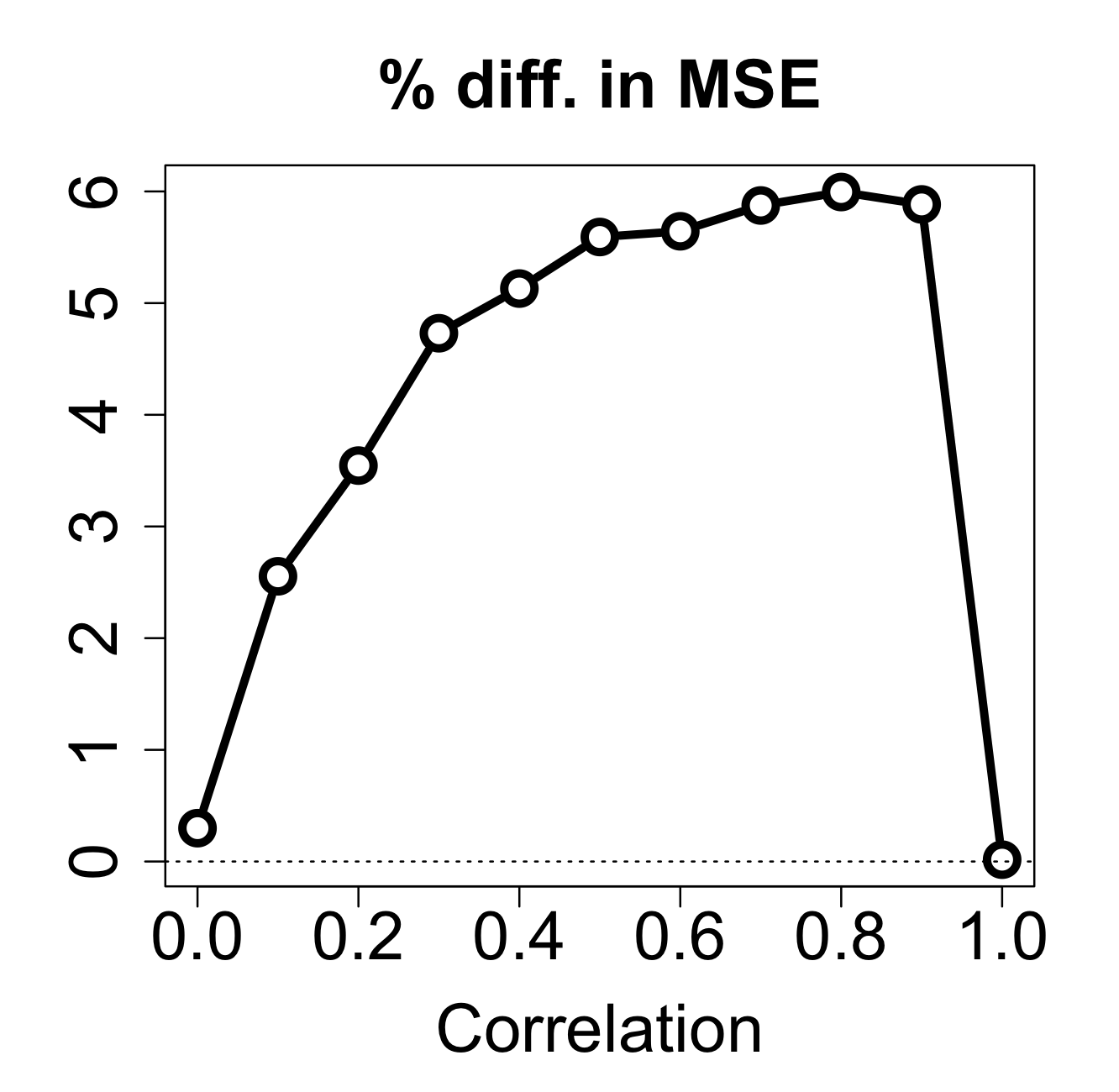}
\end{subfigure}
\begin{subfigure}{0.32\textwidth}
	\includegraphics[width=\textwidth]{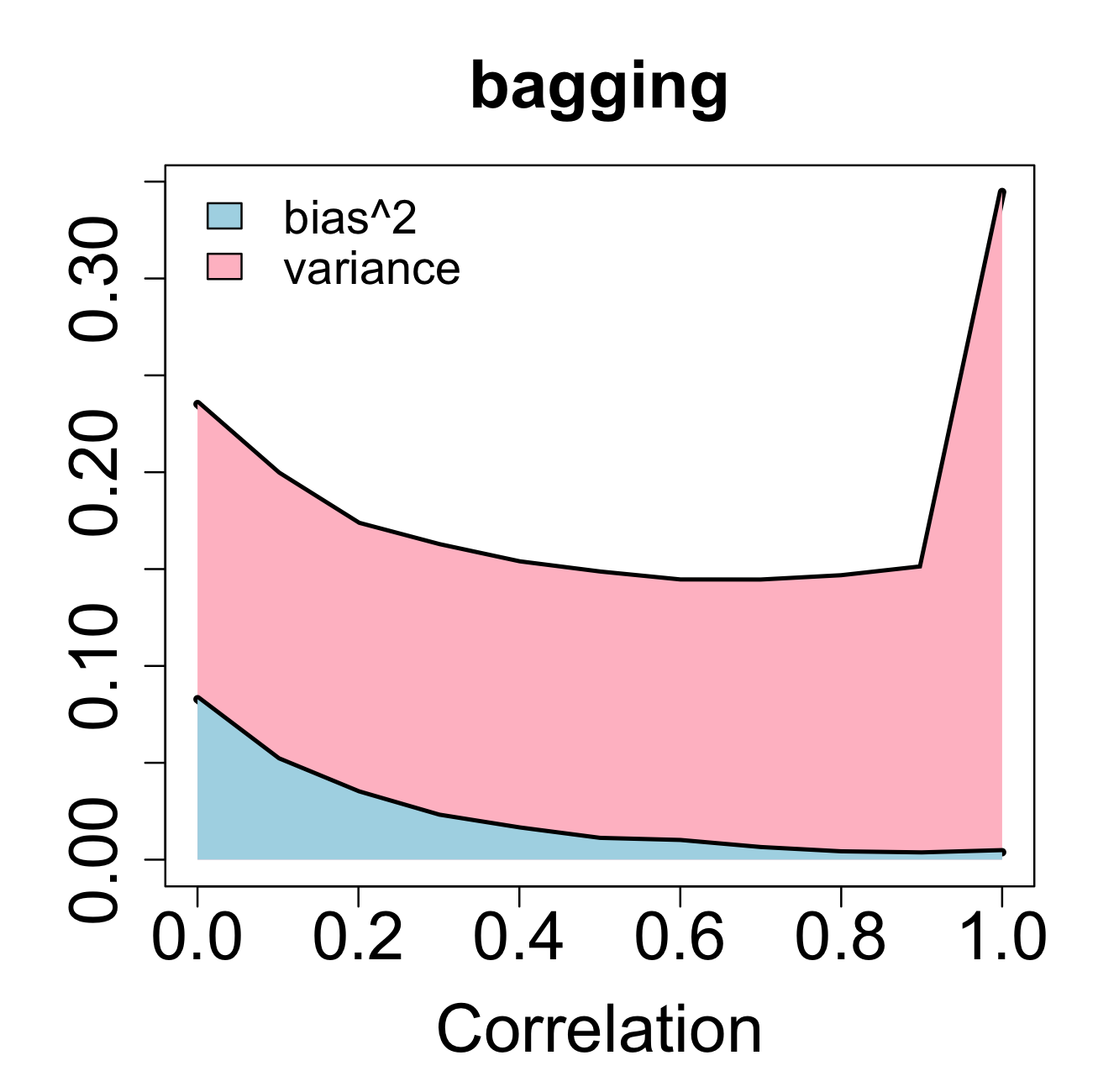}
\end{subfigure}
\begin{subfigure}{0.32\textwidth}
	\includegraphics[width=\textwidth]{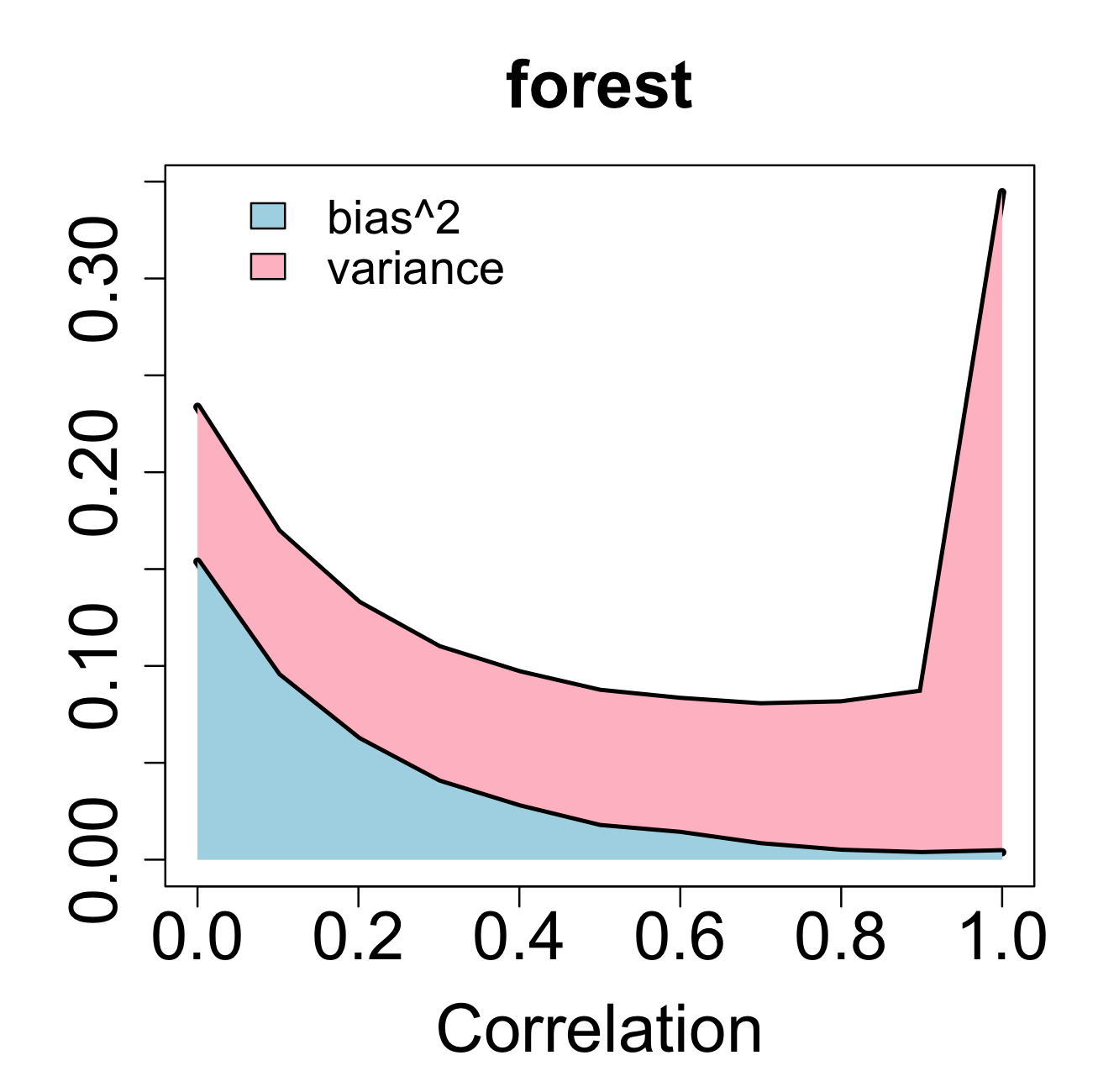}
\end{subfigure}
\caption{ Effect of correlated covariates for $\mathcal{N}$-LINEAR.}
\label{plots_dependencies_N_LINEAR}
\end{figure}

\begin{figure}[htbp]
\begin{subfigure}{0.32\textwidth}
	\includegraphics[width=\textwidth]{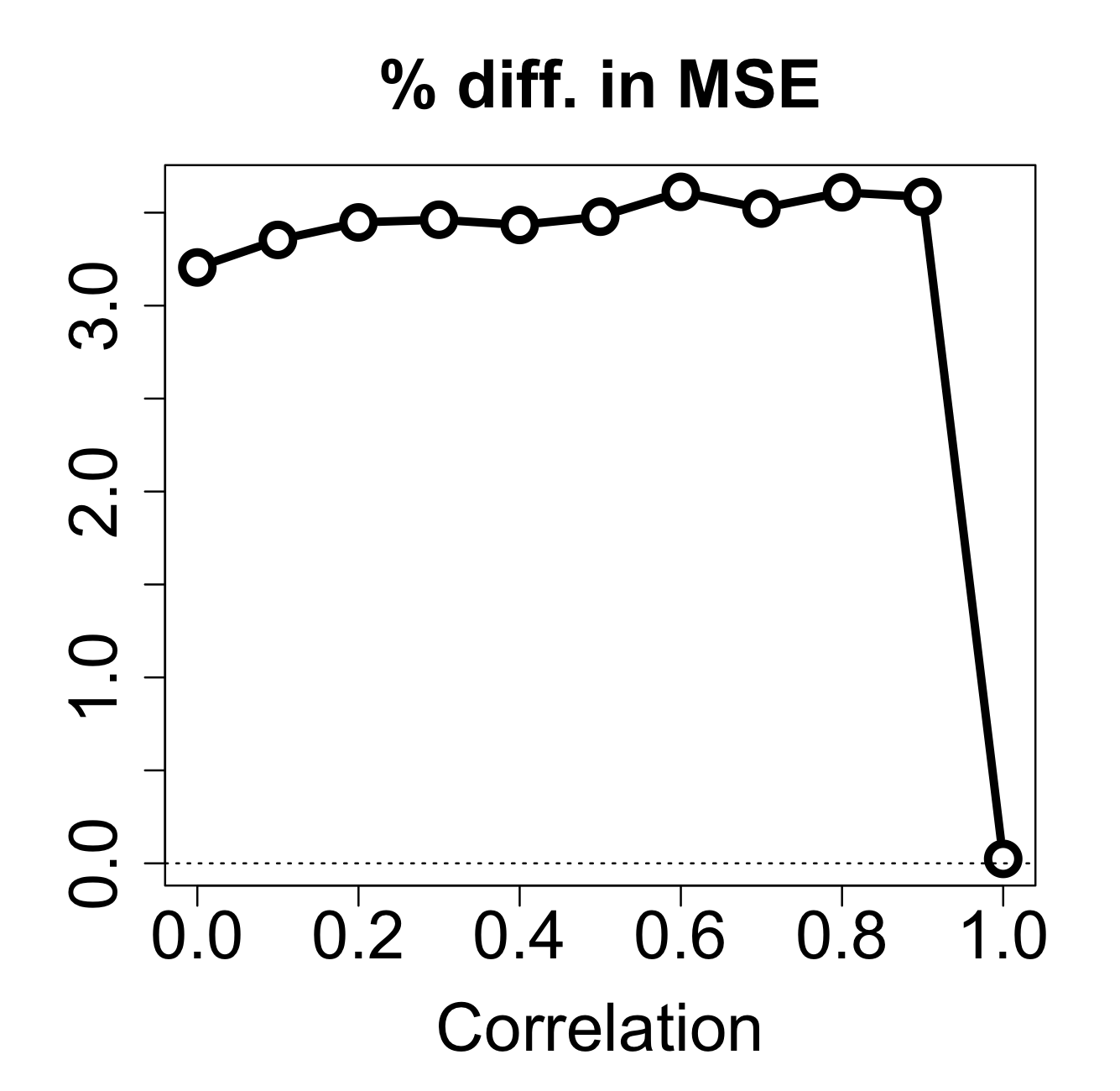}
\end{subfigure}
\begin{subfigure}{0.32\textwidth}
	\includegraphics[width=\textwidth]{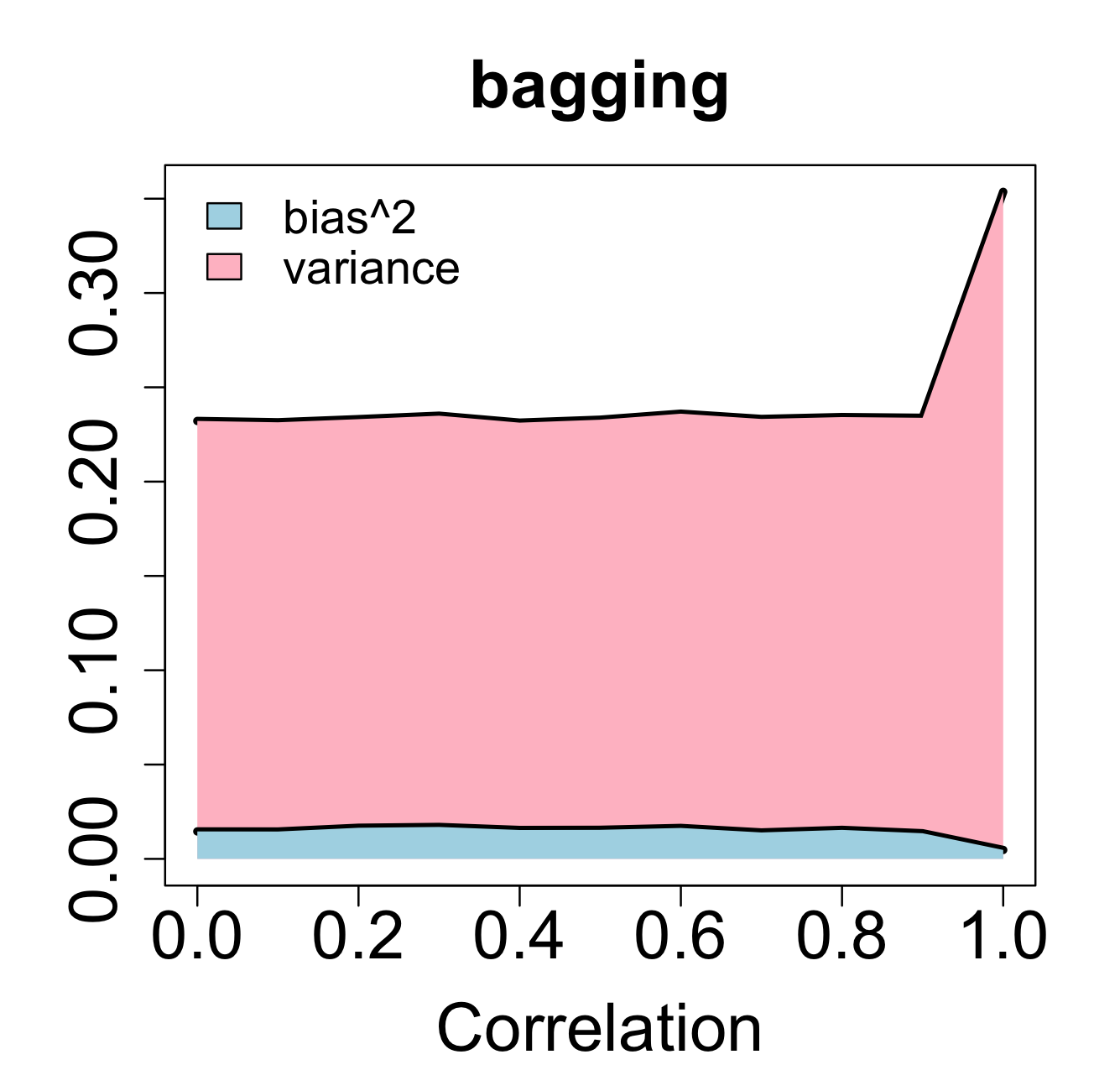}
\end{subfigure}
\begin{subfigure}{0.32\textwidth}
	\includegraphics[width=\textwidth]{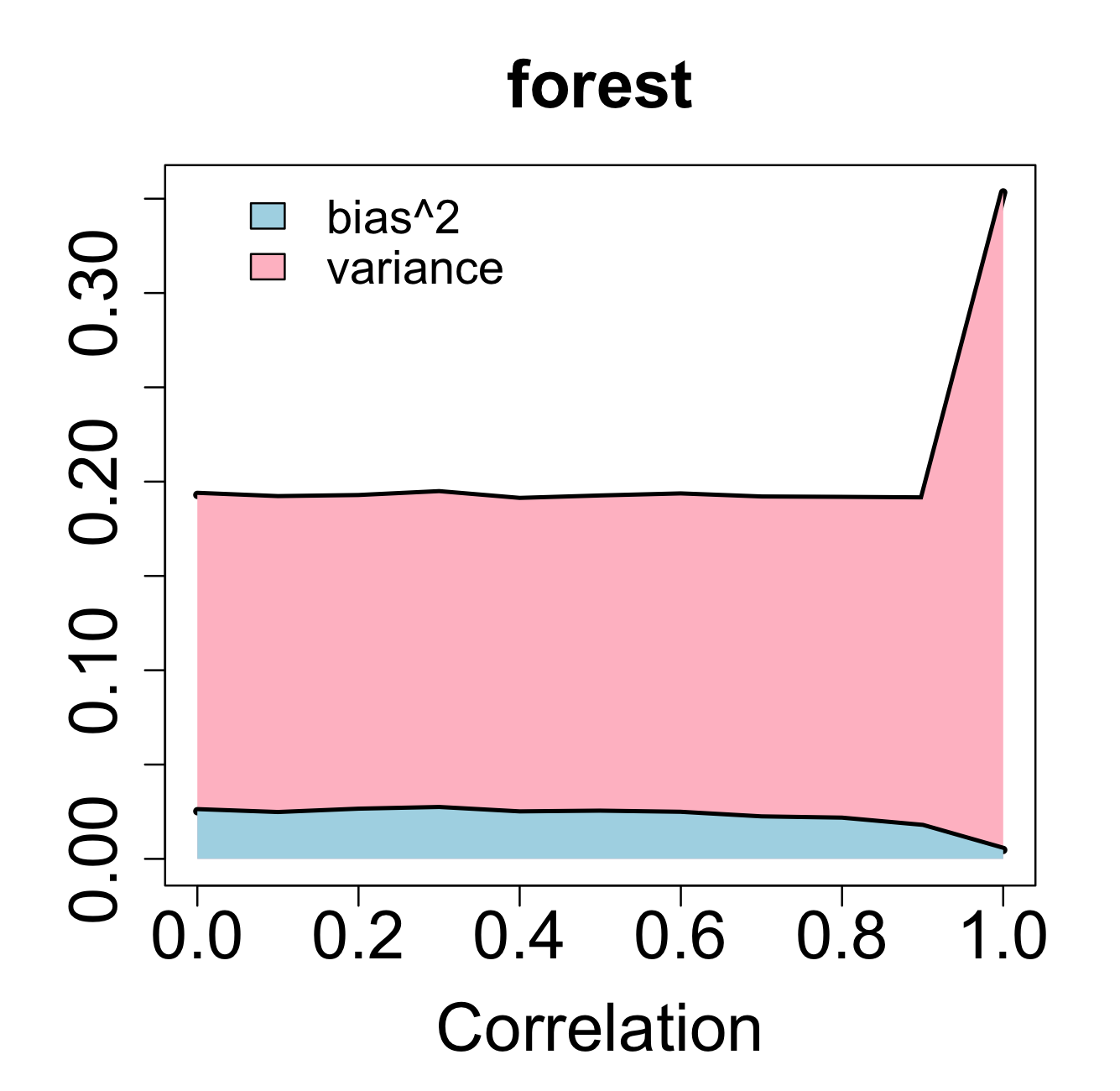}
\end{subfigure}
\caption{ Effect of correlated covariates for $\mathcal{N}$-HIDDEN.}
\label{plots_dependencies_N_HIDDEN}
\end{figure}

\newpage
\section{Tables of Section \ref{section_findings} for low and high SNR}
\label{appendix_simulations_SNR}

\subsection{Distribution of $X_j$}
Here we replicate Table \ref{extension_table_distribution_of_X} (SNR=1) for low and high SNR (0.05 and 6 respectively).
\begin{table}[ht]
\small
\centering
\begin{tabular}{ccccccc}
  & \multicolumn{2}{c}{LINEAR} & \multicolumn{2}{c}{MARS} & \multicolumn{2}{c}{HIDDEN} \\
    \hline
  distribution of $X_j$ & $\mathcal{N}(0,1)$ & $\mathcal{U}(0,1)$ & $\mathcal{N}(0,1)$ & $\mathcal{U}(0,1)$ & $\mathcal{N}(0,1)$ & $\mathcal{U}(0,1)$ \\
  \hline
  $Bias^2$ bagging & 0.05 & 0.03 & 0.15 & 0.06 & 0.02 & 0.13 \\ 
  $Bias^2$ forest & 0.11 & 0.08 & 0.30 & 0.12 & 0.02 & 0.13 \\ 
  Variance bagging & 2.17 & 2.20 & 2.30 & 2.23 & 3.74 & 3.80 \\ 
  Variance forest & 1.28 & 1.29 & 1.36 & 1.33 & 3.07 & 3.13 \\ 
  Tree variance bagging & 20.04 & 20.35 & 20.25 & 20.15 & 18.03 & 18.50 \\ 
  Tree variance forest & 17.71 & 17.97 & 18.32 & 17.87 & 17.02 & 17.31 \\ 
  Correlation bagging & 0.10 & 0.11 & 0.11 & 0.11 & 0.20 & 0.20 \\ 
  Correlation forest & 0.07 & 0.07 & 0.07 & 0.07 & 0.18 & 0.17 \\ 
  Irreducible & 20.11 & 20.32 & 20.11 & 20.32 & 20.00 & 19.62 \\ 
  MSE bagging & 22.29 & 22.56 & 22.61 & 22.60 & 23.78 & 23.55 \\ 
  MSE forest & 21.46 & 21.72 & 21.82 & 21.76 & 23.12 & 22.88 \\ 
  test statistic & 70.98 & 66.12 & 58.75 & 66.43 & 76.20 & 74.43 \\ 
  \hline
  relative difference (\%)  & \textbf{3.89} & \textbf{3.88} & \textbf{3.60} & \textbf{3.86} & \textbf{2.87} & \textbf{2.92} \\ 
   \hline
\end{tabular}
\caption{Change in the distribution of $X_j$ -- \textbf{SNR = 0.05}. } 
\label{extension_table_distribution_of_X_low_SNR}
\end{table}

\begin{table}[ht]
\small
\centering
\begin{tabular}{ccccccc}
  & \multicolumn{2}{c}{LINEAR} & \multicolumn{2}{c}{MARS} & \multicolumn{2}{c}{HIDDEN} \\
    \hline
  distribution of $X_j$ & $\mathcal{N}(0,1)$ & $\mathcal{U}(0,1)$ & $\mathcal{N}(0,1)$ & $\mathcal{U}(0,1)$ & $\mathcal{N}(0,1)$ & $\mathcal{U}(0,1)$ \\
  \hline
  $Bias^2$ bagging & 0.12 & 0.08 & 0.10 & 0.06 & 0.01 & 0.17 \\ 
  $Bias^2$ forest & 0.18 & 0.14 & 0.39 & 0.15 & 0.03 & 0.08 \\ 
  Variance bagging & 0.06 & 0.06 & 0.06 & 0.05 & 0.04 & 0.10 \\ 
  Variance forest & 0.03 & 0.03 & 0.04 & 0.03 & 0.04 & 0.05 \\ 
  Tree variance bagging & 0.52 & 0.50 & 0.35 & 0.38 & 0.17 & 0.42 \\ 
  Tree variance forest & 0.64 & 0.63 & 0.75 & 0.63 & 0.27 & 0.40 \\ 
  Correlation bagging & 0.10 & 0.11 & 0.15 & 0.14 & 0.24 & 0.23 \\ 
  Correlation forest & 0.04 & 0.04 & 0.05 & 0.05 & 0.13 & 0.13 \\ 
  Irreducible & 0.17 & 0.17 & 0.17 & 0.17 & 0.17 & 0.16 \\ 
  MSE bagging & 0.343 & 0.303 & 0.334 & 0.285 & 0.226 & 0.427 \\ 
  MSE forest & 0.377 & 0.343 & 0.604 & 0.341 & 0.231 & 0.297 \\ 
  test statistic & -52.27 & -59.13 & -212.90 & -67.34 & -14.76 & 66.32 \\ 
  \hline
  relative difference (\%)  & \textbf{-8.79} & \textbf{-11.61} & \textbf{-44.78} & \textbf{-16.24} & \textbf{-2.25} & \textbf{43.73} \\ 
   \hline
\end{tabular}
\caption{Change in the distribution of $X_j$ -- \textbf{SNR = 6}. } 
\label{extension_table_distribution_of_X_high_SNR}
\end{table}

\newpage
\subsection{Irrelevant Covariates}
Here we replicate Table \ref{extension_table_irrelevant_covariates} (SNR=1) for low and high SNR (0.05 and 6 respectively).

\begin{table}[ht]
\small
\centering
\begin{tabular}{ccccccc}
  & \multicolumn{2}{c}{$\mathcal{N}$-LINEAR} & \multicolumn{2}{c}{$\mathcal{U}$-MARS} & \multicolumn{2}{c}{$\mathcal{U}$-HIDDEN} \\
  \hline
  Relevant covariates & 5 & 5 & 5 & 5 & 2 & 2 \\
  Irrelevant covariates & 1 & 25 & 1 & 25 & 1 & 13 \\
  Total covariates ($p$) & 6 & 30 & 6 & 30 & 3 & 15 \\
  \hline
  $Bias^2$ bagging & 0.05 & 0.35 & 0.06 & 0.30 & 0.20 & 0.36 \\ 
  $Bias^2$ forest & 0.08 & 0.40 & 0.09 & 0.37 & 0.22 & 0.41 \\ 
  Variance bagging & 2.05 & 1.14 & 2.06 & 1.17 & 2.91 & 1.49 \\ 
  Variance forest & 1.52 & 0.87 & 1.52 & 0.89 & 2.10 & 1.12 \\ 
  Tree variance bagging & 20.38 & 21.27 & 20.76 & 21.43 & 19.58 & 21.22 \\ 
  Tree variance forest & 19.23 & 20.89 & 19.63 & 21.08 & 17.59 & 20.93 \\ 
  Correlation bagging & 0.10 & 0.05 & 0.10 & 0.05 & 0.14 & 0.07 \\ 
  Correlation forest & 0.08 & 0.04 & 0.08 & 0.04 & 0.12 & 0.05 \\ 
  Irreducible & 20.29 & 19.64 & 19.59 & 20.23 & 19.89 & 19.94 \\ 
  MSE bagging & 22.41 & 21.08 & 21.67 & 21.67 & 22.98 & 21.77 \\ 
  MSE forest & 21.92 & 20.86 & 21.15 & 21.46 & 22.18 & 21.44 \\ 
  test statistic & 55.34 & 37.72 & 58.60 & 33.30 & 69.27 & 38.43 \\ 
  \hline
  relative difference (\%)  & \textbf{2.24} & \textbf{1.03} & \textbf{2.45} & \textbf{0.97} & \textbf{3.59} & \textbf{1.53} \\ 
   \hline
\end{tabular}
\caption{Adding irrelevant covariates -- \textbf{SNR = 0.05}. } 
\label{extension_table_irrelevant_covariates_low_SNR}
\end{table}

\begin{table}[ht]
\small
\centering
\begin{tabular}{ccccccc}
  & \multicolumn{2}{c}{$\mathcal{N}$-LINEAR} & \multicolumn{2}{c}{$\mathcal{U}$-MARS} & \multicolumn{2}{c}{$\mathcal{U}$-HIDDEN} \\
  \hline
  Relevant covariates & 5 & 5 & 5 & 5 & 2 & 2 \\
  Irrelevant covariates & 1 & 25 & 1 & 25 & 1 & 13 \\
  Total covariates ($p$) & 6 & 30 & 6 & 30 & 3 & 15 \\
  \hline
  $Bias^2$ bagging & 0.13 & 0.24 & 0.07 & 0.14 & 0.22 & 0.33 \\ 
  $Bias^2$ forest & 0.16 & 0.31 & 0.10 & 0.20 & 0.17 & 0.33 \\ 
  Variance bagging & 0.06 & 0.07 & 0.05 & 0.06 & 0.09 & 0.08 \\ 
  Variance forest & 0.04 & 0.04 & 0.03 & 0.03 & 0.04 & 0.04 \\
  Tree variance bagging & 0.56 & 0.78 & 0.40 & 0.54 & 0.47 & 0.58 \\ 
  Tree variance forest & 0.63 & 0.87 & 0.56 & 0.74 & 0.54 & 0.75 \\ 
  Correlation bagging & 0.11 & 0.09 & 0.13 & 0.10 & 0.18 & 0.13 \\ 
  Correlation forest & 0.06 & 0.04 & 0.06 & 0.03 & 0.08 & 0.05 \\ 
  Irreducible & 0.17 & 0.16 & 0.16 & 0.17 & 0.17 & 0.17 \\ 
  MSE bagging & 0.36 & 0.47 & 0.28 & 0.35 & 0.48 & 0.57 \\ 
  MSE forest & 0.37 & 0.51 & 0.29 & 0.39 & 0.38 & 0.52 \\ 
  test statistic & -23.63 & -56.12 & -11.09 & -45.13 & 45.57 & 30.14 \\ 
  \hline
  relative difference (\%)  & \textbf{-3.20} & \textbf{-7.81} & \textbf{-2.53} & \textbf{-9.38} & \textbf{26.04} & \textbf{9.57} \\ 
   \hline
\end{tabular}
\caption{Adding irrelevant covariates -- \textbf{SNR = 6}. } 
\label{extension_table_irrelevant_covariates_high_SNR}
\end{table}

\newpage
\subsection{Correlated Covariates}
Here we replicate Table \ref{extension_table_dependencies} (SNR=1) for low and high SNR (0.05 and 6 respectively).

\begin{table}[ht]
\small
\centering
\begin{tabular}{ccccccc}
  & \multicolumn{2}{c}{$\mathcal{N}$-LINEAR} & \multicolumn{2}{c}{$\mathcal{N}$-MARS} & \multicolumn{2}{c}{$\mathcal{N}$-HIDDEN} \\
  \hline
  correlation $\rho$ & 0 & 0.5 & 0 & 0.5 & 0 & 0.5 \\
  \hline
  $Bias^2$ bagging & 0.05 & 0.01 & 0.15 & 0.14 & 0.02 & 0.02 \\ 
  $Bias^2$ forest & 0.11 & 0.01 & 0.30 & 0.25 & 0.02 & 0.03 \\ 
  Variance bagging & 2.17 & 2.33 & 2.30 & 2.41 & 3.74 & 3.83 \\ 
  Variance forest & 1.28 & 1.39 & 1.36 & 1.45 & 3.07 & 3.15 \\
  Tree variance bagging & 20.04 & 19.76 & 20.25 & 20.06 & 18.03 & 18.05 \\ 
  Tree variance forest & 17.71 & 17.61 & 18.32 & 17.87 & 17.02 & 17.02 \\ 
  Correlation bagging & 0.10 & 0.11 & 0.11 & 0.11 & 0.20 & 0.21 \\ 
  Correlation forest & 0.07 & 0.07 & 0.07 & 0.08 & 0.18 & 0.18 \\
  Irreducible & 20.11 & 20.11 & 20.11 & 20.11 & 20.00 & 20.00 \\ 
  MSE bagging & 22.29 & 22.45 & 22.61 & 22.69 & 23.78 & 23.84 \\ 
  MSE forest & 21.46 & 21.51 & 21.82 & 21.82 & 23.12 & 23.16 \\ 
  test statistic & 70.98 & 72.66 & 58.75 & 67.22 & 76.20 & 76.62 \\ 
  \hline
  relative difference (\%)  & \textbf{3.89} & \textbf{4.36} & \textbf{3.60} & \textbf{3.95} & \textbf{2.87} & \textbf{2.93} \\ 
   \hline
\end{tabular}
\caption{Adding correlation -- \textbf{SNR = 0.05}. } 
\label{extension_table_dependencies_low_SNR}
\end{table}

\begin{table}[ht]
\small
\centering
\begin{tabular}{ccccccc}
  & \multicolumn{2}{c}{$\mathcal{N}$-LINEAR} & \multicolumn{2}{c}{$\mathcal{N}$-MARS} & \multicolumn{2}{c}{$\mathcal{N}$-HIDDEN} \\
  \hline
  correlation $\rho$ & 0 & 0.5 & 0 & 0.5 & 0 & 0.5 \\
  \hline
  $Bias^2$ bagging & 0.12 & 0.02 & 0.10 & 0.09 & 0.01 & 0.02 \\ 
  $Bias^2$ forest & 0.18 & 0.03 & 0.39 & 0.29 & 0.03 & 0.03 \\ 
  Variance bagging & 0.06 & 0.04 & 0.06 & 0.06 & 0.04 & 0.04 \\ 
  Variance forest & 0.03 & 0.02 & 0.04 & 0.04 & 0.04 & 0.03 \\ 
  Tree variance bagging & 0.52 & 0.28 & 0.35 & 0.33 & 0.17 & 0.17 \\ 
  Tree variance forest & 0.64 & 0.32 & 0.75 & 0.67 & 0.27 & 0.25 \\ 
  Correlation bagging & 0.10 & 0.12 & 0.15 & 0.15 & 0.24 & 0.25 \\ 
  Correlation forest & 0.04 & 0.05 & 0.05 & 0.05 & 0.13 & 0.14 \\ 
  Irreducible & 0.17 & 0.17 & 0.17 & 0.17 & 0.17 & 0.17 \\ 
  MSE bagging & 0.343 & 0.223 & 0.334 & 0.321 & 0.226 & 0.227 \\ 
  MSE forest & 0.377 & 0.211 & 0.604 & 0.505 & 0.231 & 0.228 \\ 
  test statistic & -52.27 & 38.80 & -212.90 & -182.35 & -14.76 & -4.03 \\ 
  \hline
  relative difference (\%)  & \textbf{-8.79} & \textbf{5.76} & \textbf{-44.78} & \textbf{-36.52} & \textbf{-2.25} & \textbf{-0.54} \\ 
   \hline
\end{tabular}
\caption{Adding correlation -- \textbf{SNR = 6}. } 
\label{extension_table_dependencies_high_SNR}
\end{table}


\newpage
\vskip 0.2in
\bibliography{biblio}

\end{document}